\documentclass[letterpaper]{article} %
\usepackage[draft]{aaai2026}  %
\usepackage{times}  %
\usepackage{helvet}  %
\usepackage{courier}  %
\usepackage[hyphens]{url}  %
\usepackage{graphicx} %
\usepackage{natbib}  %
\usepackage{caption} %
\usepackage{algorithm}
\usepackage{algorithmic}

\usepackage{newfloat}
\usepackage{listings}
\DeclareCaptionStyle{ruled}{labelfont=normalfont,labelsep=colon,strut=off} %
\floatstyle{ruled}
\newfloat{listing}{tb}{lst}{}
\floatname{listing}{Listing}
\usepackage{amsmath}
\usepackage{subcaption}
\usepackage[capitalize,noabbrev]{cleveref}

\usepackage{multirow}
\usepackage{tikz}
\usepackage{pgfplots}
\usepackage{algorithm}
\usepackage{algorithmic}

\usepackage{comment}
\usepackage{array}

\usepackage{alphalph}

\usepackage{listings}
\usepackage{fancyvrb}
\usepackage{fvextra}

\def\ours{Unmasking with Contrastive Attention Guidance}
\def\oursabbr{UNCAGE}

\usepackage{booktabs}
\usepackage{titlecaps}
\usepackage{enumitem}
\usepackage{placeins}

\usepackage{amssymb}

\usepackage{colortbl}

\colorlet{avg}{blue!5}

\title{UNCAGE: Contrastive Attention Guidance for Masked Generative Transformers in Text-to-Image Generation}

\author {
    Wonjun Kang\textsuperscript{\rm 1,2}\quad
    Byeongkeun Ahn\textsuperscript{\rm 3}\quad
    Minjae Lee\textsuperscript{\rm 2}\quad
    Kevin Galim\textsuperscript{\rm 2}\quad
    Seunghyuk Oh\textsuperscript{\rm 2}\\
    Hyung Il Koo\textsuperscript{\rm 2,4}\quad
    Nam Ik Cho\textsuperscript{\rm 1}
}
\affiliations {
    \textsuperscript{\rm 1}Seoul National University\quad
    \textsuperscript{\rm 2}FuriosaAI\quad
    \textsuperscript{\rm 3}Independent Researcher\quad
    \textsuperscript{\rm 4}Ajou University\\
    {\tt\small \{kangwj1995, minjae.lee, kevin.galim, seunghyukoh, hikoo\}@furiosa.ai, byeongkeunahn0@gmail.com, nicho@snu.ac.kr}
}

\usepackage{bibentry}

\begin{document}

\twocolumn[{
\renewcommand\twocolumn[1][]{#1}
\maketitle
\begin{center}
    \centering
    \includegraphics[width=\linewidth]{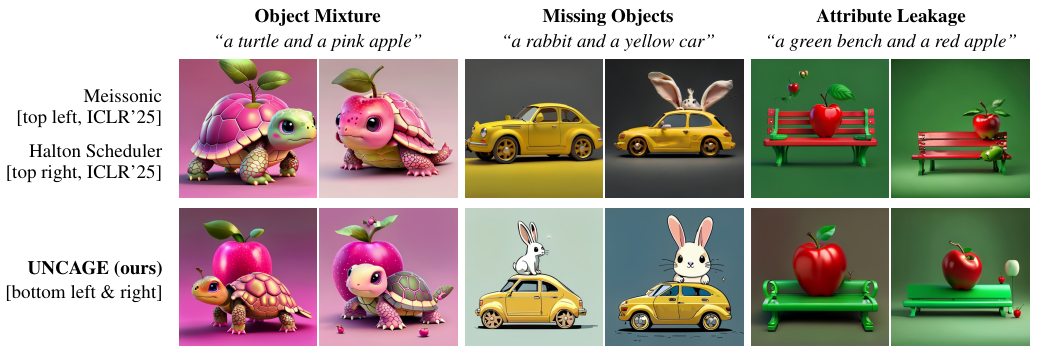}
    \captionof{figure}{
    Our training-free, \textit{\ours{} (\oursabbr{})} enhances the performance of Masked Generative Transformers in compositional T2I generation, with negligible inference overhead.
    }
    \vspace{0.5cm}
\label{fig:teaser}
\end{center}
}
]

\begin{abstract}
Text-to-image (T2I) generation has been actively studied using Diffusion Models and Autoregressive Models. Recently, Masked Generative Transformers have gained attention as an alternative to Autoregressive Models to overcome the inherent limitations of causal attention and autoregressive decoding through bidirectional attention and parallel decoding, enabling efficient and high-quality image generation. However, compositional T2I generation remains challenging, as even state-of-the-art Diffusion Models often fail to accurately bind attributes and achieve proper text-image alignment. While Diffusion Models have been extensively studied for this issue, Masked Generative Transformers exhibit similar limitations but have not been explored in this context. To address this, we propose \textit{\ours{} (\oursabbr{})}, a novel training-free method that improves compositional fidelity by leveraging attention maps to prioritize the unmasking of tokens that clearly represent individual objects. \oursabbr{} consistently improves performance in both quantitative and qualitative evaluations across multiple benchmarks and metrics, with negligible inference overhead. Our code is available at \url{https://github.com/furiosa-ai/uncage}.%

\end{abstract}

\section{Introduction}
\label{sec:intro}

Text-to-image (T2I) generation~\citep{imagen,dalle} has emerged as a crucial task in computer vision, offering vast potential for creative applications. To achieve high-quality generation, various generative models, including GANs~\citep{gigagan}, Diffusion Models~\citep{sd}, and Autoregressive Models~\citep{llamagen,var}, have been extensively studied and rapidly advanced. Diffusion Models~\citep{ddpm,ddim} have significant advantages over GANs~\citep{gan} in terms of stable training and sample diversity. As a result, Diffusion Models have become the most widely adopted approach for visual generation tasks~\citep{svd,sv3d}, including T2I generation. However, with the recent success of autoregressive modeling in the NLP domain, exemplified by models like GPT~\citep{gpt3,gpt4} and LLaMA~\citep{llama,llama2}, there has been growing interest in applying Autoregressive Models to image generation as well~\citep{llamagen,lumina}.

Autoregressive Models generate image tokens sequentially, producing only one token per step, which makes inference computationally expensive. Moreover, the attention mechanism in these models is applied causally rather than bidirectionally, limiting the model’s ability to fully leverage contextual information during image generation.
Masked Generative Transformers (MGTs) address these limitations~\citep{maskgit,muse,meissonic}, offering an alternative generative paradigm that enhances efficiency through parallel decoding and leverages information more effectively through bidirectional attention.

As T2I generation advances, there is growing focus on compositional T2I generation, where the goal is to correctly render multiple objects and their attributes based on the prompt.
However, most T2I generation models struggle with misaligned attribute binding, resulting in discrepancies between the prompt and the generated image. For example, as shown in \cref{fig:teaser}, given the prompt \textit{“a turtle and a pink apple,”} the model may generate a turtle with a pink apple-shaped shell, instead of depicting both objects separately.

Attend-and-Excite~\citep{attend} is the first study to address this issue in Diffusion Models. \citet{attend} observe that attention maps in Diffusion Models correlate strongly with object regions, but attribute binding problems occur when these maps fail to accurately localize individual objects and overlap with each other.
To mitigate this, they introduce an attention-based loss and refine the output at each denoising step by computing gradients and updating the generation process accordingly.
Subsequent studies~\citep{syngen,predicated} build on Attend-and-Excite to further improve attribute binding in compositional T2I generation.
However, these methods rely on gradient-based iterative refinement during inference, resulting in substantial inference overhead.

MGTs, like most Diffusion Models, utilize attention-based architectures, which can lead to misaligned attribute binding due to inaccurate attention maps. However, no prior work has addressed this issue for MGTs. Moreover, Diffusion Models and MGTs generate outputs in fundamentally different ways depending on the timestep.
Diffusion Models progressively refine arbitrary regions of the image through iterative denoising, whereas MGTs predict all token positions in parallel at each timestep, and only a subset of these tokens is unmasked, which then remains fixed throughout the process.
Due to this fundamental difference, directly applying refinement strategies~\citep{attend,syngen,predicated} developed for Diffusion Models to MGTs is non-trivial. This highlights the need for methods specifically designed for MGTs.

\begin{figure*}[!htbp]
    \centering
    \includegraphics[width=\linewidth]{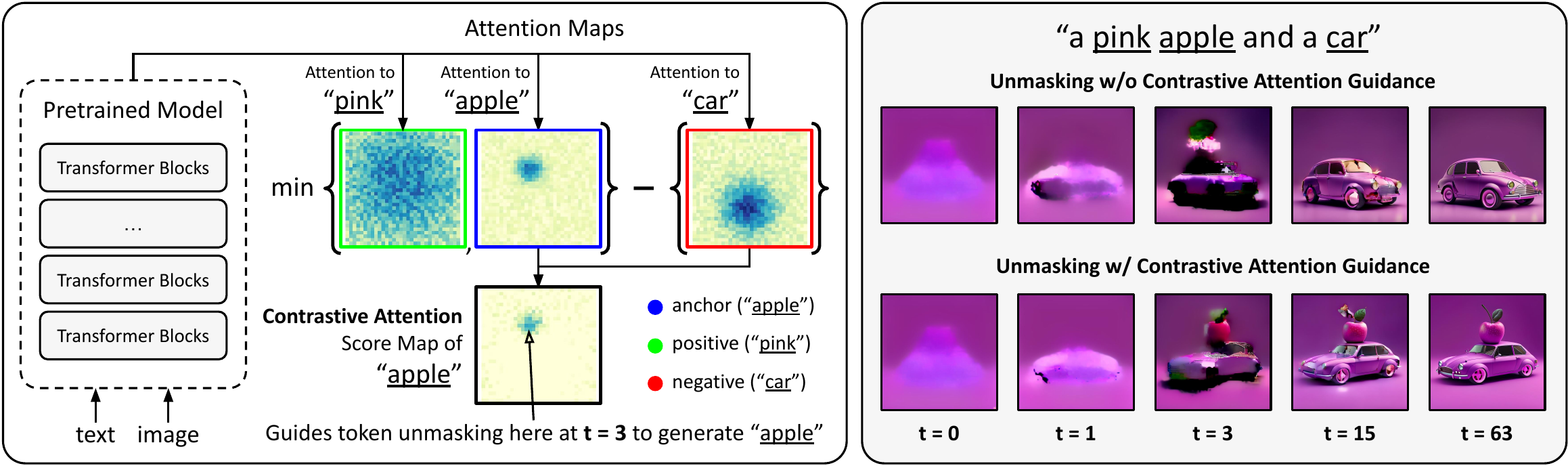}
    \caption{
    Overview of \textit{\ours{} (\oursabbr{})}. Without \oursabbr{}, as shown at $t = 3$, the model fails to select (unmask) the apple-related token despite generating a reasonable candidate, resulting in an image that contains only the car. In contrast, with \oursabbr{}, it effectively identifies tokens that clearly represent each object, allowing the model to select the apple-related token at $t = 3$ and successfully generate both the car and the apple in the final image.
    }
    \label{fig:method}
\end{figure*}

In MGTs, the overall structure is largely determined during the early generation steps, as shown in the examples on the right of \cref{fig:method}.
In the upper right example, the model failed to generate \textit{“a pink apple and a car”} and instead produced only a car.
We hypothesize that if the apple-related token predicted at $t = 3$ had been unmasked and retained, it could have served as a foundation for generating the apple in subsequent steps (the lower right example), thereby improving alignment.
This observation indicates that \textit{unmasking order strategies}, particularly during early steps, are crucial for improved text-image alignment.

Based on these insights, we propose \textit{\ours{} (\oursabbr{})} to address misaligned attribute binding in compositional T2I generation by leveraging attention maps to prioritize the unmasking of tokens that clearly represent individual objects. This process is training-free and incurs negligible additional computational cost, as it relies solely on attention maps that have already been computed to guide the unmasking order.

In summary, our main contributions are:

\begin{itemize}[leftmargin=*]
\item We are the first to address inaccurate attribute binding in compositional T2I generation with Masked Generative Transformers, improving text-image alignment.
\item We propose a novel training-free method, \textit{\ours{} (\oursabbr{})}, which mitigates this issue while incurring negligible additional cost.
\item We demonstrate the effectiveness of \oursabbr{} through quantitative and qualitative experiments across multiple benchmarks and evaluation metrics.
\end{itemize}

\section{Related Works}
\label{sec:related_works}
\subsection{Masked Generative Transformers (MGTs)}

While the success of GPT~\cite{gpt3,gpt4} has inspired autoregressive approaches for image generation~\citep{igpt}, MaskGIT~\citep{maskgit} takes a different approach, drawing inspiration from BERT~\citep{bert}. By leveraging a bidirectional Transformer and masked token prediction, MaskGIT enables parallel decoding, significantly accelerating inference while preserving high image quality. Its iterative refinement mechanism also makes it highly effective for tasks such as inpainting, outpainting, and local editing. Muse~\citep{muse} extends the masked generative approach of MaskGIT to T2I generation, leveraging a pre-trained language model for text conditioning. This method enables highly efficient parallel decoding while maintaining state-of-the-art image quality, outperforming both Autoregressive and Diffusion models in terms of speed and fidelity. %
Recently, Meissonic~\citep{meissonic} has incorporated novel architectural designs, enhanced positional encoding methods, and refined sampling techniques to achieve T2I generation performance comparable to state-of-the-art Diffusion Models like Stable Diffusion XL~\citep{sdxl}.

\paragraph{Unmasking Order of MGTs}
MGTs typically employ random order or confidence-based order for token unmasking~\cite{maskgit,meissonic}.
There are a few studies on the advanced unmasking (sampling) order of MGTs, but no research has yet focused on attribute binding in compositional T2I generation.
\citet{tcts-fas} propose Text-Conditioned Token Selection (TCTS) and Frequency Adaptive Sampling (FAS), which refine the sampling process to improve image quality. \citet{token-critic} introduce Token-Critic, an auxiliary Transformer that enhances MaskGIT’s sampling by identifying implausible tokens.
MaskSketch~\citep{masksketch} tackles the sketch-to-image synthesis task by extracting attention maps from pretrained masked image generation models when a sketch input is provided, and using them to prioritize the sampling of tokens with high structural similarity.
\citet{halton} introduce the Halton Scheduler, which improves token selection by leveraging a low-discrepancy sequence to enhance the image diversity and quality while ensuring a more uniform distribution of information across sampling steps.

While these methods improve image quality, most of them require additional training, and none of them address compositional T2I generation or the challenge of attribute binding. Unlike prior works, we focus on attribute binding in compositional T2I generation and propose an effective training-free method with negligible additional cost.

\subsection{Text-to-Image (T2I) Generation}
During the era of VAEs~\citep{vae} and GANs~\citep{gan}, research focused mainly on simpler image generation tasks, such as unconditional or class-to-image generation. However, with the success of Diffusion Models~\citep{ddpm}, which enable more stable and diverse generation, research on T2I generation has significantly advanced~\citep{sd,sd3}.
Meanwhile, inspired by the success of Autoregressive Models like GPT~\citep{gpt3} in language modeling, there has been a growing interest in applying them to T2I generation~\citep{llamagen}, demonstrating that autoregressive modeling is also effective in this domain.
Furthermore, research has recently begun to explore T2I generation by MGTs~\citep{meissonic}, which aims to address the limitations of autoregressive modeling.

\subsubsection{\textbf{Compositional T2I Generation}}
As T2I generation has advanced, focus has shifted to compositional T2I generation that renders multiple objects and their attributes based on the prompt.
In this setting, ensuring alignment between the prompt and the image is crucial, particularly because attributes of different objects can become entangled during compositional generation.
Attend-and-Excite~\citep{attend} incorporates an attention map-based loss, refining the output at each denoising step by computing gradients and adjusting the generation process accordingly. By leveraging syntactic structure, SynGen~\citep{syngen} adjusts cross-attention maps to better align with linguistic bindings, outperforming Attend-and-Excite. In addition, Predicated Diffusion~\citep{predicated} introduces a predicate logic-based attention guidance framework, leveraging fuzzy logic to enforce text-image alignment.

While these methods improve compositional T2I generation, they are all designed for Diffusion Models, and this problem has not been explored for MGTs. In addition, these methods rely on iterative refinement via gradient computation and incur substantial inference overhead, while we propose a training-free method with negligible inference cost.

\section{Preliminaries}

\paragraph{Unmasking Order of MGTs}

Unlike Autoregressive Models, MGTs predict all tokens in parallel at each timestep. However, only a subset of these predictions is selected (unmasked) for all subsequent timesteps, while the rest are discarded (re-masked) for future prediction. %
For MGTs in image generation, unmasking tokens in a purely random order likely leads to artifacts in the final image. To address this issue, most MGTs, including MaskGIT~\citep{maskgit} and Meissonic~\citep{meissonic}, use the logits as confidence scores to guide the unmasking order. Additionally, random noise is added for maintaining stochasticity.
In practice, after sampling each token, the logit value of each token is used as confidence scores $F_c(t)$, where $F_c(t)[i, j]$ denotes the logit value of the token at position $(i, j)$ at timestep $t$. Stochasticity is implemented via adding Gumbel noises $F_g(t)$, where the temperature of the Gumbel noise is linearly decreased from 1.0 to 0.01 across timesteps~\citep{meissonic}.
Finally, the unmasking order scores $F(t)$ are computed as the sum of $F_c(t)$ and $F_g(t)$. The top-$k$ tokens are unmasked based on their scores, where $k$ is determined by a cosine schedule:
\begin{equation}
F(t) = F_c(t) + F_g(t).
\end{equation}
Recently, ~\citet{halton} proposed Halton Scheduler that utilizes input-independent quasi-random Halton sequences for pre-determined unmasking patterns, reducing the correlation between tokens unmasked in the same step and maximizing information gain.

\paragraph{Challenges in Compositional T2I Generation}
As shown in \cref{fig:teaser}, due to misaligned attribute binding, existing unmasking order methods in MGTs, including Halton Scheduler, often fail at compositional T2I generation.
Similar issues occur in Diffusion Models, where inaccurate attention maps have been identified as a key cause of attribute mismatch, and addressed through attention-based losses and gradient-based refinement.
However, these methods are not directly applicable to MGTs. Unlike in Diffusion Models, once a token is unmasked during the generation in MGTs, it becomes fixed and cannot be refined further, as it is directly used in the final image. This highlights the importance of the unmasking strategy for compositional T2I generation.%

\section{Method}
In this section, we propose \textit{\ours{} (\oursabbr{})}, a novel training-free unmasking order method for compositional T2I generation, which is outlined in \cref{fig:method} and \cref{alg:algorithm1}.

\paragraph{\textit{Notation}}

From a given text prompt P (e.g., \textit{“an Attribute1 Object1 and an Attribute2 Object2”}), we extract a set of subject tokens $\mathcal{S}$. This set is composed of object tokens (forming the set $\mathcal{O}$) and attribute tokens (forming the set $\mathcal{A}$), such that $\mathcal{S} = \mathcal{O} \cup \mathcal{A}$.
For each subject token $s \in \mathcal{S}$, we define $M_t^s$ as the attention map of image tokens attending to token $s$ at timestep $t$.
For each object token $o \in \mathcal{O}$ (e.g., Object1), we define $\mathcal{N}_o$ as the set of subject tokens forming \textbf{\textit{negative pairs}} with $o$ (e.g., Attribute2 and Object2), and $\mathcal{P}_o$ as the set of subject tokens forming \textbf{\textit{positive pairs}} with $o$ (e.g., Attribute1 and Object1), where $\mathcal{P}_o$ also includes $o$ itself.

\subsection{Motivation}

 Suppose the image token at position $[i, j]$ is unmasked at timestep $t$. We want this token to clearly represent the individual object $o$ (e.g., Object1) with its corresponding attributes. For this, we identify two key constraints: i) for positive pairs ($p_o \in \mathcal{P}_o$), and ii) for negative pairs ($n_o \in \mathcal{N}_o$).

\paragraph{Positive Pair Constraint}
For position $[i,j]$ to distinctively represent object $o$ (e.g., Object1), not only object $o$ itself but also the other subjects that form positive pairs with object $o$ ($p_o \in \mathcal{P}_o$, e.g., Attribute1) should have high attention scores at the same position $[i,j]$.

For example, if $M_t^{o}[i, j]$ is low, the token may not contribute meaningfully to object $o$ (e.g., Object1) at timestep $t$, potentially resulting in a \textit{\textbf{missing object}} in the final image. If $M_t^{o}[i, j]$ is high but $M_t^{p_o}[i, j]$ is low, while object $o$ (e.g., Object1) may be generated successfully, its corresponding attributes (e.g., Attribute1) may fail to be properly represented, resulting in \textit{\textbf{attribute leakage}}. Therefore, the value of $\min_{p_o \in \mathcal{P}_o}(M_t^{p_{o}}[i,j])$ should be large enough.%

\paragraph{Negative Pair Constraint}
In contrast, for position $[i,j]$ to distinctively represent object $o$ (e.g., Object1), the attention scores of subjects that form negative pairs with object $o$ ($n_o \in \mathcal{N}_o$, e.g., Object2) should be low at position $[i,j]$.

For example, if both $M_t^{o}[i, j]$ and $M_t^{n_o}[i, j]$ are high, the image token is likely to encode attributes from both objects, increasing the risk of \textit{\textbf{object mixture}} when it is selected and preserved in the final image. 
Therefore, the value of $\max_{n_o \in \mathcal{N}_o}(M_t^{n_{o}}[i,j])$ should be small enough.%

\begin{algorithm}[tb]
\caption{\ours{} (\oursabbr)}
\label{alg:algorithm1}
{\raggedright
\textbf{Input}: prompt \text{P}, previously unmasked tokens $\texttt{prev\_tokens}$, current timestep $t$, Masked Generative Transformer $MGT$, a set of subject token indices $\mathcal{S}$\\
\textbf{Parameter}: guidance weight $w_a$\\
\textbf{Output}: unmasked tokens at timestep $t$\\
\par}
\begin{algorithmic}[1] %

\STATE \texttt{pred\_tokens}, $M_t \leftarrow$ $MGT(\text{P}, \texttt{prev\_tokens}, t)$%
\FOR{$s \in \mathcal{S}$}
    \STATE $M_t^s \gets M_t[:, :, s]$
    \STATE $M_t^s \gets \text{Gaussian}(M_t^s)$
\ENDFOR
\FOR{each position $[i, j]$ in the attention map}
    \STATE $F_a(t)[i,j] \gets \max_{o \in \mathcal{O}}(\min_{p_o \in \mathcal{P}_o}(M_t^{p_{o}}[i,j]) - \max_{n_o \in \mathcal{N}_o}(M_t^{n_{o}}[i,j]))$%
\ENDFOR
\STATE $F(t) \leftarrow F_c(t) + F_g(t) + w_a F_a(t)$
\STATE $\texttt{unmask\_pos} \leftarrow \operatorname*{arg\,top\textit{k}}(F(t), k =  \texttt{cos\_sched}(t))$
\STATE \textbf{return} $\text{unmask}(\texttt{unmask\_pos}, \texttt{pred\_tokens})$ %
\end{algorithmic}
\end{algorithm}

\subsection{Proposed Method: \oursabbr{}}

Building on this insight, we propose \textit{\ours{} (\oursabbr{})} to unmask tokens that clearly represent individual objects.
\cref{fig:method} shows the overview of \oursabbr{}.

\paragraph{Contrastive Attention Guidance}
For each object $o$, we construct two guidance signals: i) $\min_{p_o \in \mathcal{P}_o}(M_t^{p_{o}}[i,j])$ for positive pairs, and ii) $-\max_{n_o \in \mathcal{N}_o}(M_t^{n_{o}}[i,j])$ for negative pairs. Combining these, we compute the contrastive attention score $F_a^o(t)[i,j]$ to measure how distinctively position $[i,j]$ represents object $o$ as below:
\begin{equation}
\label{eq:individual_score}
\overbrace{F_a^o(t)[i,j]}^{\textbf{contrastive}} = \overbrace{\min_{p_o \in \mathcal{P}_o}(M_t^{p_{o}}[i,j])}^{\textbf{positive pair}} - \overbrace{\max_{n_o \in \mathcal{N}_o}(M_t^{n_{o}}[i,j])}^{\textbf{negative pair}}
\end{equation}
Then, the final contrastive attention score of each position $[i,j]$ corresponds to the most distinctive object's score, obtained by taking the maximum across all objects:
\begin{equation}
\label{eq:final_score}
F_a(t)[i,j] = \max_{o \in \mathcal{O}}F_a^o(t)[i,j].
\end{equation}
A high $F_a(t)[i,j]$ indicates that a token at position $[i,j]$ clearly represents an object with its corresponding attributes.
Note that \cref{eq:final_score} reduces to absolute difference as $F_a(t)[i,j] = |M_t^{Object1}[i, j]-M_t^{Object2}[i, j]|$ in the simple case, \textit{“an Object1 and an Object2."}
The method can also be generalized for use with more than two objects, with results shown in the experiment section.
\paragraph{Implementation of \oursabbr{}}

We first obtain attention maps $M_t$ at the given timestep $t$ (\cref{alg:algorithm1}, line 1), similar to Attend-and-Excite~\citep{attend}. Specifically, the attention maps from the single-modal Transformer blocks are averaged over all blocks and all attention heads, and then rescaled. We use the attention weights after softmax, with image tokens as queries and text tokens as keys. The intuition is that the image tokens will be semantically related to the text tokens in proportion to the amount of attention they pay to the text tokens. After extracting these attention maps, we smooth them using Gaussian blur with a standard deviation of 2.0 (\cref{alg:algorithm1}, lines 2–5).

Next, we compute the contrastive attention guidance $F_a$ using Eq.2 and Eq.3 at each spatial location $[i, j]$ (\cref{alg:algorithm1}, lines 6–8). Then, we add $F_a(t)$ to the baseline unmasking score $F_c(t) + F_g(t)$ to obtain the final unmasking score $F(t)$ (\cref{alg:algorithm1}, line 9):
\begin{equation}
F(t) = F_c(t) + F_g(t) + w_a F_a(t).
\end{equation}
The scores can be combined using weights, with $w_a$ denoting the scale of guidance. The original unmasking scheme corresponds to $w_a = 0.0$. Finally, we select the top-$k$ tokens with the highest $F(t)$ values for unmasking, and proceed to the next timestep (\cref{alg:algorithm1}, lines 10-11).

\begin{figure*}[t]
    \centering
    \setlength{\tabcolsep}{0.5pt}
    \renewcommand{\arraystretch}{0.3}
    \resizebox{1.0\linewidth}{!}{
    {\scriptsize
    \begin{tabular}{c c c @{\hspace{0.1cm}} c c @{\hspace{0.1cm}} c c @{\hspace{0.1cm}} c c }

        &
        \multicolumn{2}{c}{\textit{``a bird and a horse''}} &
        \multicolumn{2}{c}{\textit{``a cat and a frog''}} &
        \multicolumn{2}{c}{\textit{``a turtle and a pink balloon''}} &
        \multicolumn{2}{c}{\textit{``a red bench and a yellow clock''}} \\
\\
        {\raisebox{0.45in}{
        \multirow{2}{*}{\rotatebox{90}{\small Meissonic (baseline)}}}} &         \hspace{0.05cm}
        \includegraphics[width=0.1\textwidth]{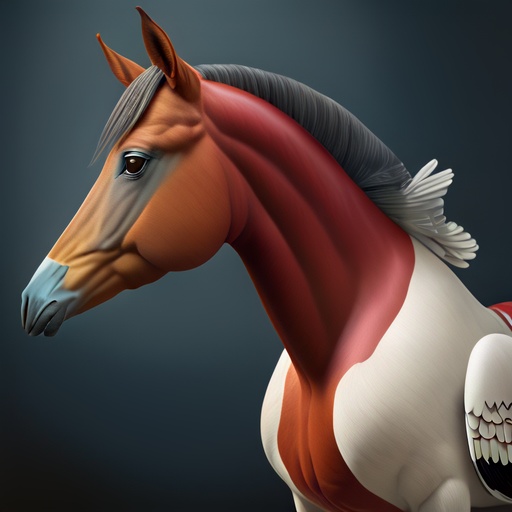} &
        \includegraphics[width=0.1\textwidth]{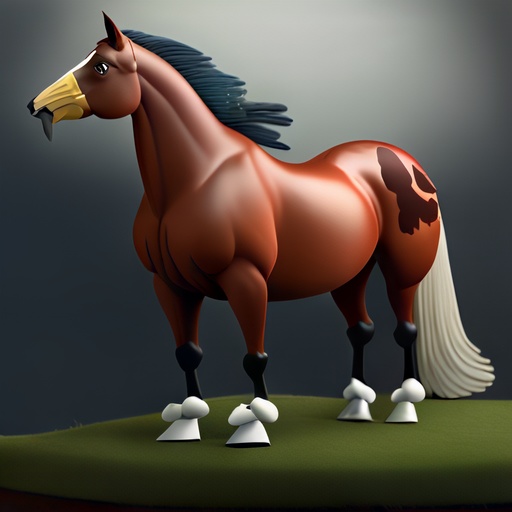} &
        \includegraphics[width=0.1\textwidth]{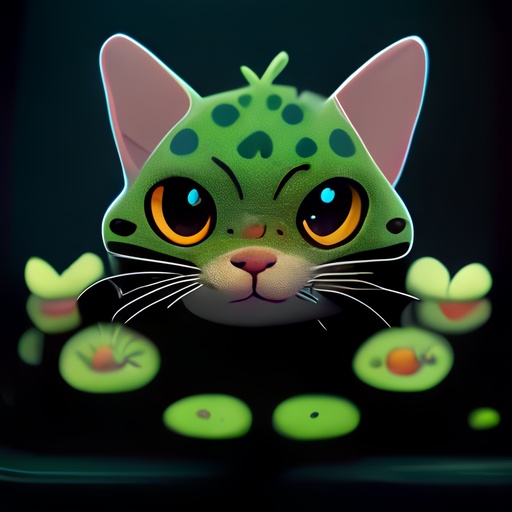} &
        \includegraphics[width=0.1\textwidth]{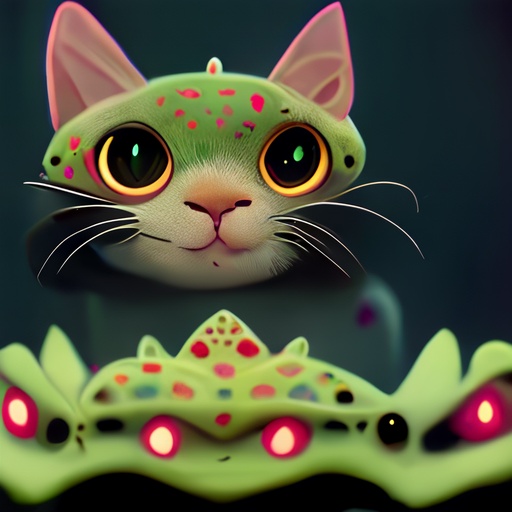} &
        \includegraphics[width=0.1\textwidth]{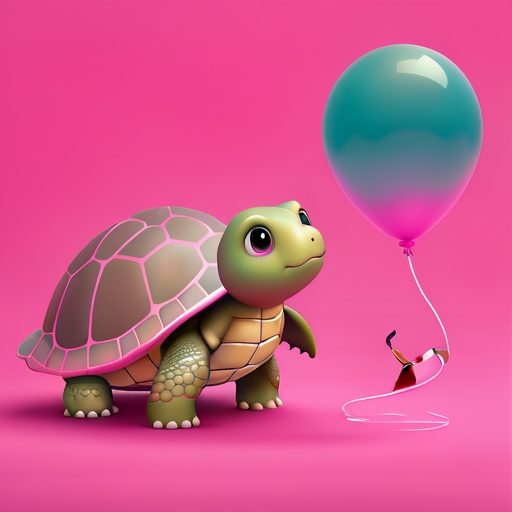} &
        \includegraphics[width=0.1\textwidth]{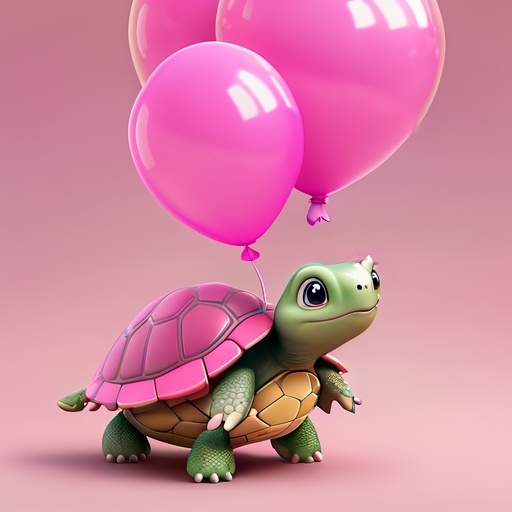} & 
        \includegraphics[width=0.1\textwidth]{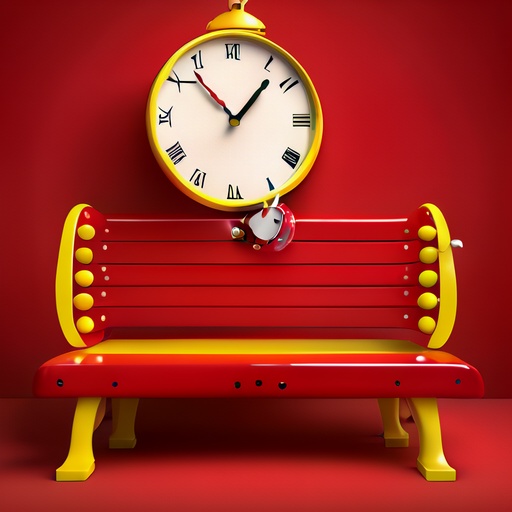} &
        \includegraphics[width=0.1\textwidth]{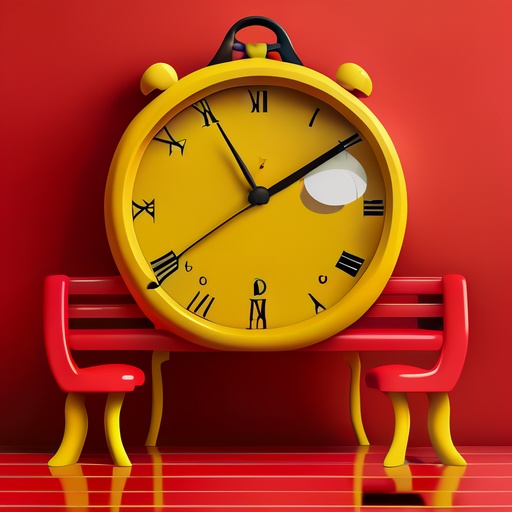} \\

        &
        \hspace{0.05cm}
        \includegraphics[width=0.1\textwidth]{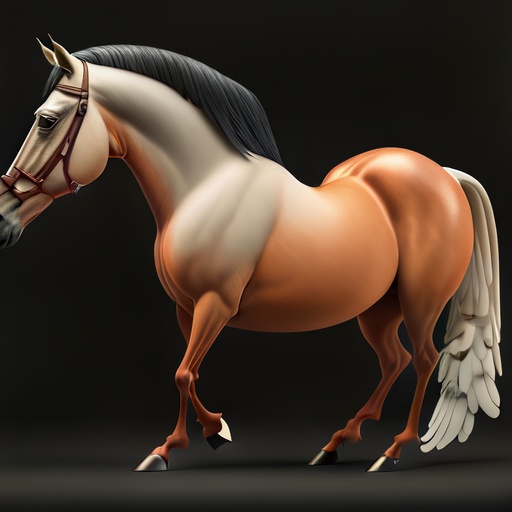} &
        \includegraphics[width=0.1\textwidth]{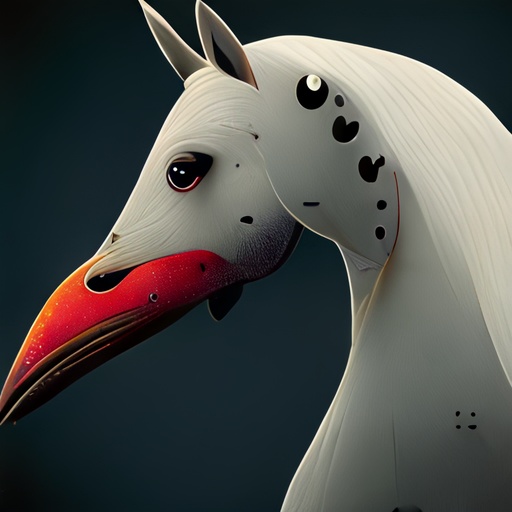} &
        \includegraphics[width=0.1\textwidth]{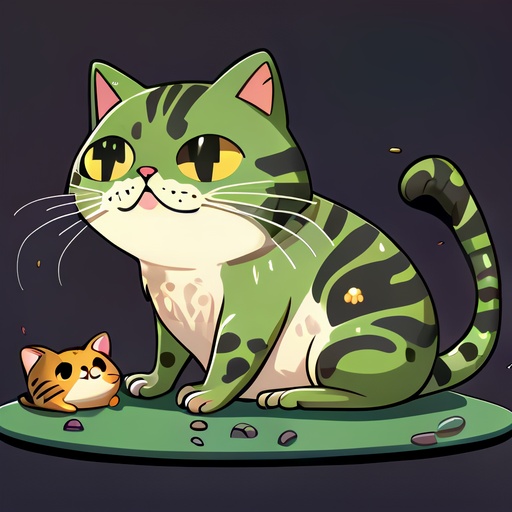} &
        \includegraphics[width=0.1\textwidth]{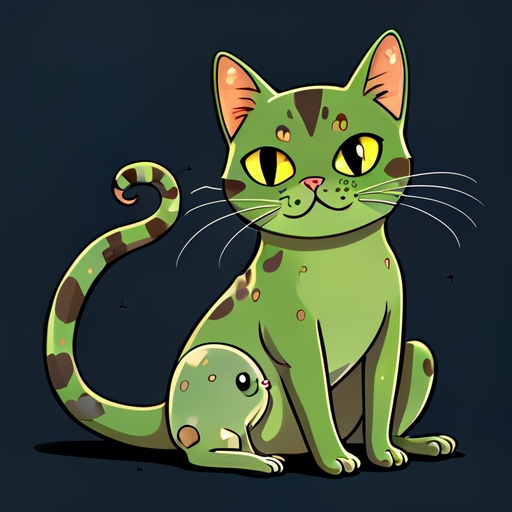} &
        \includegraphics[width=0.1\textwidth]{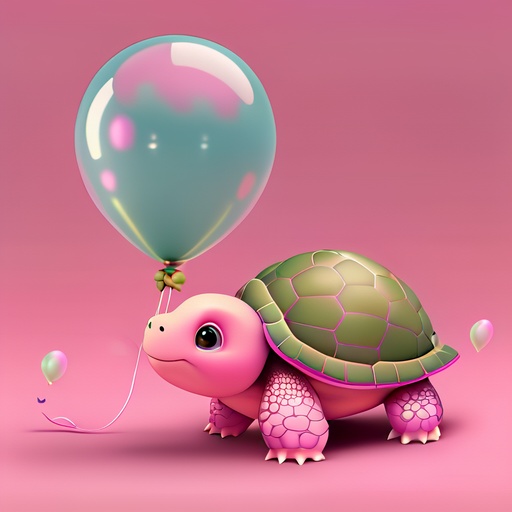} &
        \includegraphics[width=0.1\textwidth]{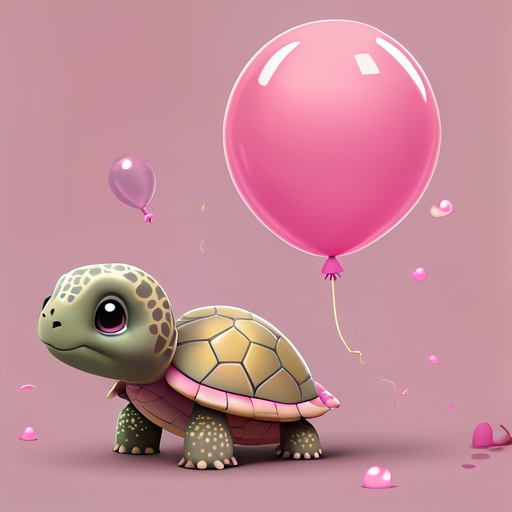} &
        \includegraphics[width=0.1\textwidth]{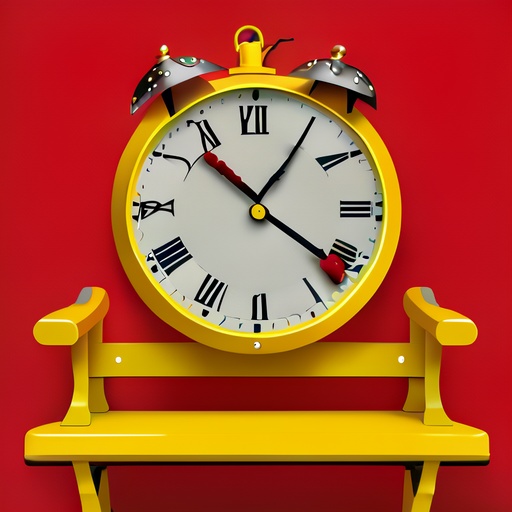} &
        \includegraphics[width=0.1\textwidth]{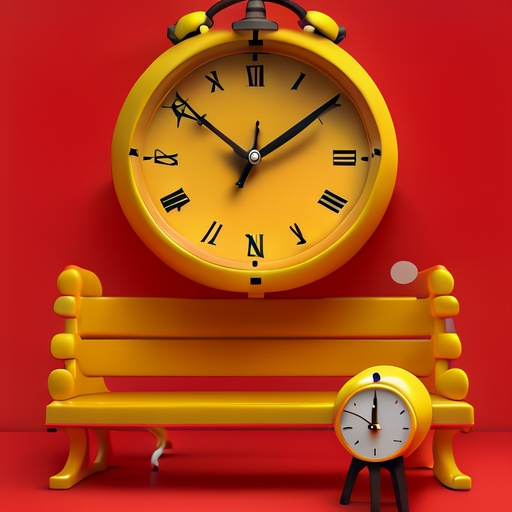} \\ \\

        {\raisebox{0.4in}{
        \multirow{2}{*}{\rotatebox{90}{\small Halton Scheduler}}}} &         \hspace{0.05cm}
        \includegraphics[width=0.1\textwidth]{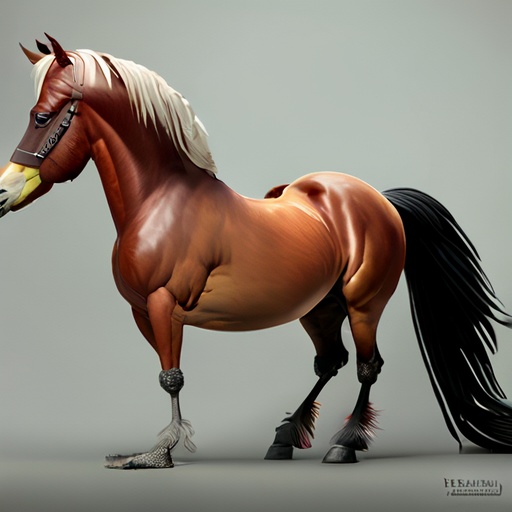} &
        \includegraphics[width=0.1\textwidth]{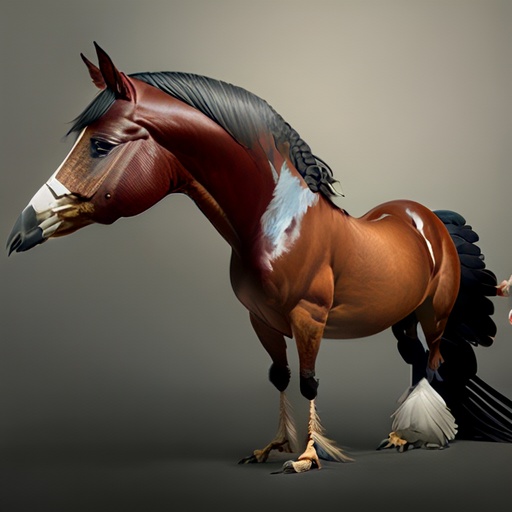} &
        \includegraphics[width=0.1\textwidth]{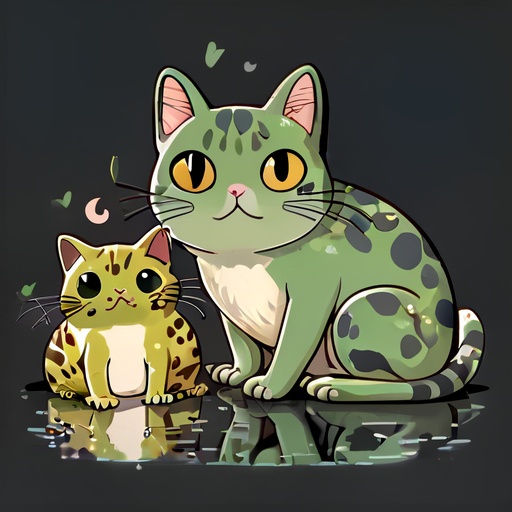} &
        \includegraphics[width=0.1\textwidth]{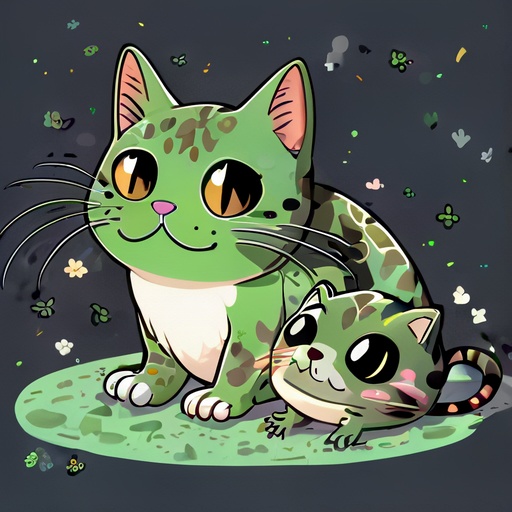} &
        \includegraphics[width=0.1\textwidth]{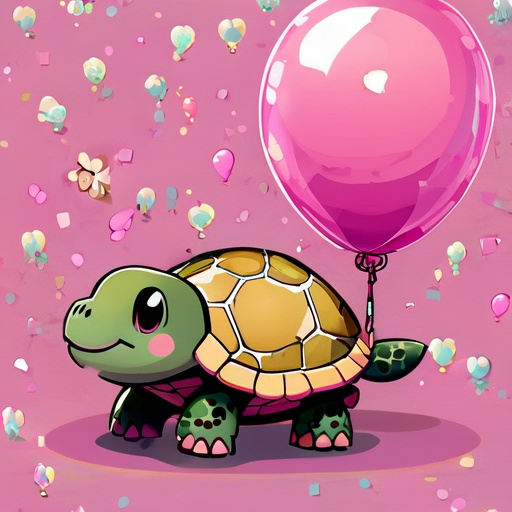} &
        \includegraphics[width=0.1\textwidth]{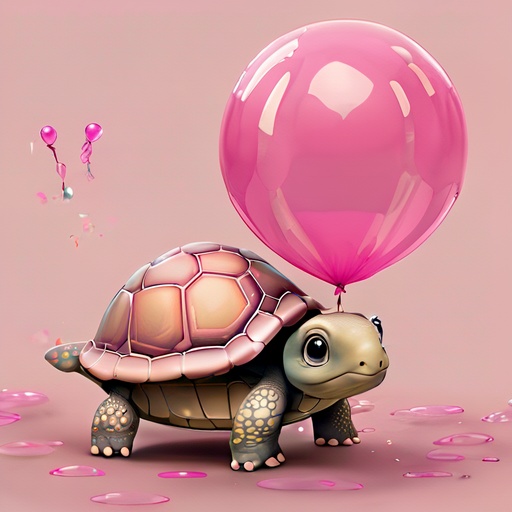} & 
        \includegraphics[width=0.1\textwidth]{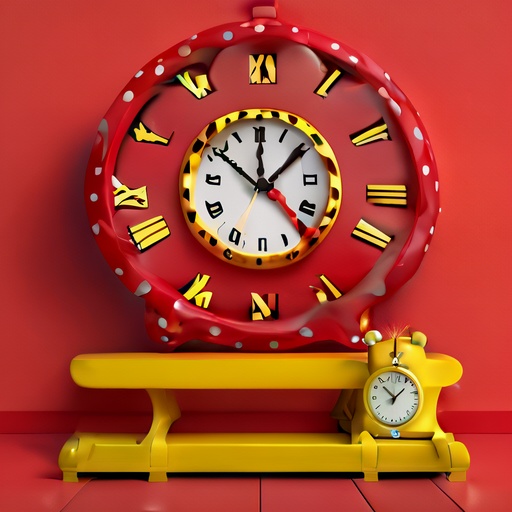} &
        \includegraphics[width=0.1\textwidth]{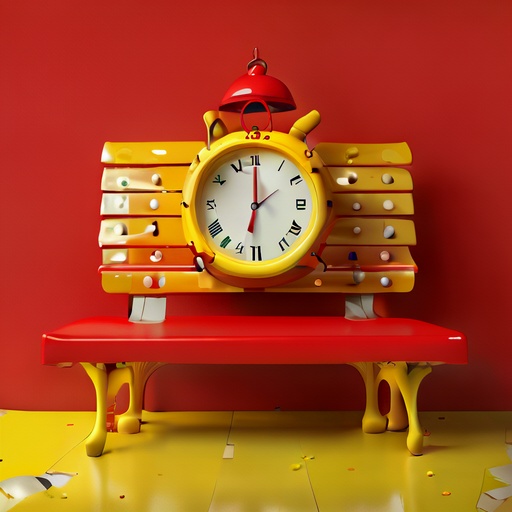} \\

        &
        \hspace{0.05cm}
        \includegraphics[width=0.1\textwidth]{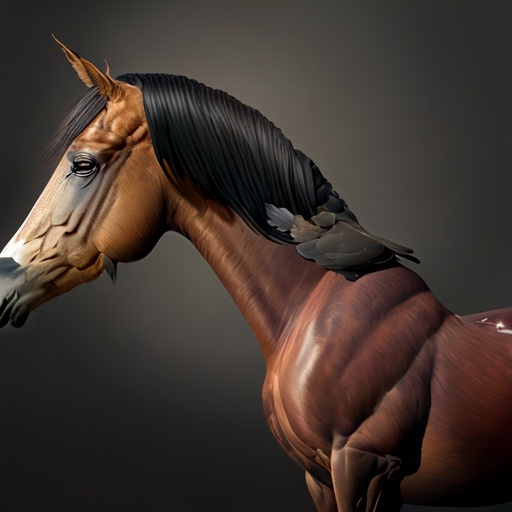} &
        \includegraphics[width=0.1\textwidth]{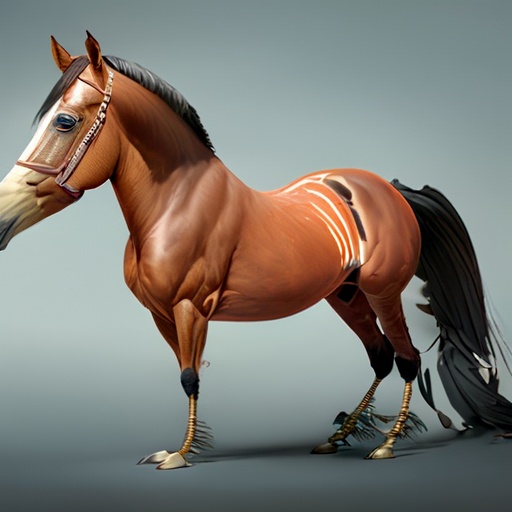} &
        \includegraphics[width=0.1\textwidth]{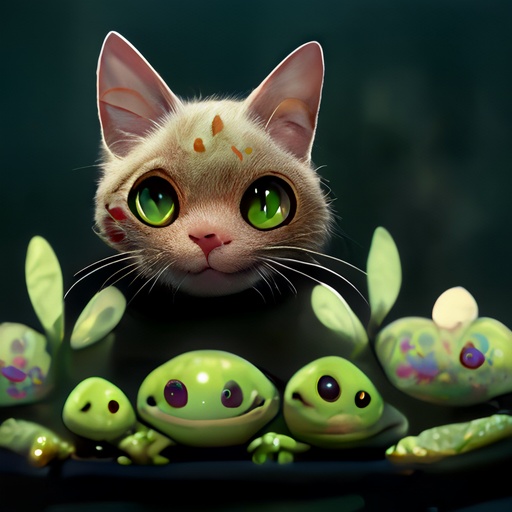} &
        \includegraphics[width=0.1\textwidth]{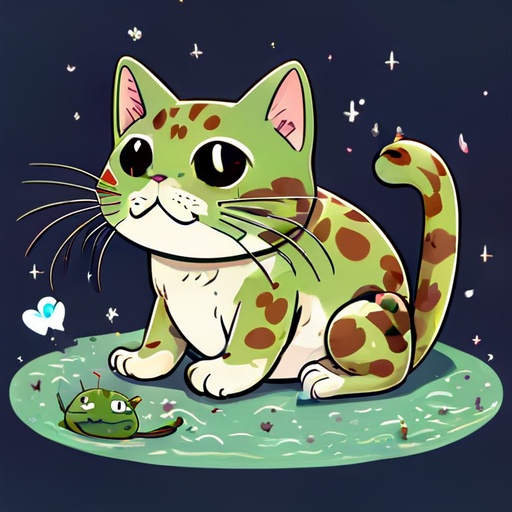} &
        \includegraphics[width=0.1\textwidth]{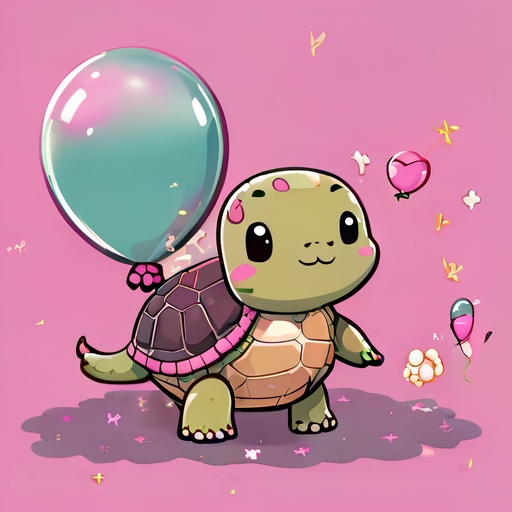} &
        \includegraphics[width=0.1\textwidth]{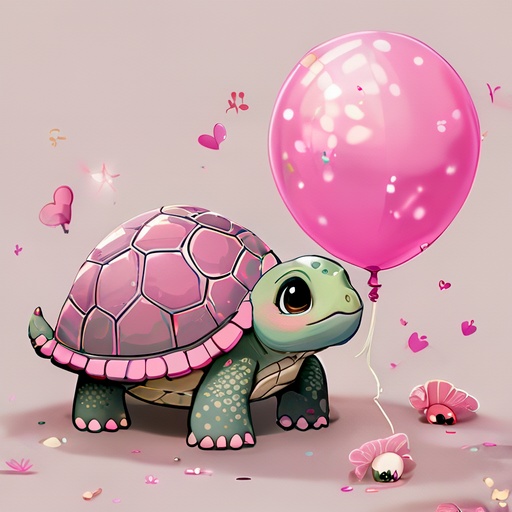} &
        \includegraphics[width=0.1\textwidth]{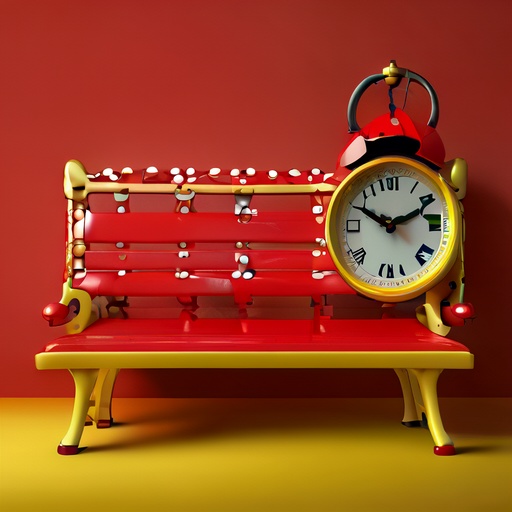} &
        \includegraphics[width=0.1\textwidth]{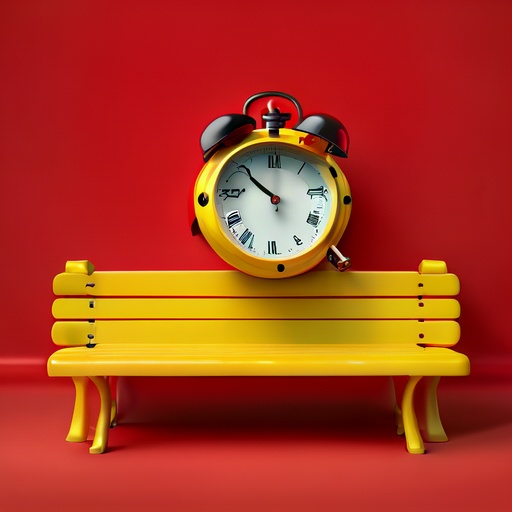} \\ \\

        {\raisebox{0.4in}{
        \multirow{2}{*}{\rotatebox{90}{\small \textbf{\oursabbr{} (ours)}}}}} &        \hspace{0.05cm}
        \includegraphics[width=0.1\textwidth]{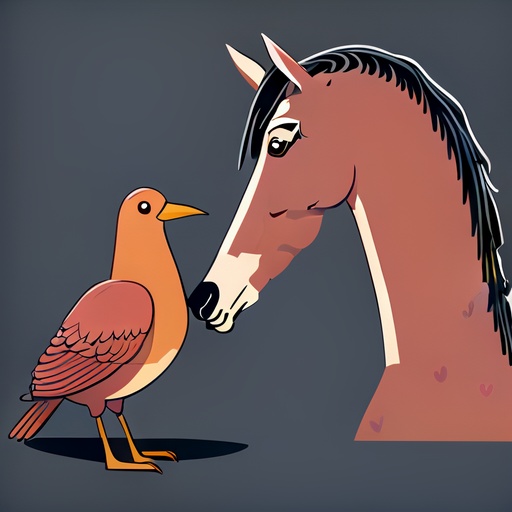} &
        \includegraphics[width=0.1\textwidth]{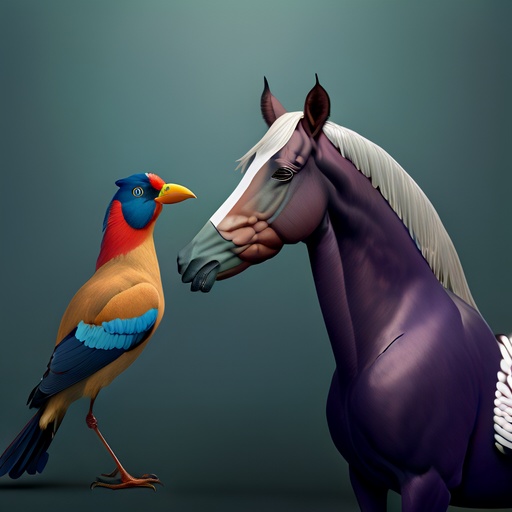} &
        \includegraphics[width=0.1\textwidth]{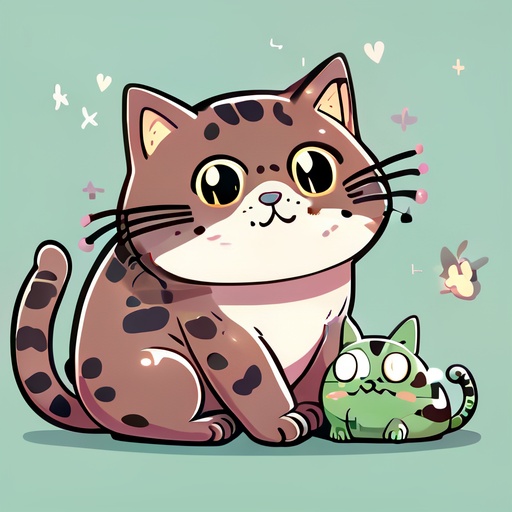} &
        \includegraphics[width=0.1\textwidth]{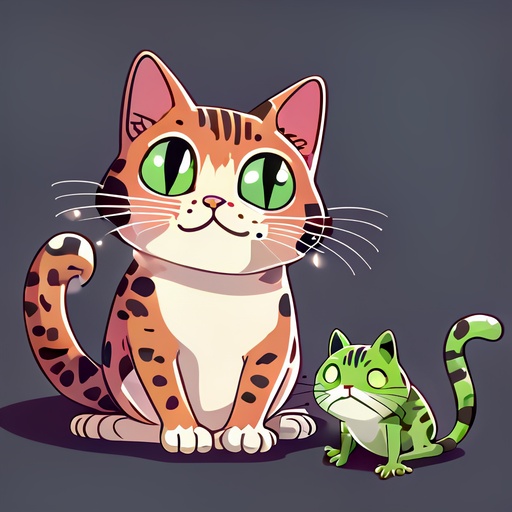} &
         \includegraphics[width=0.1\textwidth]{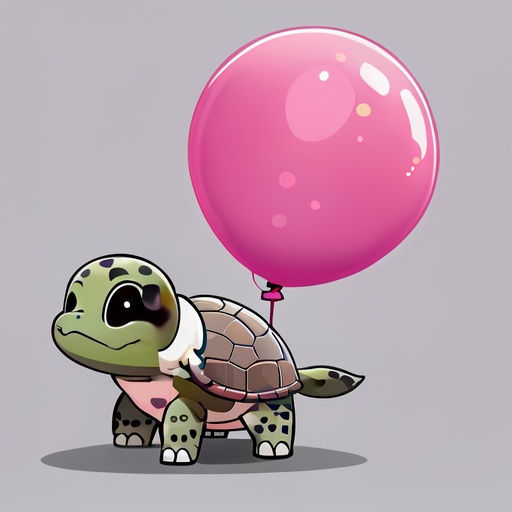} &
        \includegraphics[width=0.1\textwidth]{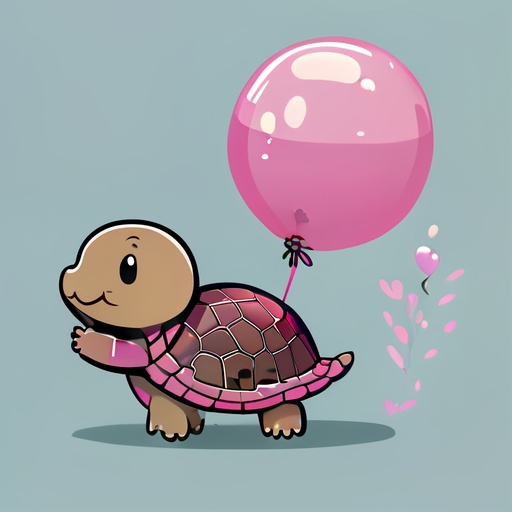} &
        \includegraphics[width=0.1\textwidth]{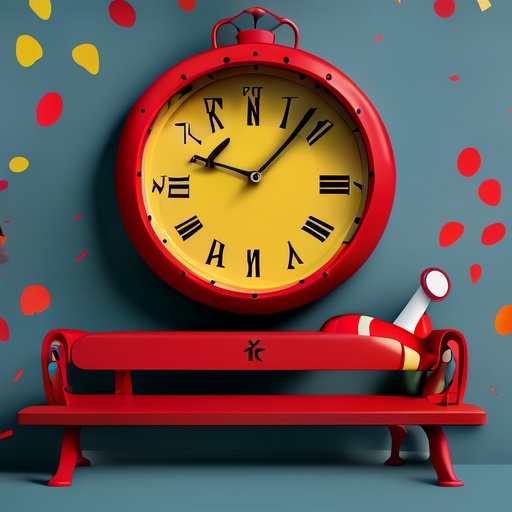} &
        \includegraphics[width=0.1\textwidth]{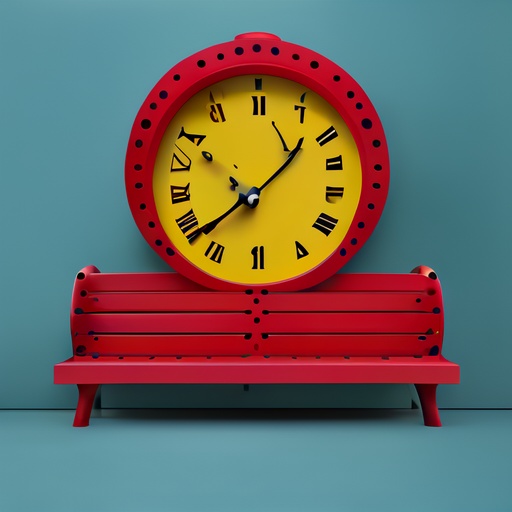} \\

        &
                \hspace{0.05cm}
        \includegraphics[width=0.1\textwidth]{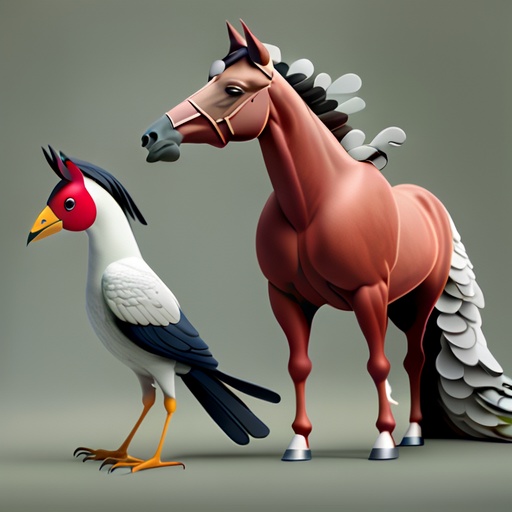} &
        \includegraphics[width=0.1\textwidth]{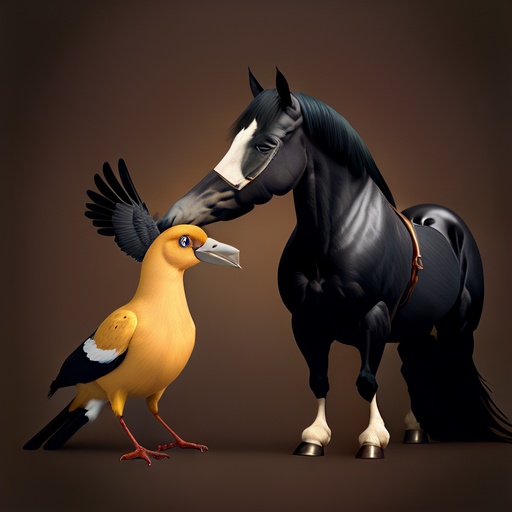} &
        \includegraphics[width=0.1\textwidth]{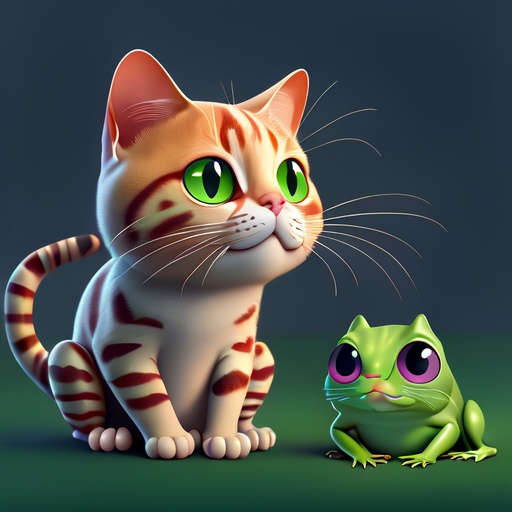} &
        \includegraphics[width=0.1\textwidth]{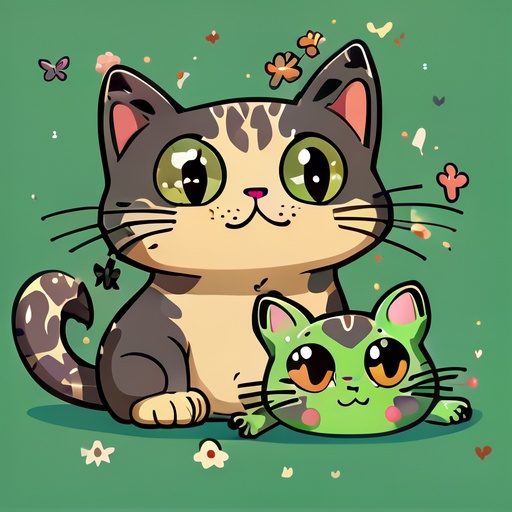} &
        \includegraphics[width=0.1\textwidth]{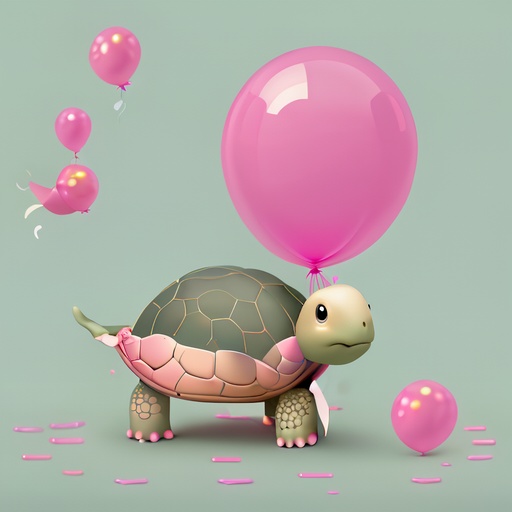} &
        \includegraphics[width=0.1\textwidth]{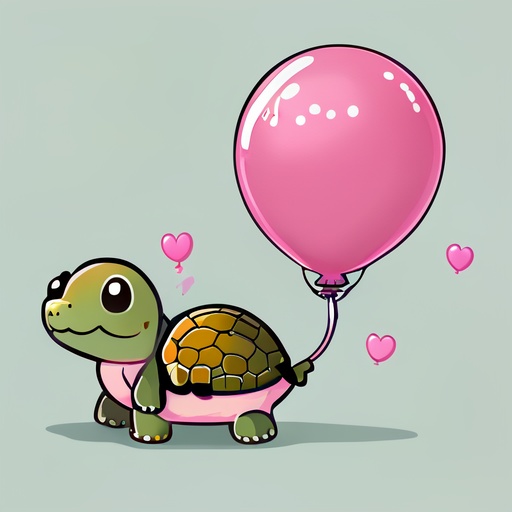} &
        \includegraphics[width=0.1\textwidth]{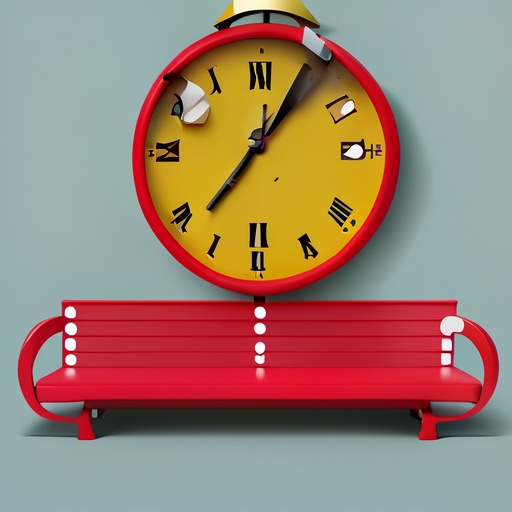} &
        \includegraphics[width=0.1\textwidth]{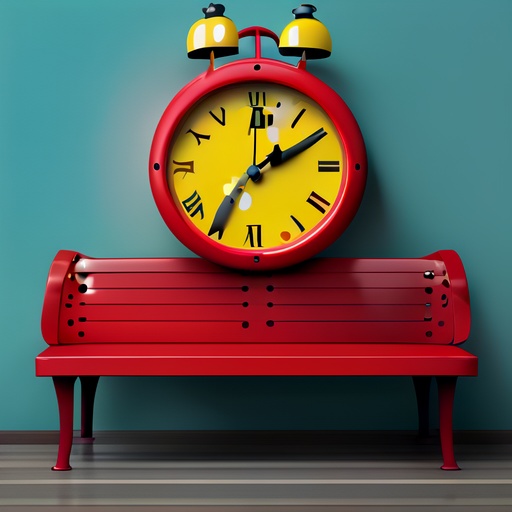} \\

    \end{tabular}
    }
    }
    \caption{
    Qualitative results on Attend-and-Excite dataset. The same seeds are applied to each prompt across all methods.
    }
    \label{fig:qualitative_results_1}
\end{figure*}

\section{Experiments}
\label{sec:experiments}

\begin{table}[ht!]
\centering
\caption{Quantitative results on the Attend-and-Excite and SSD datasets. \oursabbr{} is applied for the first 16 steps; \dag{} indicates all 64 steps. \textbf{Bold} and \underline{underlined} denote best and second-best results. \oursabbr{} consistently outperforms existing methods across benchmarks and evaluation metrics.} 
\label{tab:quantitative_results}
        \begin{subtable}[t]{\linewidth}
\caption{CLIP text-image similarities.} 
\label{tab:clip_score}
\resizebox{\linewidth}{!}{
\begin{tabular}{l c c c c c c} 
\toprule
\multirow{2.5}{*}{\textbf{Method / Dataset}}  &\multicolumn{3}{c}{\textbf{Attend-and-Excite}} & \multicolumn{2}{c}{\textbf{SSD}} & \multirow{2.5}{*}{\textbf{Avg.}}\\
\cmidrule(lr){2-4}\cmidrule(lr){5-6}
 & A.-A. & A.-O. & O.-O. & Two & Three\\
\midrule
Random ($F_g$) & 26.50 & 31.81 & 30.97 & 27.90 & 28.14 & \cellcolor{avg}29.06 \\
Confidence ($F_c$) & 28.37 & 35.33 & 36.69 & 29.34 & 29.64 & \cellcolor{avg}31.87 \\
Meissonic ($F_c + F_g$) & 29.82 & 36.15 & \underline{37.65} & 29.73 & 30.24 & \cellcolor{avg}32.72\\
Halton Scheduler & \textbf{30.81} & 36.15 & 37.52 & 30.01 & 30.46 & \cellcolor{avg}32.99\\
\oursabbr{} (n. only) & 30.43 & 36.11 & 37.43 & 30.08 & 30.52 & \cellcolor{avg}32.91\\
\oursabbr{} (p. only) & 30.22 & 36.05 & 37.45 & \underline{30.11} & \underline{30.53} & \cellcolor{avg}32.87\\
\textbf{\oursabbr{}\phantom{\dag{}} (ours)} & 30.43 & \underline{36.28} & \textbf{37.65} & 30.08 & 30.52 & \cellcolor{avg}\underline{32.99}\\
\textbf{\oursabbr{}\dag{} (ours)} & \underline{30.50} & \textbf{36.31} & 37.54 & \textbf{30.17} & \textbf{30.61} & \cellcolor{avg}\textbf{33.03} \\
\bottomrule
\end{tabular}
}
\end{subtable}

\medskip

\begin{subtable}[t]{\linewidth}
\caption{CLIP text-text similarities.} 
\label{tab:blip_score}
\resizebox{1.0\linewidth}{!}{
\begin{tabular}{l c c c c c c} 
\toprule
\multirow{2.5}{*}{\textbf{Method / Dataset}}  &\multicolumn{3}{c}{\textbf{Attend-and-Excite}} & \multicolumn{2}{c}{\textbf{SSD}} & \multirow{2.5}{*}{\textbf{Avg.}}\\
\cmidrule(lr){2-4}\cmidrule(lr){5-6}
 & A.-A. & A.-O. & O.-O. & Two & Three\\
\midrule
Random ($F_g$) & 66.70 & 76.98 & 69.76 & 64.03 & 65.76 & \cellcolor{avg}68.65\\
Confidence ($F_c$) & 73.12 & 85.14 & 84.88 & 67.71 & 71.04 & \cellcolor{avg}76.38 \\
Meissonic ($F_c + F_g$) & 73.98 & 87.05 & \underline{86.34} & 69.21 & 72.24 & \cellcolor{avg}77.76\\
Halton Scheduler & 76.47 & \textbf{87.27} & 86.10 & 69.85 & \underline{72.97} & \cellcolor{avg}78.53\\
\oursabbr{} (n. only) & \underline{77.05} & 87.00 & 86.33 & 69.94 & 72.66 & \cellcolor{avg}78.60\\
\oursabbr{} (p. only) & 76.36 & 86.53 & 85.55 & \underline{70.12} & 72.58 & \cellcolor{avg}78.23\\
\textbf{\oursabbr{}\phantom{\dag{}} (ours)} & \underline{77.05} & \underline{87.16} & \textbf{86.77} & 69.94 & 72.66 & \cellcolor{avg}\underline{78.72}\\
\textbf{\oursabbr{}\dag{} (ours)} & \textbf{77.21} & 86.92 & 85.88 & \textbf{70.44} & \textbf{73.50} & \cellcolor{avg}\textbf{78.79} \\
\bottomrule
\end{tabular}
}
\end{subtable}

\medskip

\begin{subtable}[t]{\linewidth}
\caption{GPT-based evaluation.} 
\label{tab:gpt_evaluation}
\resizebox{1.0\linewidth}{!}{
\begin{tabular}{l c c c c c c} 
\toprule
\multirow{2.5}{*}{\textbf{Method / Dataset}}  &\multicolumn{3}{c}{\textbf{Attend-and-Excite}} & \multicolumn{2}{c}{\textbf{SSD}} & \multirow{2.5}{*}{\textbf{Avg.}}\\
\cmidrule(lr){2-4}\cmidrule(lr){5-6}
 & A.-A. & A.-O. & O.-O. & Two & Three\\
\midrule
Random ($F_g$) & 5.94 & 8.52 & 6.57 & 5.65 & \textbf{5.59} & \cellcolor{avg}6.45 \\
Confidence ($F_c$) & 5.78 & 9.19 & 8.13 & 5.18 & 4.57 & \cellcolor{avg}6.57 \\
Meissonic ($F_c + F_g$) & 6.18 & \underline{9.65} & 8.52 &  5.59 & 5.03 & \cellcolor{avg}6.99 \\
Halton Scheduler & 6.65 & \textbf{9.67} & \textbf{8.88} & 5.79 & 5.35 & \cellcolor{avg}7.27\\
\oursabbr{} (n. only) & 6.80 & 9.65 & 8.66 & 6.16 & 5.37 & \cellcolor{avg}7.33 \\
\oursabbr{} (p. only) & \textbf{6.95} & 9.56 & 8.66 & \underline{6.20} & \underline{5.40} & \cellcolor{avg}\underline{7.35}\\
\textbf{\oursabbr{}\phantom{\dag{}} (ours)} & 6.80 & 9.62 & 8.74 & 6.16 & 5.37 & \cellcolor{avg}7.34\\
\textbf{\oursabbr{}\dag{} (ours)} & \underline{6.86} & 9.61 & \underline{8.81} & \textbf{6.27} & 5.36 & \cellcolor{avg}\textbf{7.38} \\
\bottomrule
\end{tabular}
}
\end{subtable}

\medskip

\begin{subtable}[t]{\linewidth}
\caption{User study results.} 
\label{tab:user_study}
\resizebox{1.0\linewidth}{!}{
\begin{tabular}{l c c c} 
\toprule
\multirow{2.5}{*}{\textbf{Method / Dataset}} & \multicolumn{1}{c}{\textbf{Attend-and-Excite}} & \multicolumn{1}{c}{\textbf{SSD}} & \multirow{2.5}{*}{\textbf{Avg.}}\\
\cmidrule(lr){2-2}\cmidrule(lr){3-3}
 & \multicolumn{1}{c}{Animal-Animal}  & \multicolumn{1}{c}{\phantom{..}Two Objects\phantom{..}}\\
\midrule
Tie & \multicolumn{1}{c}{18.2\%} & \multicolumn{1}{c}{\underline{31.7\%}} & \cellcolor{avg}25.0\% \\
Meissonic (baseline)    & \multicolumn{1}{c}{\underline{30.2\%}}    & \multicolumn{1}{c}{23.0\%} & \cellcolor{avg}\underline{26.6\%} \\
\textbf{\oursabbr{} (ours)}   & \multicolumn{1}{c}{\textbf{51.6\%}} & \multicolumn{1}{c}{\textbf{45.3\%}} & \cellcolor{avg}\textbf{48.4\%} \\
\bottomrule
\end{tabular}
}
\end{subtable}
\end{table}

\subsection{Experimental Setup}

For the pretrained MGT, we use Meissonic~\citep{meissonic}, a state-of-the-art MGT for T2I generation.
The model generates 1024×1024 images using parallel decoding with 64 timesteps and a cosine schedule.
For existing unmasking methods, we test i) random unmasking ($F_g$), ii) confidence-based unmasking ($F_c$), iii) Meissonic's approach ($F_c + F_g$), and iv) Halton Scheduler~\citep{halton}, a state-of-the-art unmasking method for image generation.

For \oursabbr{}, we test: i) using only negative pair guidance, ii) using only positive pair guidance, and iii) using both (our full contrastive guidance).
We use $w_a = 3$ as the default hyperparameter setting, and we also test the effect of varying $w_a$ in an ablation study.
As the overall structure of the final output in MGTs is largely determined during the early timesteps, we apply \oursabbr{} only during the first 16 steps and use the baseline ($F_c + F_g$) for the remaining 48 steps as our default setting. We also test applying \oursabbr{} for 64 (all) timesteps, denoted as \oursabbr{}\dag{}. A detailed ablation study on guidance timesteps is in the Appendix.

\paragraph{\textbf{Benchmark Datasets}}

We use two benchmark datasets: the Attend-and-Excite dataset~\citep{attend} and the SSD dataset~\citep{ssd}. The Attend-and-Excite dataset is categorized into three main groups: Animal-Animal, Animal-Object, and Object-Object. For instance, the Animal-Object category includes prompts such as \textit{“a mouse and a red car.”} This category is constructed using combinations of 12 animals and 12 objects, where each object is assigned one of 11 different colors. Each set contains 66, 144, and 66 prompts, respectively. The Similar Subjects dataset (SSD) is designed to evaluate compositional T2I generation in challenging settings. It consists of 31 prompts with two objects and 22 prompts with three objects involving semantically similar subjects, such as \textit{“A leopard and a tiger.”} For evaluation, we generate 4 images per prompt for the Attend-and-Excite dataset and 16 images per prompt for the SSD dataset using different random seeds. In total, 1,968 images are generated and used for evaluation.

\paragraph{\textbf{Evaluation Metrics}}

We measure CLIP~\citep{clip} text-image similarity and CLIP text-text similarity, following \citet{attend}.
CLIP text-image similarity is computed via cosine similarity between the CLIP embeddings of the generated image and the prompt.
Text-text similarity uses BLIP~\citep{blip} to convert the generated images to text captions, which are then compared with the prompt using CLIP.
However, as CLIP-based metrics are insufficient for accurately evaluating compositional T2I generation~\citep{attend,ssd}, we perform an additional GPT-based evaluation to  thoroughly validate the effectiveness of \oursabbr{}. 
Similar to prior works~\citep{viescore,etainv,counting}, we conduct GPT-based evaluation using the template in the Appendix.
Finally, we conduct a user study to compare the Meissonic (baseline) and \oursabbr{}. Ten respondents compared both methods for each prompt in the Attend-and-Excite (Animal-Animal) and SSD (Two Objects) datasets, evaluating 4 and 16 images per prompt, respectively. In total, each respondent evaluated 760 images and selected the better result for each prompt, or chose a tie.

\subsection{Quantitative Results}
\label{subsec:quantitative_results}

\cref{tab:quantitative_results} presents the quantitative results. CLIP text-image similarity, CLIP text-text similarity, and GPT-based evaluation exhibit consistent trends, showing general alignment across different settings.

First, Meissonic (baseline, $F_c + F_g$) performs poorly on the Animal-Animal and the SSD dataset, but excels on the Animal-Object and Object-Object categories.
This is likely because, in the Animal-Object and Object-Object categories, the objects are semantically distinct, making object mixture less likely. In contrast, in the Animal-Animal category and the SSD dataset, object mixture is more likely due to higher semantic similarity between entities.

For \oursabbr{}, using only negative guidance (n. only) or positive guidance (p. only) already outperforms Meissonic across all metrics in terms of average scores.
Moreover, when using both (contrastive), \oursabbr{} overall outperforms Halton Scheduler.
This demonstrates that both our proposed negative pair guidance and positive pair guidance work effectively, and combining them into contrastive guidance makes the method even more powerful.
The gain is especially significant on the Animal-Animal dataset and the SSD dataset, where the baseline performs poorly.
In contrast, for the Animal-Object and Object-Object datasets, where the baseline already performs well, the benefit of \oursabbr{} is relatively modest.
In summary, \oursabbr{} achieves state-of-the-art performance across multiple benchmarks and metrics compared to Meissonic and Halton Scheduler, with especially greater gain on challenging tasks. 

Notably, applying \oursabbr{} for all 64 timesteps (\oursabbr{}\dag{}) yields even greater gains and the best average scores across all metrics.
Nevertheless, as \oursabbr{}\dag{} exhibits dataset-dependent variance in its effectiveness, we default to applying \oursabbr{} for the first 16 steps to ensure more uniform performance gains across all datasets.

\paragraph{User Study Results}
As shown in \cref{tab:user_study}, \oursabbr{} was preferred nearly twice as often as the baseline. This suggests that \oursabbr{} is effective not only in improving quantitative metrics but also in better aligning with human preferences.

\subsection{Qualitative Results}
\label{subsec:qualitative_results}

\cref{fig:qualitative_results_1} presents the qualitative results on the Attend-and-Excite dataset, demonstrating that \oursabbr{} generally produces better results compared to Meissonic and Halton Scheduler.
In the case of \textit{“a bird and a horse,”} Meissonic and Halton Scheduler fail to generate both animals independently. Instead, they either generate only a horse or produce a fused entity where the bird and the horse are not clearly distinguishable.
In contrast, \oursabbr{} effectively separates and accurately generates both the bird and the horse.
For \textit{“a turtle and a pink balloon,”} Meissonic, Halton Scheduler, and \oursabbr{} all successfully generate the turtle and the balloon as separate objects.
However, while the generated balloon by \oursabbr{} is correctly pink in all cases, other methods often produce a balloon in a different color instead of pink.

    \begin{figure}[!tp]
\centering
    \resizebox{1.0\linewidth}{!}{
\begin{tikzpicture}[scale=0.88]
\begin{axis}[
    width=5cm,
    height=4cm,
    xlabel={guidance weight $w_a$},
    xlabel style={yshift=0.2cm},
    ylabel={similarity score},
    ylabel style={yshift=-0.2cm},
    xmin=0, xmax=5,
    ymin=29.7, ymax=30.9,
    xtick={0,1,2,3,4,5},
    ytick={30.0,30.5},
    yticklabel style={
        /pgf/number format/fixed,
        /pgf/number format/precision=1,
        /pgf/number format/zerofill
    },
    legend pos=north east,
    ymajorgrids=true,
    grid style=dashed,
    every axis plot/.append style={thick},
]
\addplot[
    color=red,
    mark=square,
    ]
    coordinates {
    (0.0, 29.82)
    (0.5, 29.93)
    (1.0, 30.24)
    (1.5, 30.14)
    (2.0, 30.40)
    (2.5, 30.45)
    (3.0, 30.43)
    (3.5, 30.32)
    (4.0, 30.46)
    (4.5, 30.20)
    (5.0, 30.25)
    };
    \addlegendentry{CLIP (txt-img)}
\addplot[mark=none, dashed, red, domain=0:5]{29.82};
\end{axis}
\end{tikzpicture}
\begin{tikzpicture}[scale=0.88]
\begin{axis}[
    width=5cm,
    height=4cm,
    xlabel={guidance weight $w_a$},
    xlabel style={yshift=0.2cm},
    xmin=0, xmax=5,
    ymin=73.5, ymax=79,
    xtick={0,1,2,3,4,5},
    ytick={74,76,78},
    yticklabel style={
        /pgf/number format/fixed,
        /pgf/number format/precision=1,
        /pgf/number format/zerofill
    },
    legend pos=north east,
    ymajorgrids=true,
    grid style=dashed,
    every axis plot/.append style={thick},
]
\addplot[
    color=blue,
    mark=square,
    ]
    coordinates {
    (0.0, 73.98)
    (0.5, 75.56)
    (1.0, 75.55)
    (1.5, 75.76)
    (2.0, 76.67)
    (2.5, 76.42)
    (3.0, 77.05)
    (3.5, 76.61)
    (4.0, 76.49)
    (4.5, 76.65)
    (5.0, 76.51)
    };
    \addlegendentry{CLIP (txt-txt)}
\addplot[mark=none, dashed, blue, domain=0:5]{73.98};
\end{axis}
\end{tikzpicture}
}
\caption{Ablation results for contrastive attention guidance weight on the Attend-and-Excite (Animal-Animal) dataset.}
\label{clip_plot}
\end{figure}
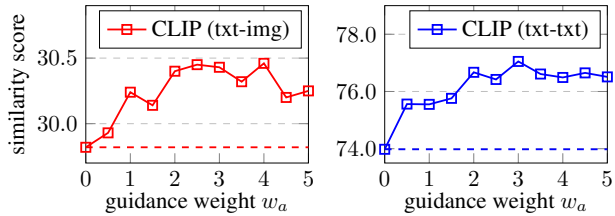
    \begin{figure}[tp]
    \centering
    \setlength{\tabcolsep}{0.5pt}
    \renewcommand{\arraystretch}{0.3}
    \resizebox{1.0\linewidth}{!}{
    {\small
    \begin{tabular}{c c c @{\hspace{0.1cm}} c c }

        &
        \multicolumn{2}{c}{\textit{“a lion and a frog”}} &
        \multicolumn{2}{c}{\textit{“a dog and a black apple”}} \\
\\
        {\raisebox{0.4in}{
        \multirow{2}{*}{\rotatebox{0}{\shortstack[c]{Meissonic \\ {(baseline)}}}}}} &         \hspace{0.1cm}
        \includegraphics[width=0.1\textwidth]{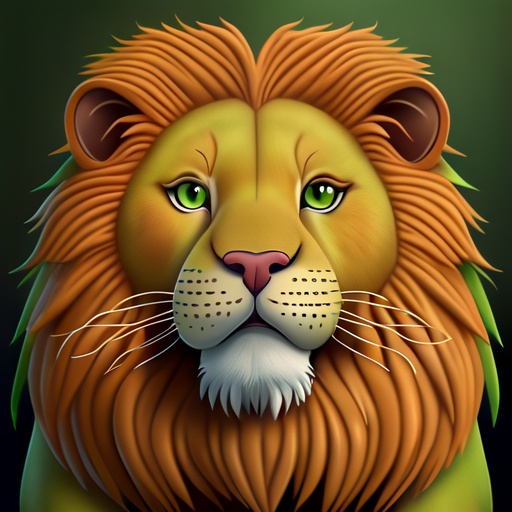} &
        \includegraphics[width=0.1\textwidth]{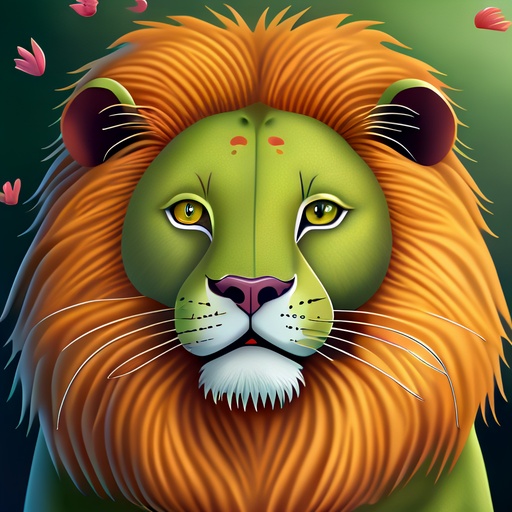} &
        \includegraphics[width=0.1\textwidth]{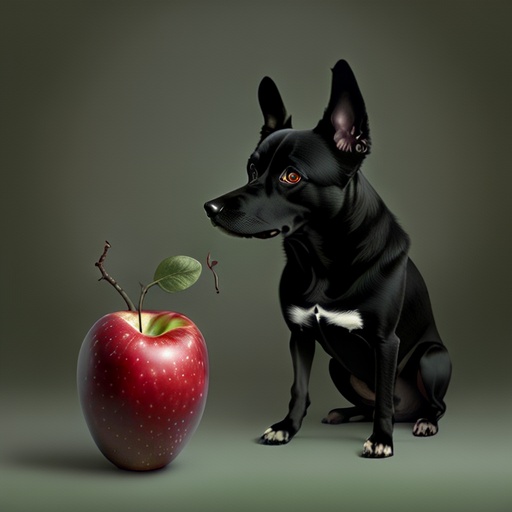} &
        \includegraphics[width=0.1\textwidth]{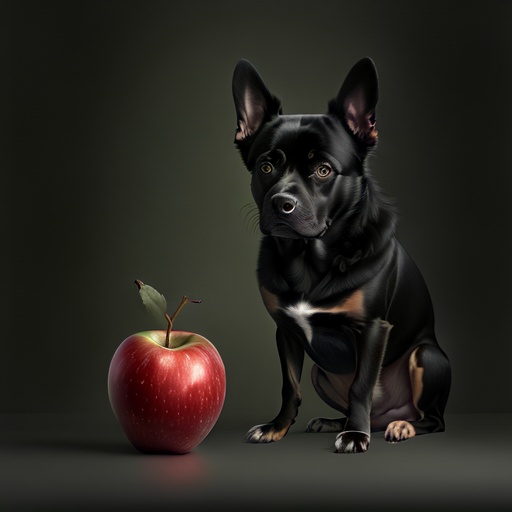} \\ \\

        {\raisebox{0.4in}{
        \multirow{2}{*}{\rotatebox{0}{\shortstack[c]{\textbf{\oursabbr} \\ {(ours)}}}}}} &        \hspace{0.1cm}
        \includegraphics[width=0.1\textwidth]{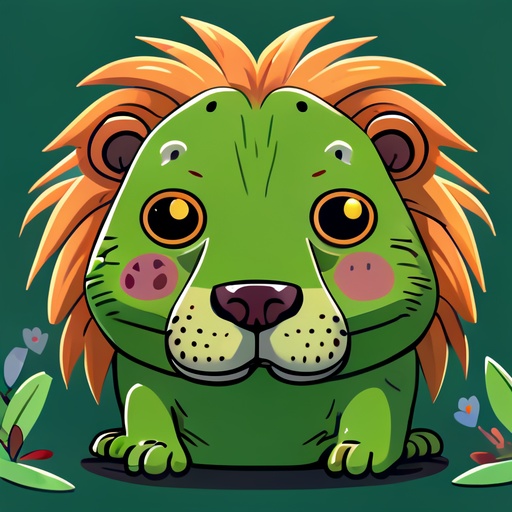} &
        \includegraphics[width=0.1\textwidth]{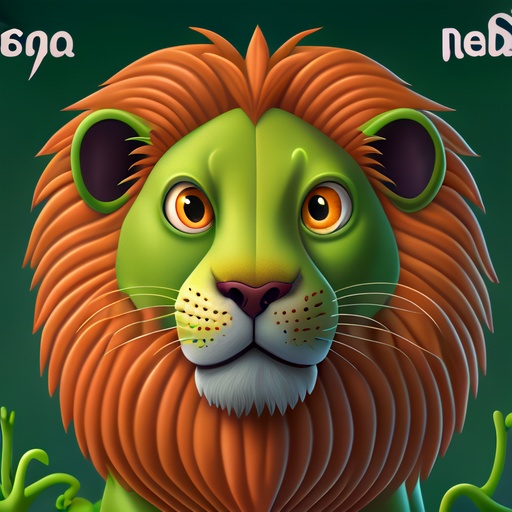} &
         \includegraphics[width=0.1\textwidth]{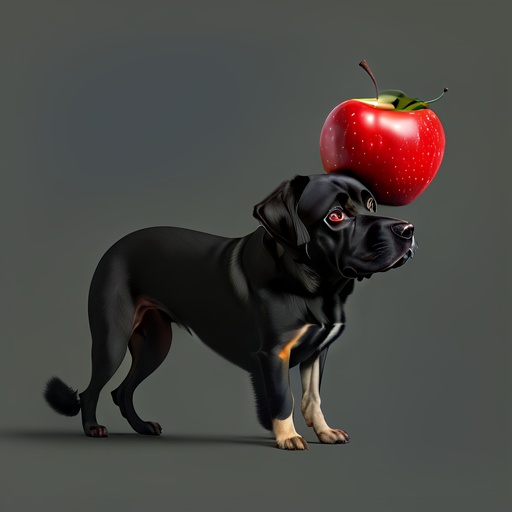} &
        \includegraphics[width=0.1\textwidth]{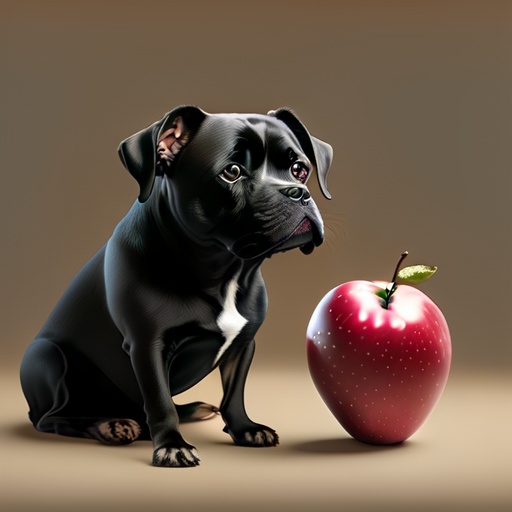} 
        \\

    \end{tabular}
    }
    }
\caption{Cases without improvement.}%
\label{fig:failure}
\end{figure}

\subsection{Ablation Study}
\label{subsec:ablation}

Additional ablation studies, including the number of guidance timesteps, Gaussian smoothing of the attention map, $F_c$ and $F_g$ terms, are provided in the Appendix.

\paragraph{Ablation on Guidance Weight $w_a$}

To study the effect of $w_a$, we conduct an ablation study on the Animal-Animal dataset by varying $w_a$ from 0 to 5 in increments of 0.5. As shown in \cref{clip_plot}, increasing $w_a$ from 0 gradually improves both CLIP text-image and text-text similarity scores, until around $w_a = 3$, after which performance begins to saturate or slightly decline.
These upward-concave trends suggest that \oursabbr{} is effective and functioning as intended, a behavior commonly observed in other guidance methods such as classifier-free guidance~\citep{cfg}.

\subsection{Inference Overhead Analysis}
\label{subsec:latency}

To demonstrate that \oursabbr{} introduces negligible inference overhead, we measure the runtime on an A100 GPU. Meissonic takes 16 seconds to generate an image over 64 steps. \oursabbr{} adds 0.0013 seconds per step for the first 16 steps, resulting in 0.0208 seconds of additional latency, which corresponds to 0.13\% of the total runtime. In contrast, \citet{attend} report that Attend-and-Excite increases Stable Diffusion’s inference time on an A100 from 5.6 seconds to 9.7–15.4 seconds, roughly doubling the runtime.

\section{Limitations and Future Work}

First, while \oursabbr{} improves compositional T2I generation, it does not always yield better results. Since \oursabbr{} is training-free and modifies only the unmasking order without changing the pretrained model, its effectiveness can be limited.
For example, in \cref{fig:failure}, the prompt \textit{“a dog and a black apple”} results in \textit{“a black dog and an apple”} instead.
This likely reflects pretrained bias, where black dogs are common and apples are typically red.

Second, while \oursabbr{} outperforms existing methods in MGTs with negligible inference overhead, the gain is relatively modest compared to guidance methods in Diffusion Models~\citep{attend}.
This is primarily due to the fundamental differences between \oursabbr{} and the methods commonly used in Diffusion Models. In Diffusion Models, guidance is explicitly provided through iterative refinement via gradient computation, which leads to larger gains at the cost of significantly increased inference time.
As future work, it would be interesting to explore methods that achieve greater gains in MGTs, even at higher cost.%

\section{Conclusion}
While interest in T2I generation using Masked Generative Transformers (MGTs) has recently increased, these often struggle to accurately bind attributes and achieve proper text-image alignment in compositional T2I generation. To address this, we have proposed \ours{} (\oursabbr{}), which is, to our knowledge, the first method specifically designed for compositional T2I generation with MGTs. By leveraging attention maps to prioritize tokens that clearly represent individual objects and guide the unmasking order accordingly, \oursabbr{} improves compositional fidelity in a training-free manner.
\oursabbr{} demonstrates consistent improvements via
quantitative and qualitative experiments across multiple benchmarks and metrics, with negligible inference overhead.

\bibliography{aaai2026}

\clearpage

\section{Details of GPT-based Evaluation}

While CLIP-based metrics are straightforward and widely used, they have limitations in accurately evaluating compositional T2I generation~\citep{attend,ssd}. Therefore, we conduct an additional GPT-based evaluation to obtain more accurate and fine-grained assessments, and to further validate the effectiveness of \oursabbr{}.
In this section, we describe the evaluation template used and provide sample responses generated by GPT during the evaluation.

\subsection{GPT-based Evaluation Template}
We use the template in \cref{lst:template} for GPT-based evaluation.
It is adapted from the prompt proposed by \citet{viescore}, with modifications tailored to our task.
We use the model \texttt{gpt-4o-2024-11-20} provided by the OpenAI API.

\begin{lstlisting}[caption=GPT-based evaluation template.,basicstyle=\ttfamily,breaklines=true,frame=single,breakindent=0pt,numbers=none,xleftmargin=0pt, framexleftmargin=0pt,label=lst:template]
You are a professional digital artist. You will have to evaluate the effectiveness of the AI-generated image(s) based on the given rules. You will have to give your output in this way (Keep your reasoning concise and short.):
{
\"score\" : [...],
\"reasoning\" : \"...\"
}
RULES:
Prompt and image will be provided:
The objective is to evaluate how the image is well aligned to the prompt.
On scale of 0 to 10:
A score from 0 to 10 will be given based on the success of the alignment. (0 indicates that the image does not follow the prompt at all. 10 indicates that the image in the image follows the prompt perfectly.)
Put the score in a list such that output score = [score]
\end{lstlisting}

\subsection{Example GPT-based Evaluation Outputs}

\begin{figure}[!htp]

\centering
\begin{minipage}{\linewidth}
\centering
\textit{\textbf{\large “a monkey and a frog”}}
\end{minipage}

\medskip

\subfloat[Output from Meissonic (baseline), failing to generate distinct objects.\label{sample1}]{
\includegraphics[width=.45\linewidth]{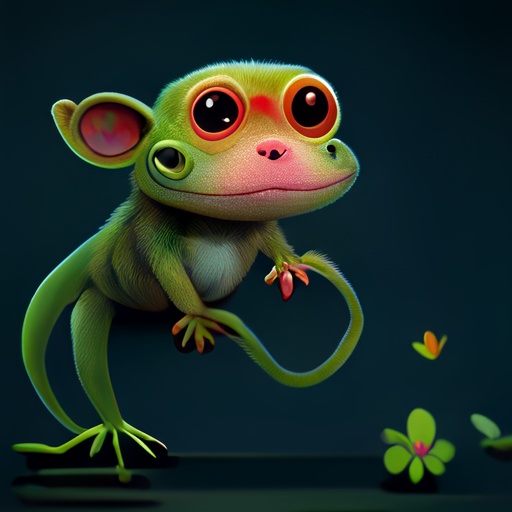}
}
\hfill
\subfloat[Output from \oursabbr{} (ours), generating a clearly separated monkey and frog.\label{sample2}]{
\includegraphics[width=.45\linewidth]{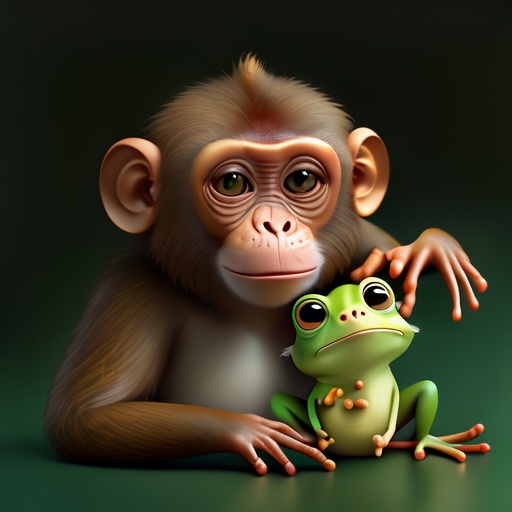}
}

\caption{Example images used for GPT-based evaluation.}
\label{fig:sample}
\end{figure}

To demonstrate that our GPT-based evaluation provides reliable assessments, we include sample responses generated by GPT. Specifically, we test the prompt \textit{“a monkey and a frog”} from the Animal-Animal category of the Attend-and-Excite dataset. The outputs generated by Meissonic and \oursabbr{} are shown in \cref{fig:sample}.

As shown in \cref{sample1}, Meissonic fails to generate a distinct monkey and frog, instead producing a green creature that appears to be a mixture of both. In contrast, as shown in \cref{sample2}, \oursabbr{} successfully generates a clearly separated and recognizable monkey and frog.

GPT assigns a score of 5 to the baseline output, accurately describing it as \textit{“a creature that appears to be a hybrid of a monkey and a frog.”} In comparison, it gives a score of 10 to our output, noting that it \textit{“depicts both a monkey and a frog clearly and accurately.”} These examples illustrate that GPT’s assessments closely align with human judgment, supporting the reliability of our GPT-based evaluation.

\begin{lstlisting}[caption=GPT evaluation output for the image generated by Meissonic (baseline)., basicstyle=\ttfamily,breaklines=true,frame=single,breakindent=0pt,numbers=none,xleftmargin=0pt, framexleftmargin=0pt]
{
\"score\" : [5],
\"reasoning\" : \"The image depicts a creature that appears to be a hybrid of a monkey and a frog, rather than showing a distinct monkey and a frog as separate entities. While creative, it does not fully align with the prompt's request for both animals to be present.\"
}
\end{lstlisting}

\begin{lstlisting}[caption=GPT evaluation output for the image generated by \oursabbr{} (ours)., basicstyle=\ttfamily,breaklines=true,frame=single,breakindent=0pt,numbers=none,xleftmargin=0pt, framexleftmargin=0pt]
{
\"score\" : [10],
\"reasoning\" : \"The image perfectly aligns with the prompt, depicting both a monkey and a frog clearly and accurately. The composition is well-executed, and the subjects are easily identifiable.\"
}
\end{lstlisting}

\section{Comparison with Stable Diffusion}

To better understand the effectiveness of \oursabbr{} in Masked Generative Transformers, we compare it with the effectiveness of Attend-and-Excite~\citep{attend} applied to Diffusion Models. In \cref{tab:stable_diffusion}, we report the performance of Attend-and-Excite applied to Stable Diffusion 2.1.

First, we observe that the Stable Diffusion 2.1 baseline performs consistently well across all datasets, with relatively small performance differences between them. This contrasts with Meissonic~\citep{meissonic}, which performs poorly on the Animal-Animal and SSD datasets, but achieves strong performance on the Animal-Object and Object-Object datasets.

Overall, we observe that the performance gains from Attend-and-Excite on Stable Diffusion 2.1 are larger than those from \oursabbr{} on Meissonic. However, this difference stems from the fundamental nature of the two approaches. Attend-and-Excite relies on gradient-based iterative refinement, which introduces significant inference overhead compared to its baseline. In contrast, our method is training-free and guides the sampling order with negligible inference cost.

As discussed in the experiment section, \oursabbr{} incurs only 0.13\% additional runtime, whereas Attend-and-Excite roughly doubles the inference time. Considering this inference overhead, \oursabbr{} demonstrates effective performance gains at minimal cost.

\begin{table}[ht!]
\centering
\caption{Comparison with improvements in Diffusion Models. We report the performance of Attend-and-Excite~\citep{attend} applied to Stable Diffusion 2.1~\citep{sd}.} 
\label{tab:stable_diffusion}
        \begin{subtable}[t]{\linewidth}
\caption{CLIP text-image similarities.} 
\label{tab:clip_stable_diffusion}
\resizebox{\linewidth}{!}{
\begin{tabular}{l c c c c c c} 
\toprule
\multirow{2.5}{*}{\textbf{Method / Dataset}}  &\multicolumn{3}{c}{\textbf{Attend-and-Excite}} & \multicolumn{2}{c}{\textbf{SSD}} & \multirow{2.5}{*}{\textbf{Avg.}}\\
\cmidrule(lr){2-4}\cmidrule(lr){5-6}
 & A.-A. & A.-O. & O.-O. & Two & Three\\
\midrule
Meissonic & 29.82 & 36.15 & \underline{37.65} & 29.73 & 30.24 & \cellcolor{avg}32.72\\
\textbf{\oursabbr{}\phantom{\dag{}} (ours)} & \underline{30.43} & \underline{36.28} & \textbf{37.65} & \underline{30.08} & \underline{30.52} & \cellcolor{avg}\underline{32.99}\\
\textbf{\oursabbr{}\dag{} (ours)} & \textbf{30.50} & \textbf{36.31} & 37.54 & \textbf{30.17} & \textbf{30.61} & \cellcolor{avg}\textbf{33.03} \\
\midrule
Stable Diffusion 2.1 & 33.12 & 35.39 & 34.27 & 32.07 & 31.51 & \cellcolor{avg}33.27\\
Attend-and-Excite & \textbf{34.24} &  \textbf{36.03} & \textbf{36.74} & \textbf{32.45} & \textbf{32.15} & \cellcolor{avg}\textbf{34.32}\\
\bottomrule
\end{tabular}
}
\end{subtable}

\medskip

\begin{subtable}[t]{\linewidth}
\caption{CLIP text-text similarities.} 
\label{tab:blip_stable_diffusion}
\resizebox{1.0\linewidth}{!}{
\begin{tabular}{l c c c c c c} 
\toprule
\multirow{2.5}{*}{\textbf{Method / Dataset}}  &\multicolumn{3}{c}{\textbf{Attend-and-Excite}} & \multicolumn{2}{c}{\textbf{SSD}} & \multirow{2.5}{*}{\textbf{Avg.}}\\
\cmidrule(lr){2-4}\cmidrule(lr){5-6}
 & A.-A. & A.-O. & O.-O. & Two & Three\\
\midrule
Meissonic & 73.98 & \underline{87.05} & \underline{86.34} & 69.21 & 72.24 & \cellcolor{avg}77.76\\
\textbf{\oursabbr{}\phantom{\dag{}} (ours)} & \underline{77.05} & \textbf{87.16} & \textbf{86.77} & \underline{69.94} & \underline{72.66} & \cellcolor{avg}\underline{78.72}\\
\textbf{\oursabbr{}\dag{} (ours)} & \textbf{77.21} & 86.92 & 85.88 & \textbf{70.44} & \textbf{73.50} & \cellcolor{avg}\textbf{78.79} \\
\midrule
Stable Diffusion 2.1 & 81.52 & 82.67 & 78.04 & 72.29 & 73.00 & \cellcolor{avg}77.50\\
Attend-and-Excite & \textbf{85.13} & \textbf{85.15} & \textbf{83.33} & \textbf{73.25} & \textbf{74.80} & \cellcolor{avg}\textbf{80.33}\\
\bottomrule
\end{tabular}
}
\end{subtable}

\medskip

\begin{subtable}[t]{\linewidth}
\caption{GPT-based evaluation.} 
\label{tab:gpt_stable_diffusion}
\resizebox{1.0\linewidth}{!}{
\begin{tabular}{l c c c c c c} 
\toprule
\multirow{2.5}{*}{\textbf{Method / Dataset}}  &\multicolumn{3}{c}{\textbf{Attend-and-Excite}} & \multicolumn{2}{c}{\textbf{SSD}} & \multirow{2.5}{*}{\textbf{Avg.}}\\
\cmidrule(lr){2-4}\cmidrule(lr){5-6}
 & A.-A. & A.-O. & O.-O. & Two & Three\\
\midrule
Meissonic & 6.18 & \textbf{9.65} & 8.52 &  5.59 & 5.03 & \cellcolor{avg}6.99 \\
\textbf{\oursabbr{}\phantom{\dag{}} (ours)} & \underline{6.80} & \underline{9.62} & \underline{8.74} & \underline{6.16} & \textbf{5.37} & \cellcolor{avg}\underline{7.34}\\
\textbf{\oursabbr{}\dag{} (ours)} & \textbf{6.86} & 9.61 & \textbf{8.81} & \textbf{6.27} & \underline{5.36} & \cellcolor{avg}\textbf{7.38} \\
\midrule
Stable Diffusion 2.1 & 7.62 & 8.33 & 6.41 & 7.48 & 6.87 & \cellcolor{avg}7.34\\
Attend-and-Excite & \textbf{8.85} & \textbf{9.36} & \textbf{8.14} & \textbf{8.67} & \textbf{7.89} & \cellcolor{avg}\textbf{8.58}\\
\bottomrule
\end{tabular}
}
\end{subtable}
\end{table}

\section{Additional Ablation Study}
In this section, we conduct additional ablation studies to further analyze \oursabbr{}. In the ablation study, we set $w_a = 3$.

\subsection{Ablation on the Number of Guidance Timesteps}

In our default setting, \oursabbr{} is applied for the first 16 timesteps, with the remaining 48 following the baseline. To further investigate its impact, we conduct an ablation with 0 (Meissonic, baseline), 1, 2, 4, 8, 16 (ours), 32, and all 64 timesteps (full), and report the results in \cref{tab:ablation_timestep}.

First, we observe a consistent trend across all metrics where the average score increases as the number of guidance timesteps grows. This indicates that \oursabbr{} effectively provides guidance in a direction that improves performance.

Notably, applying \oursabbr{} during the first 16 timesteps yields results comparable to full guidance, suggesting that the overall structure of the final output is largely determined during the early stages of generation. Furthermore, even applying \oursabbr{} only during the first 4 timesteps, or even just the first timestep, yields noticeable improvements over the Meissonic (baseline). This highlights the importance of providing guidance at early timesteps and the effectiveness of \oursabbr{}.

\begin{table}[ht!]
\centering
\caption{Ablation on the number of guidance timesteps.} 
\label{tab:ablation_timestep}
        \begin{subtable}[t]{\linewidth}
\caption{CLIP text-image similarities.} 
\label{tab:clip_ablation_timestep}
\resizebox{\linewidth}{!}{
\begin{tabular}{l c c c c c c} 
\toprule
\multirow{2.5}{*}{\textbf{Method / Dataset}}  &\multicolumn{3}{c}{\textbf{Attend-and-Excite}} & \multicolumn{2}{c}{\textbf{SSD}} & \multirow{2.5}{*}{\textbf{Avg.}}\\
\cmidrule(lr){2-4}\cmidrule(lr){5-6}
 & A.-A. & A.-O. & O.-O. & Two & Three\\
\midrule
Meissonic & 29.82 & 36.15 & 37.65 & 29.73 & 30.24 & \cellcolor{avg}32.72\\
Halton Scheduler & \textbf{30.81} & 36.15 & 37.52 & 30.01 & 30.46 & \cellcolor{avg}32.99\\
\textbf{\oursabbr{} (1)} & 29.84 & 36.27 & 37.63 & 29.87 & 30.40 & \cellcolor{avg}32.80 \\
\textbf{\oursabbr{} (2)} & 30.20 & 36.25 & \textbf{37.82} & 29.97 & 30.44 & \cellcolor{avg}32.94 \\
\textbf{\oursabbr{} (4)} & 30.27 & 36.27 & \underline{37.74} & 30.02 & \textbf{30.63} & \cellcolor{avg}32.99 \\
\textbf{\oursabbr{} (8)} & 30.44 & 36.30 & 37.62 & 30.03 & 30.60 & \cellcolor{avg}33.00 \\
\textbf{\oursabbr{} (16)} & 30.43 & 36.28 & 37.65 & 30.08 & 30.52 & \cellcolor{avg}32.99\\
\textbf{\oursabbr{} (32)} & 30.47 & \textbf{36.36} & 37.53 & \underline{30.16} & 30.59 & \cellcolor{avg}\underline{33.02} \\
\textbf{\oursabbr{} (64)} & \underline{30.50} & \underline{36.31} & 37.54 & \textbf{30.17} & \underline{30.61} & \cellcolor{avg}\textbf{33.03} \\
\bottomrule
\end{tabular}
}
\end{subtable}

\medskip

\begin{subtable}[t]{\linewidth}
\caption{CLIP text-text similarities.} 
\label{tab:blip_ablation_timestep}
\resizebox{1.0\linewidth}{!}{
\begin{tabular}{l c c c c c c} 
\toprule
\multirow{2.5}{*}{\textbf{Method / Dataset}}  &\multicolumn{3}{c}{\textbf{Attend-and-Excite}} & \multicolumn{2}{c}{\textbf{SSD}} & \multirow{2.5}{*}{\textbf{Avg.}}\\
\cmidrule(lr){2-4}\cmidrule(lr){5-6}
 & A.-A. & A.-O. & O.-O. & Two & Three\\
\midrule
Meissonic & 73.98 & 87.05 & 86.34 & 69.21 & 72.24 & \cellcolor{avg}77.76\\
Halton Scheduler & 76.47 & 87.27 & 86.10 & 69.85 & 72.97 & \cellcolor{avg}78.53\\
\textbf{\oursabbr{} (1)} & 74.55 & \textbf{87.51} & 86.33 & 70.01 & 72.94 & \cellcolor{avg}78.27 \\
\textbf{\oursabbr{} (2)} & 75.55 & \underline{87.48} & 86.89 & 69.52 & 72.57 & \cellcolor{avg}78.40\\
\textbf{\oursabbr{} (4)} & 75.75 & 87.03 & \textbf{87.02} & 70.08 & 73.17 & \cellcolor{avg}78.61 \\
\textbf{\oursabbr{} (8)} & 76.71 & 86.90 & \underline{86.92} & 70.15 & \underline{73.43} & \cellcolor{avg}\textbf{78.82}\\
\textbf{\oursabbr{} (16)} & \underline{77.05} & 87.16 & 86.77 & 69.94 & 72.66 & \cellcolor{avg}78.72\\
\textbf{\oursabbr{} (32)} & 76.98 & 86.99 & 86.34 & \underline{70.19} & 73.30 & \cellcolor{avg}78.76\\
\textbf{\oursabbr{} (64)} & \textbf{77.21} & 86.92 & 85.88 & \textbf{70.44} & \textbf{73.50} & \cellcolor{avg}\underline{78.79} \\

\bottomrule
\end{tabular}
}
\end{subtable}

\medskip

\begin{subtable}[t]{\linewidth}
\caption{GPT-based evaluation.} 
\label{tab:gpt_ablation_timestep}
\resizebox{1.0\linewidth}{!}{
\begin{tabular}{l c c c c c c} 
\toprule
\multirow{2.5}{*}{\textbf{Method / Dataset}}  &\multicolumn{3}{c}{\textbf{Attend-and-Excite}} & \multicolumn{2}{c}{\textbf{SSD}} & \multirow{2.5}{*}{\textbf{Avg.}}\\
\cmidrule(lr){2-4}\cmidrule(lr){5-6}
 & A.-A. & A.-O. & O.-O. & Two & Three\\
\midrule
Meissonic & 6.18 & 9.65 & 8.52 &  5.59 & 5.03 & \cellcolor{avg}6.99 \\
Halton Scheduler & 6.65 & \textbf{9.67} & \textbf{8.88} & 5.79 & 5.35 & \cellcolor{avg}7.27\\
\textbf{\oursabbr{} (1)} & 6.25 & \underline{9.67} & 8.59 & 5.70 & 5.21 & \cellcolor{avg}7.08 \\
\textbf{\oursabbr{} (2)} & 6.54 & 9.59 & 8.74 & 5.82 & 5.08 & \cellcolor{avg}7.15 \\
\textbf{\oursabbr{} (4)} & 6.64 & 9.64 & \underline{8.87} & 6.05 & 5.18 & \cellcolor{avg}7.28 \\
\textbf{\oursabbr{} (8)} & \textbf{6.90} & 9.62 & 8.83 & 6.07 & 5.19 & \cellcolor{avg}7.32 \\
\textbf{\oursabbr{} (16)} & 6.80 & 9.62 & 8.74 & 6.16 & \underline{5.37} & \cellcolor{avg}7.34\\
\textbf{\oursabbr{} (32)} & 6.84 & 9.64 & 8.78 & \underline{6.26} & \textbf{5.39} & \cellcolor{avg}\underline{7.38} \\
\textbf{\oursabbr{} (64)} & \underline{6.86} & 9.61 & 8.81 & \textbf{6.27} & 5.36 & \cellcolor{avg}\textbf{7.38} \\
\bottomrule
\end{tabular}
}
\end{subtable}
\end{table}

\subsection{Ablation on Gaussian Smoothing of the Attention Map}
Following Attend-and-Excite~\citep{attend}, we apply Gaussian smoothing to the attention map when obtaining it for contrastive attention guidance computation in our default setting. We investigate how this smoothing affects the performance of \oursabbr{}. 

As shown in \cref{tab:ablation_blur}, \oursabbr{} without Gaussian smoothing already outperforms Meissonic (baseline) across all metrics in terms of average scores, and in some cases even outperforms \oursabbr{} (ours). However, our method yields consistently stronger and more stable performance overall. This suggests that Gaussian smoothing helps stabilize the effectiveness of \oursabbr{}, which is consistent with findings from Attend-and-Excite.

\begin{table}[ht!]
\centering
\caption{Ablation results for Gaussian smoothing of the attention map.} 
\label{tab:ablation_blur}
        \begin{subtable}[t]{\linewidth}
\caption{CLIP text-image similarities.} 
\label{tab:clip_ablation_blur}
\resizebox{\linewidth}{!}{
\begin{tabular}{l c c c c c c} 
\toprule
\multirow{2.5}{*}{\textbf{Method / Dataset}}  &\multicolumn{3}{c}{\textbf{Attend-and-Excite}} & \multicolumn{2}{c}{\textbf{SSD}} & \multirow{2.5}{*}{\textbf{Avg.}}\\
\cmidrule(lr){2-4}\cmidrule(lr){5-6}
 & A.-A. & A.-O. & O.-O. & Two & Three\\
\midrule
Meissonic & 29.82 & \underline{36.15} & \underline{37.65} & 29.73 & 30.24 & \cellcolor{avg}32.72\\
\textbf{\oursabbr{} (ours)} & \textbf{30.43} & \textbf{36.28} & \textbf{37.65} & \textbf{30.08} & \textbf{30.52} & \cellcolor{avg}\textbf{32.99}\\
\oursabbr{} w/o blur & \underline{30.21} & 36.13 & 37.62 & \underline{29.97} & \underline{30.39} & \cellcolor{avg}\underline{32.86}\\
\bottomrule
\end{tabular}
}
\end{subtable}

\medskip

\begin{subtable}[t]{\linewidth}
\caption{CLIP text-text similarities.} 
\label{tab:blip_ablation_blur}
\resizebox{1.0\linewidth}{!}{
\begin{tabular}{l c c c c c c} 
\toprule
\multirow{2.5}{*}{\textbf{Method / Dataset}}  &\multicolumn{3}{c}{\textbf{Attend-and-Excite}} & \multicolumn{2}{c}{\textbf{SSD}} & \multirow{2.5}{*}{\textbf{Avg.}}\\
\cmidrule(lr){2-4}\cmidrule(lr){5-6}
 & A.-A. & A.-O. & O.-O. & Two & Three\\
\midrule
Meissonic & 73.98 & \underline{87.05} & 86.34 & 69.21 & 72.24 & \cellcolor{avg}77.76\\
\textbf{\oursabbr{} (ours)} & \textbf{77.05} & \textbf{87.16} & \textbf{86.77} & \textbf{69.94} & \underline{72.66} & \cellcolor{avg}\textbf{78.72}\\
\oursabbr{} w/o blur & \underline{76.07} & 86.41 & \underline{87.07} & \underline{69.55} & \textbf{73.57} & \cellcolor{avg}\underline{78.53}\\
\bottomrule
\end{tabular}
}
\end{subtable}

\medskip

\begin{subtable}[t]{\linewidth}
\caption{GPT-based evaluation.} 
\label{tab:gpt_ablation_blur}
\resizebox{1.0\linewidth}{!}{
\begin{tabular}{l c c c c c c} 
\toprule
\multirow{2.5}{*}{\textbf{Method / Dataset}}  &\multicolumn{3}{c}{\textbf{Attend-and-Excite}} & \multicolumn{2}{c}{\textbf{SSD}} & \multirow{2.5}{*}{\textbf{Avg.}}\\
\cmidrule(lr){2-4}\cmidrule(lr){5-6}
 & A.-A. & A.-O. & O.-O. & Two & Three\\
\midrule
Meissonic & 6.18 & \textbf{9.65} & 8.52 &  5.59 & 5.03 & \cellcolor{avg}6.99 \\
\textbf{\oursabbr{} (ours)} & \textbf{6.80} & \underline{9.62} & \underline{8.74} & \textbf{6.16} & \textbf{5.37} & \cellcolor{avg}\textbf{7.34}\\
\oursabbr{} w/o blur & \underline{6.50} & 9.56 & \textbf{8.77} & \underline{6.09} & \underline{5.24} & \cellcolor{avg}\underline{7.23}\\
\bottomrule
\end{tabular}
}
\end{subtable}

\end{table}

\begin{table}[ht!]
\centering
\caption{Ablation results for confidence and randomness terms.} 
\label{tab:ablation_guidance}
        \begin{subtable}[t]{\linewidth}
\caption{CLIP text-image similarities.} 
\label{tab:clip_ablation_guidance}
\resizebox{\linewidth}{!}{
\begin{tabular}{l c c c c c c} 
\toprule
\multirow{2.5}{*}{\textbf{Method / Dataset}}  &\multicolumn{3}{c}{\textbf{Attend-and-Excite}} & \multicolumn{2}{c}{\textbf{SSD}} & \multirow{2.5}{*}{\textbf{Avg.}}\\
\cmidrule(lr){2-4}\cmidrule(lr){5-6}
 & A.-A. & A.-O. & O.-O. & Two & Three\\
\midrule
Random ($F_g$) & 26.50 & 31.81 & 30.97 & 27.90 & 28.14 & \cellcolor{avg}29.06 \\
Confidence ($F_c$) & 28.37 & 35.33 & 36.69 & 29.34 & 29.64 & \cellcolor{avg}31.87 \\
Meissonic ($F_c+F_g$) & \underline{29.82} & \underline{36.15} & \underline{37.65} & 29.73 & \underline{30.24} & \cellcolor{avg}\underline{32.72}\\
\textbf{\oursabbr{} (ours)} & \textbf{30.43} & \textbf{36.28} & \textbf{37.65} & \textbf{30.08} & \textbf{30.52} & \cellcolor{avg}\textbf{32.99}\\
\oursabbr{} w/o $F_g$ & 29.36 & 35.96 & 37.53 & \underline{29.78} & 30.01 & \cellcolor{avg}32.53\\
\oursabbr{} w/o $F_c$ & 28.39 & 35.31 & 36.33 & 29.10 & 29.59 & \cellcolor{avg}31.74\\
\oursabbr{} w/o $F_c, F_g$ & 28.79 & 34.88 & 36.26 & 29.16 & 29.57 & \cellcolor{avg}31.73\\
\bottomrule
\end{tabular}
}
\end{subtable}

\medskip

\begin{subtable}[t]{\linewidth}
\caption{CLIP text-text similarities.} 
\label{tab:blip_ablation_guidance}
\resizebox{1.0\linewidth}{!}{
\begin{tabular}{l c c c c c c} 
\toprule
\multirow{2.5}{*}{\textbf{Method / Dataset}}  &\multicolumn{3}{c}{\textbf{Attend-and-Excite}} & \multicolumn{2}{c}{\textbf{SSD}} & \multirow{2.5}{*}{\textbf{Avg.}}\\
\cmidrule(lr){2-4}\cmidrule(lr){5-6}
 & A.-A. & A.-O. & O.-O. & Two & Three\\
\midrule
Random ($F_g$) & 66.70 & 76.98 & 69.76 & 64.03 & 65.76 & \cellcolor{avg}68.65\\
Confidence ($F_c$) & 73.12 & 85.14 & 84.88 & 67.71 & 71.04 & \cellcolor{avg}76.38 \\
Meissonic ($F_c+F_g$) & 73.98 & \underline{87.05} & \underline{86.34} & 69.21 & \underline{72.24} & \cellcolor{avg}77.76\\
\textbf{\oursabbr{} (ours)} & \textbf{77.05} & \textbf{87.16} & \textbf{86.77} & \textbf{69.94} & \textbf{72.66} & \cellcolor{avg}\textbf{78.72}\\
\oursabbr{} w/o $F_g$ & \underline{76.19} & 86.02 & 85.59 & \underline{69.94} & 71.32 & \cellcolor{avg}\underline{77.81}\\
\oursabbr{} w/o $F_c$ & 73.20 & 85.28 & 83.36 & 68.43 & 71.71 & \cellcolor{avg}76.40\\
\oursabbr{} w/o $F_c, F_g$ & 74.05 & 84.01 & 82.67 & 69.67 & 70.93 & \cellcolor{avg}76.27\\
\bottomrule
\end{tabular}
}
\end{subtable}

\medskip

\begin{subtable}[t]{\linewidth}
\caption{GPT-based evaluation.} 
\label{tab:gpt_ablation_guidance}
\resizebox{1.0\linewidth}{!}{
\begin{tabular}{l c c c c c c} 
\toprule
\multirow{2.5}{*}{\textbf{Method / Dataset}}  &\multicolumn{3}{c}{\textbf{Attend-and-Excite}} & \multicolumn{2}{c}{\textbf{SSD}} & \multirow{2.5}{*}{\textbf{Avg.}}\\
\cmidrule(lr){2-4}\cmidrule(lr){5-6}
 & A.-A. & A.-O. & O.-O. & Two & Three\\
\midrule
Random ($F_g$) & 5.94 & 8.52 & 6.57 & 5.65 & \textbf{5.59} & \cellcolor{avg}6.45 \\
Confidence ($F_c$) & 5.78 & 9.19 & 8.13 & 5.18 & 4.57 & \cellcolor{avg}6.57 \\
Meissonic ($F_c+F_g$) & 6.18 & \textbf{9.65} & 8.52 &  5.59 & 5.03 & \cellcolor{avg}6.99 \\
\textbf{\oursabbr{} (ours)} & \textbf{6.80} & \underline{9.62} & \textbf{8.74} & \textbf{6.16} & 5.37 & \cellcolor{avg}\textbf{7.34}\\
\oursabbr{} w/o $F_g$ & \underline{6.61} & 9.50 & 8.66 & \underline{6.15} & \underline{5.42} & \cellcolor{avg}\underline{7.27}\\
\oursabbr{} w/o $F_c$ & 6.14 & 9.20 & \underline{8.68} & 5.58 & 5.14 & \cellcolor{avg}6.95\\
\oursabbr{} w/o $F_c, F_g$ & 5.35 & 8.54 & 8.30 & 5.16 & 4.94 & \cellcolor{avg}6.46\\
\bottomrule
\end{tabular}
}
\end{subtable}

\end{table}

\subsection{Ablation on Confidence and Randomness Terms}

We use both the confidence ($F_c$) and randomness ($F_g$) terms by default when evaluating the effectiveness of \oursabbr{}. To further investigate the role of contrastive attention guidance ($F_a$) and its interaction with these terms, we conduct an ablation study with three additional settings:

\begin{enumerate}[label=(\arabic*),align=left]
    \item \oursabbr{} w/o $F_g$ term ($F(t) = F_c(t) + w_a F_a(t)$),
    \item \oursabbr{} w/o $F_c$ term ($F(t) = F_g(t) + w_a F_a(t)$),
    \item \oursabbr{} w/o $F_c, F_g$ terms ($F(t) = w_a F_a(t)$).
\end{enumerate}

We evaluate how each combination affects performance and analyze the contribution of each term to the overall effectiveness of $F_a$.
The results are shown in \cref{tab:ablation_guidance}.

Notably, both ablations (1) and (2) show clear performance improvements over the setting without \oursabbr{}. This demonstrates the effectiveness of \oursabbr{} itself, independent of the presence of the confidence or randomness terms.
Furthermore, using only $F_a$ (ablation 3) outperforms using only $F_g$ (randomness term) and achieves performance comparable to using only $F_c$ (confidence term).
This suggests that \oursabbr{} not only serves as a guidance mechanism but is also as effective as the confidence term itself.

\section{Additional Qualitative Results}

In this section, we provide additional qualitative results on the Attend-and-Excite~\cite{attend} dataset and the SSD~\cite{ssd} dataset.

\subsection{Attend-and-Excite Dataset}

\begin{figure*}[!htp]
    \centering
    \setlength{\tabcolsep}{0.5pt}
    \renewcommand{\arraystretch}{0.3}
    \resizebox{1.0\linewidth}{!}{
    {\scriptsize
\begin{tabular}{c c c @{\hspace{0.1cm}} c c @{\hspace{0.1cm}} c c @{\hspace{0.1cm}} c c}

    &
    \multicolumn{2}{c}{\textit{"a bear and a frog"}} &
    \multicolumn{2}{c}{\textit{"a cat and a lion"}} &
    \multicolumn{2}{c}{\textit{"a turtle and a pink apple"}} &
    \multicolumn{2}{c}{\textit{"a gray crown and a purple apple"}} \\ \\ 

    {\raisebox{0.45in}{\multirow{2}{*}{\rotatebox{90}{\small Meissonic (baseline)}}}} & \hspace{0.05cm}
    \includegraphics[width=0.1\textwidth]{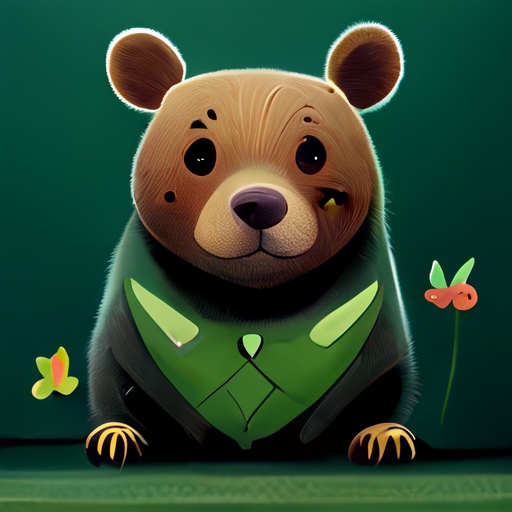} &
    \includegraphics[width=0.1\textwidth]{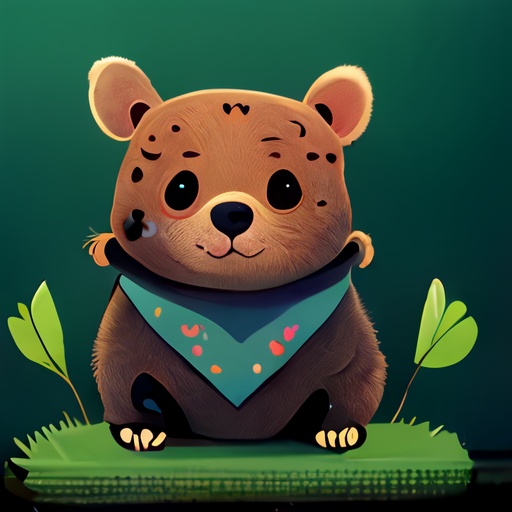} &
    \includegraphics[width=0.1\textwidth]{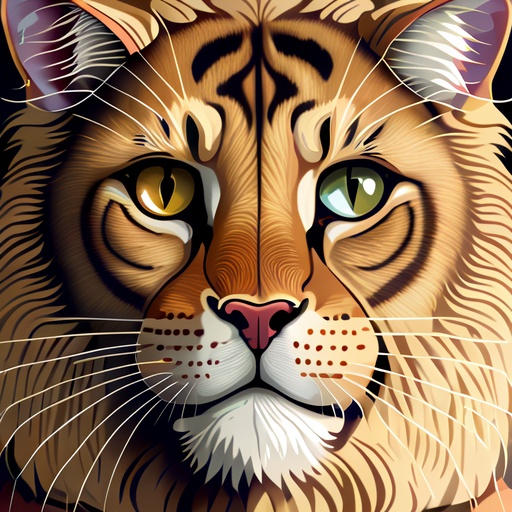} &
    \includegraphics[width=0.1\textwidth]{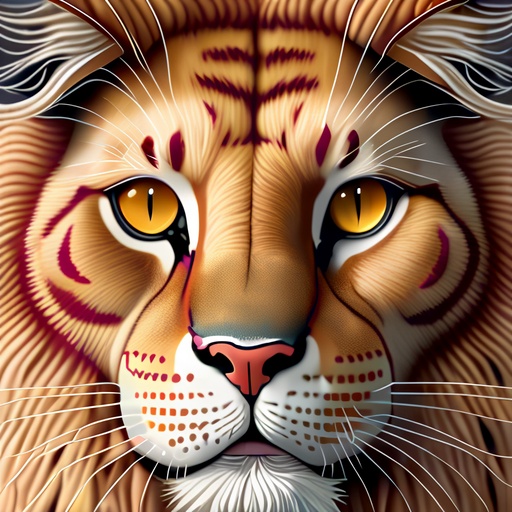} &
    \includegraphics[width=0.1\textwidth]{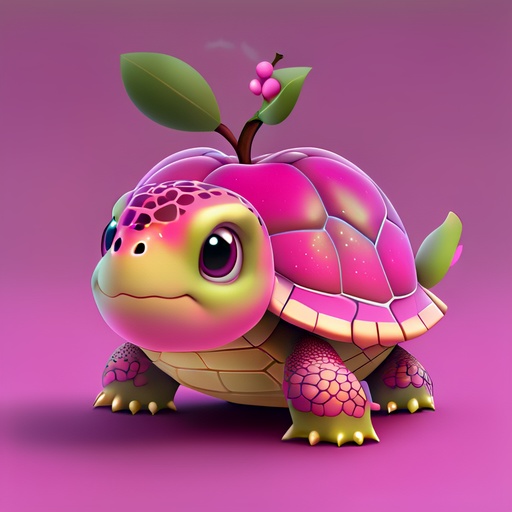} &
    \includegraphics[width=0.1\textwidth]{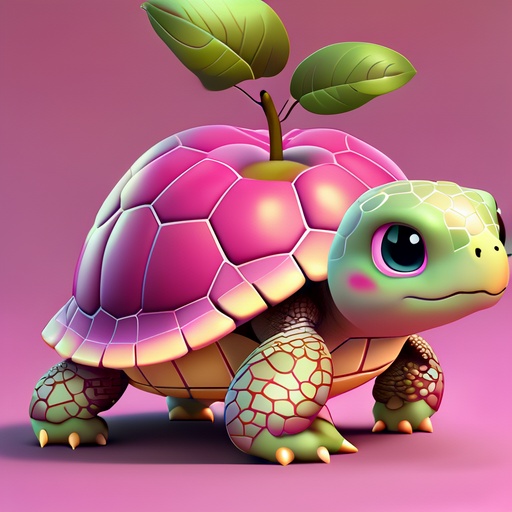} &
    \includegraphics[width=0.1\textwidth]{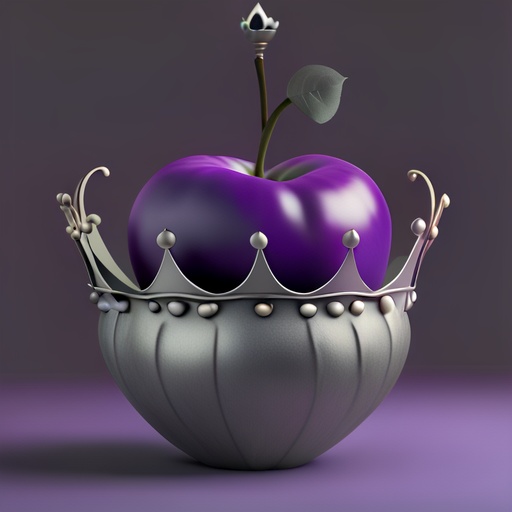} &
    \includegraphics[width=0.1\textwidth]{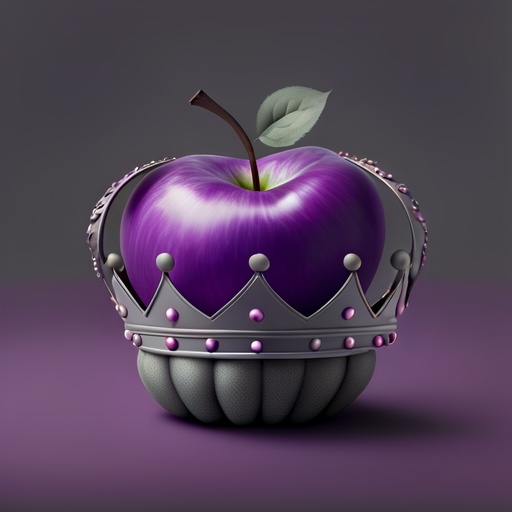} \\

    & \hspace{0.05cm}
    \includegraphics[width=0.1\textwidth]{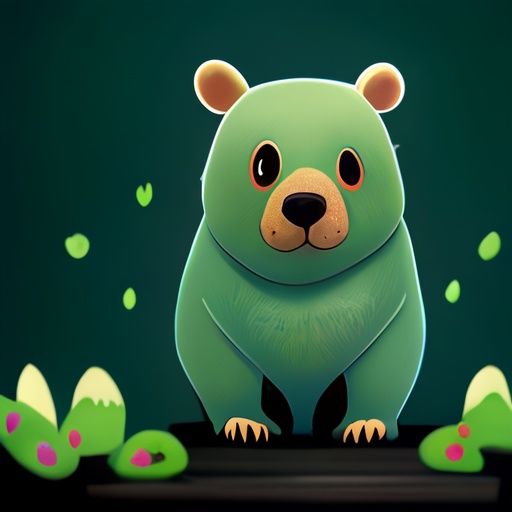} &
    \includegraphics[width=0.1\textwidth]{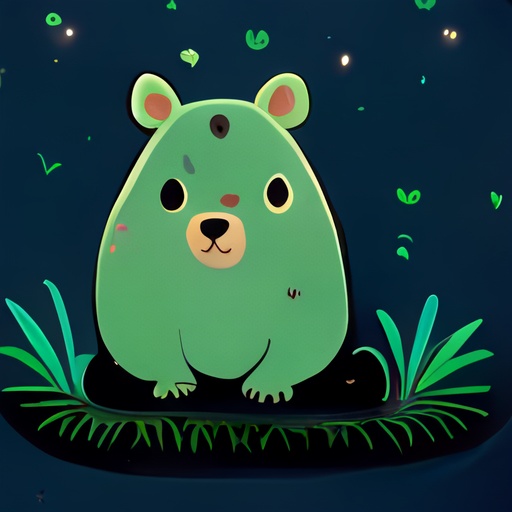} &
    \includegraphics[width=0.1\textwidth]{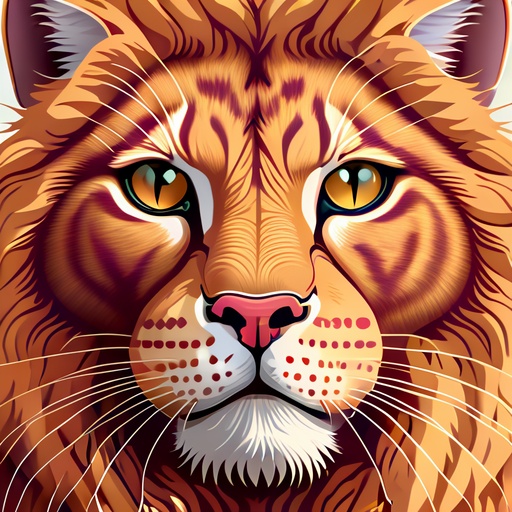} &
    \includegraphics[width=0.1\textwidth]{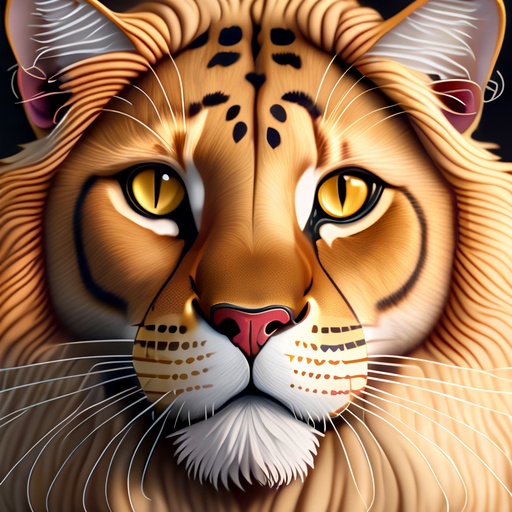} &
    \includegraphics[width=0.1\textwidth]{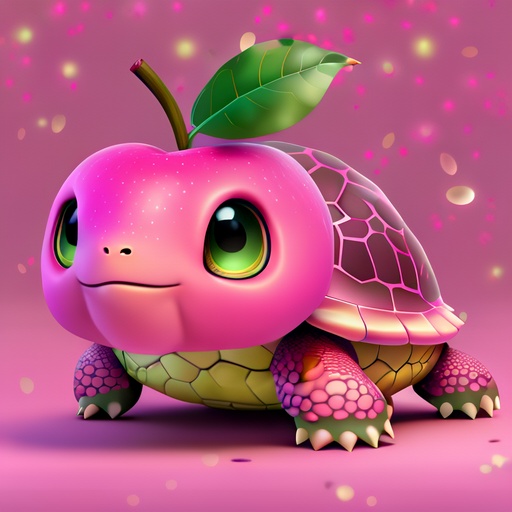} &
    \includegraphics[width=0.1\textwidth]{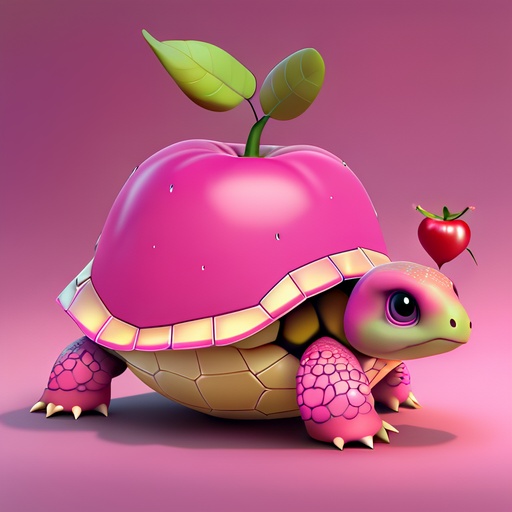} &
    \includegraphics[width=0.1\textwidth]{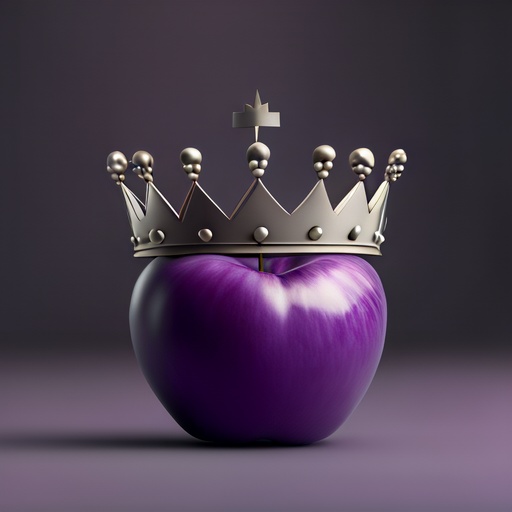} &
    \includegraphics[width=0.1\textwidth]{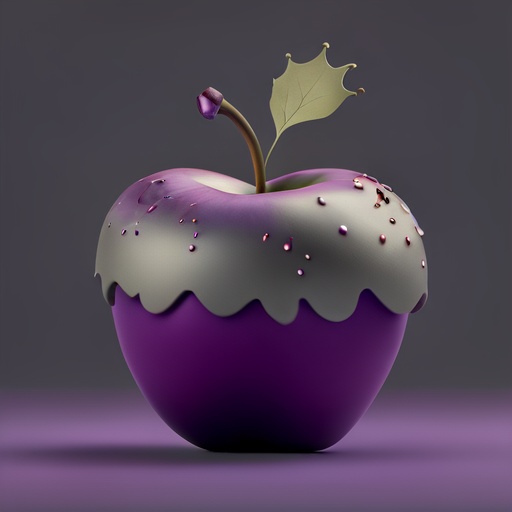} \\ \\ \\

    {\raisebox{0.4in}{\multirow{2}{*}{\rotatebox{90}{\small \textbf{\oursabbr{} (ours)}}}}} & \hspace{0.05cm}
    \includegraphics[width=0.1\textwidth]{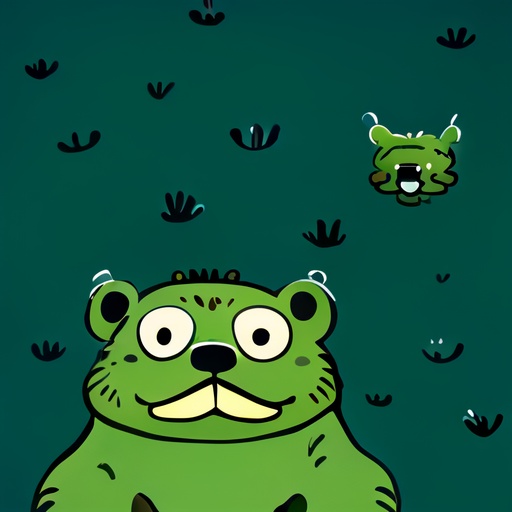} &
    \includegraphics[width=0.1\textwidth]{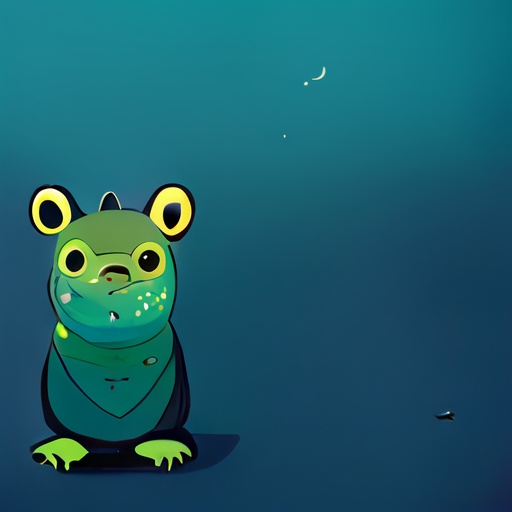} &
    \includegraphics[width=0.1\textwidth]{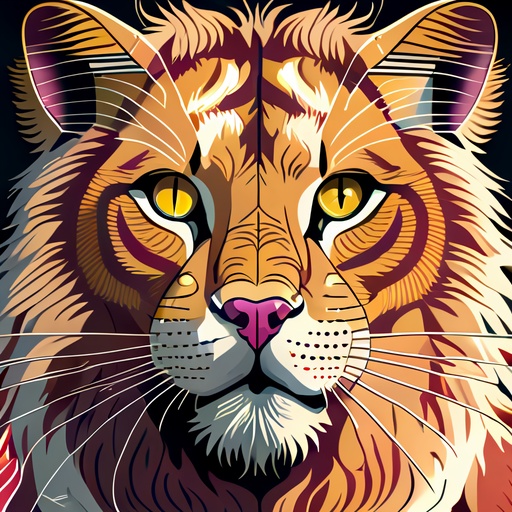} &
    \includegraphics[width=0.1\textwidth]{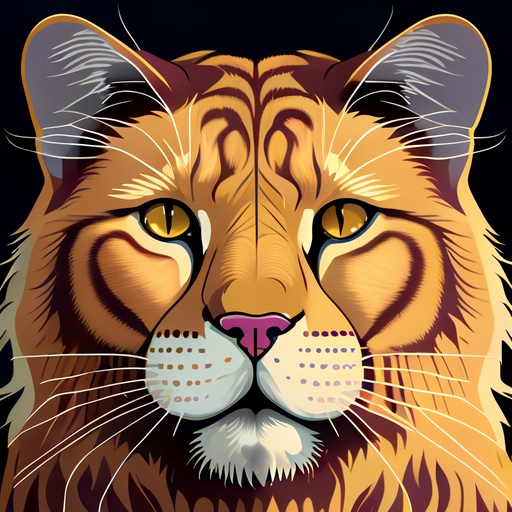} &
    \includegraphics[width=0.1\textwidth]{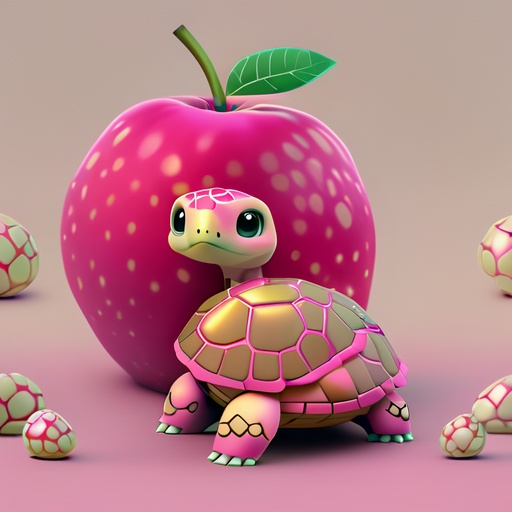} &
    \includegraphics[width=0.1\textwidth]{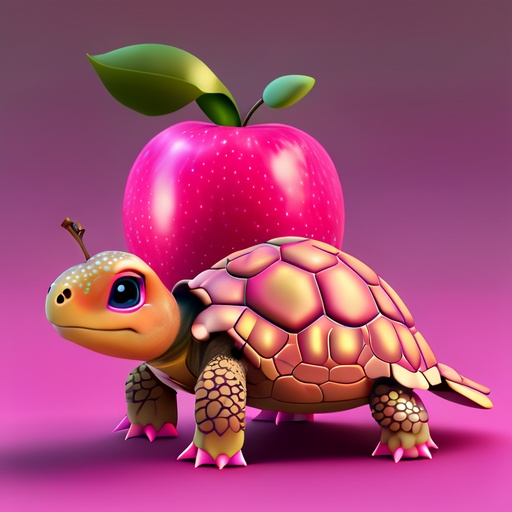} &
    \includegraphics[width=0.1\textwidth]{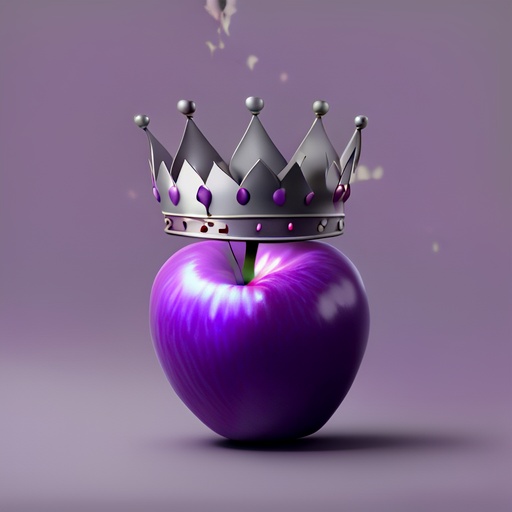} &
    \includegraphics[width=0.1\textwidth]{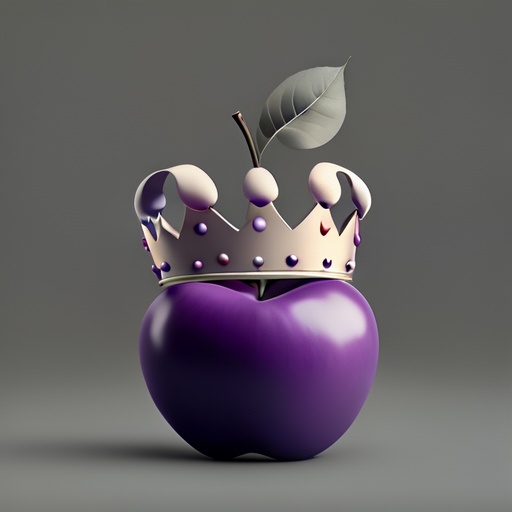} \\

    & \hspace{0.05cm}
    \includegraphics[width=0.1\textwidth]{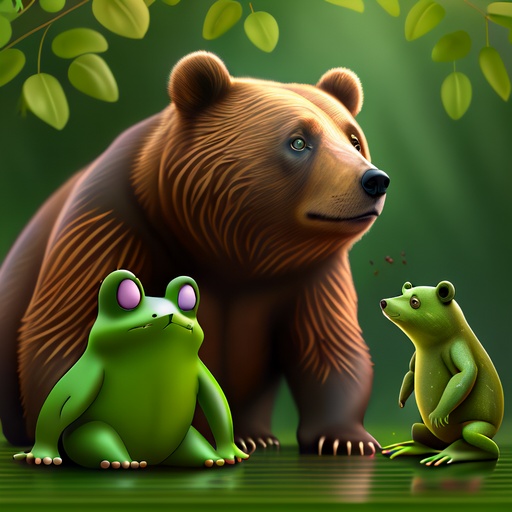} &
    \includegraphics[width=0.1\textwidth]{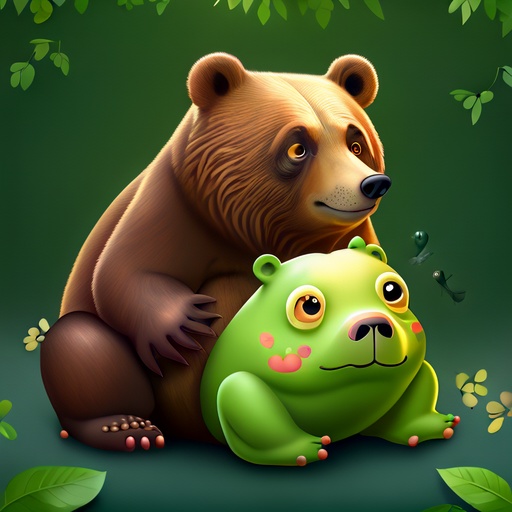} &
    \includegraphics[width=0.1\textwidth]{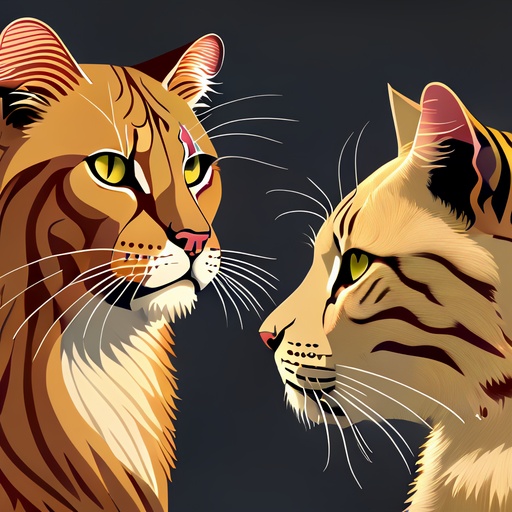} &
    \includegraphics[width=0.1\textwidth]{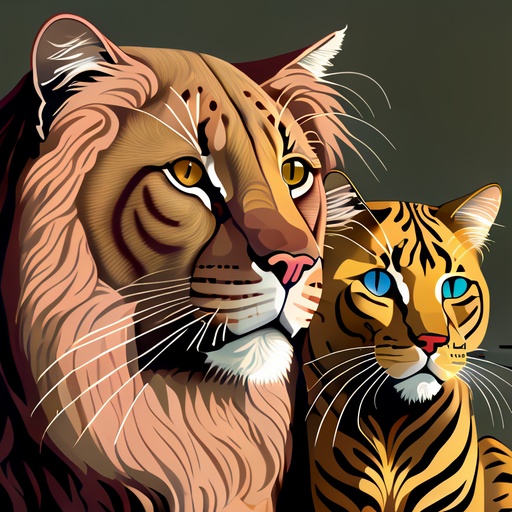} &
    \includegraphics[width=0.1\textwidth]{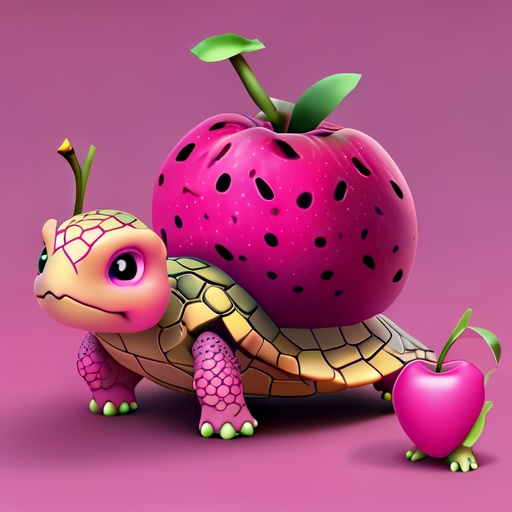} &
    \includegraphics[width=0.1\textwidth]{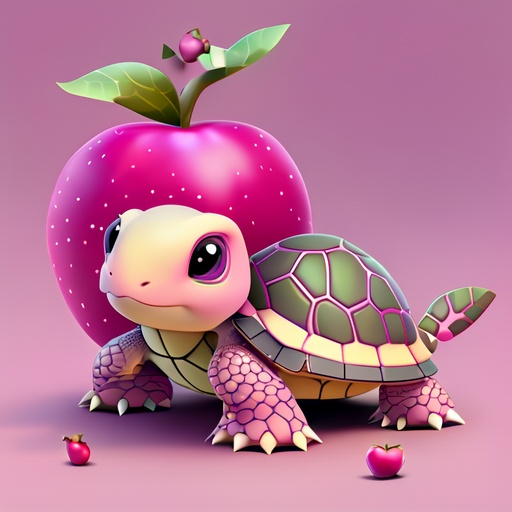} &
    \includegraphics[width=0.1\textwidth]{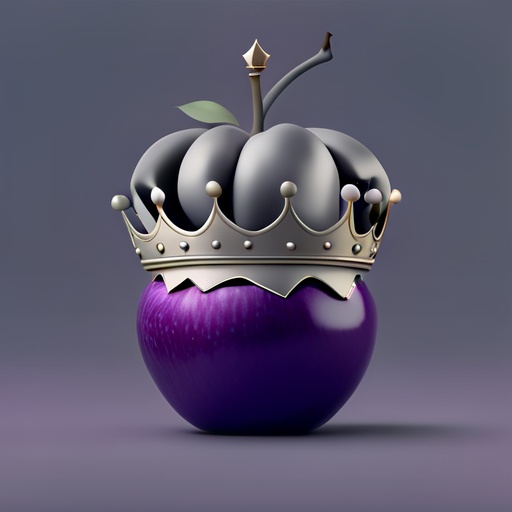} &
    \includegraphics[width=0.1\textwidth]{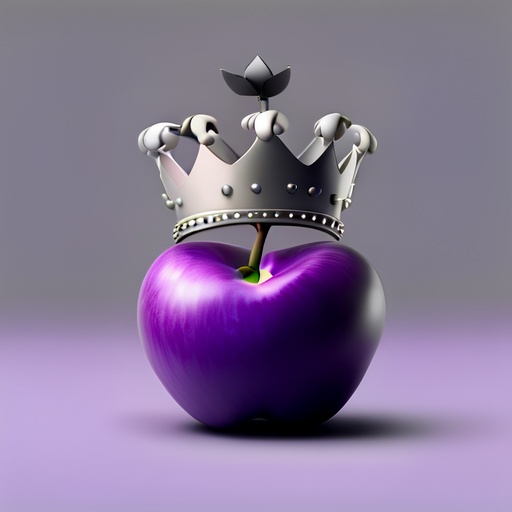} \\ \\ \\ \\ \\ \\ \\ \\ \\

    &
    \multicolumn{2}{c}{\textit{"a dog and a rabbit"}} &
    \multicolumn{2}{c}{\textit{"a rabbit and a mouse"}} &
    \multicolumn{2}{c}{\textit{"a rabbit and a yellow car"}} &
    \multicolumn{2}{c}{\textit{"a green bench and a red apple"}} \\ \\ 

    {\raisebox{0.45in}{\multirow{2}{*}{\rotatebox{90}{\small Meissonic (baseline)}}}} & \hspace{0.05cm}
    \includegraphics[width=0.1\textwidth]{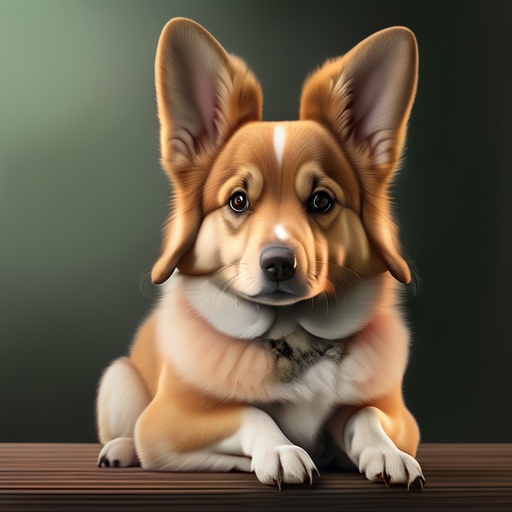} &
    \includegraphics[width=0.1\textwidth]{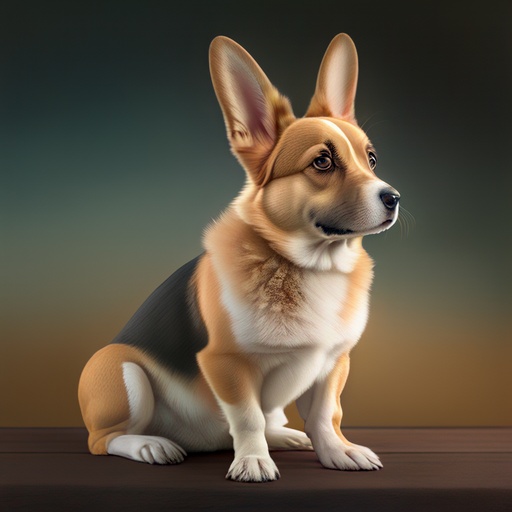} &
    \includegraphics[width=0.1\textwidth]{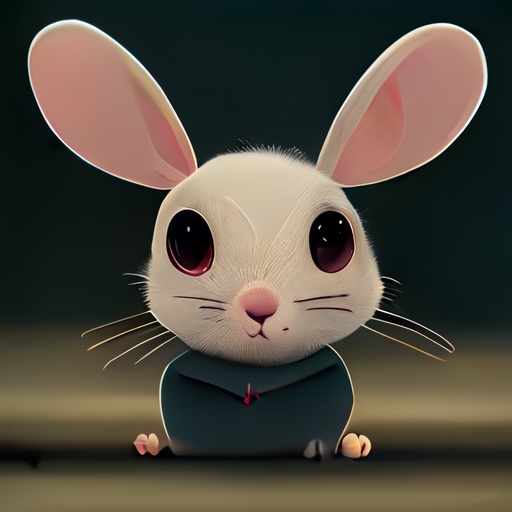} &
    \includegraphics[width=0.1\textwidth]{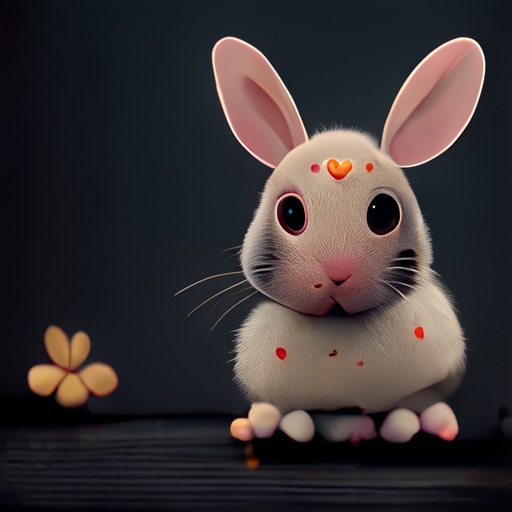} &
    \includegraphics[width=0.1\textwidth]{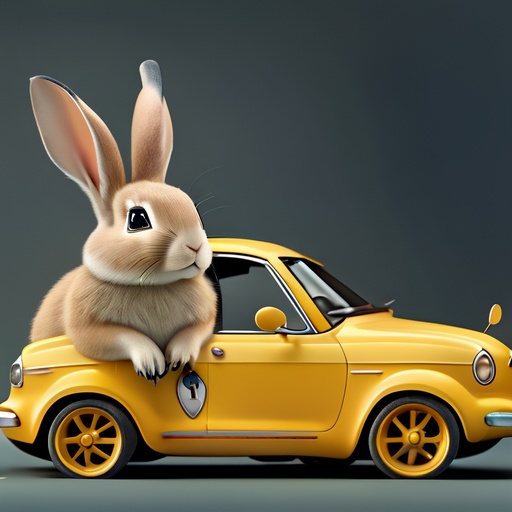} &
    \includegraphics[width=0.1\textwidth]{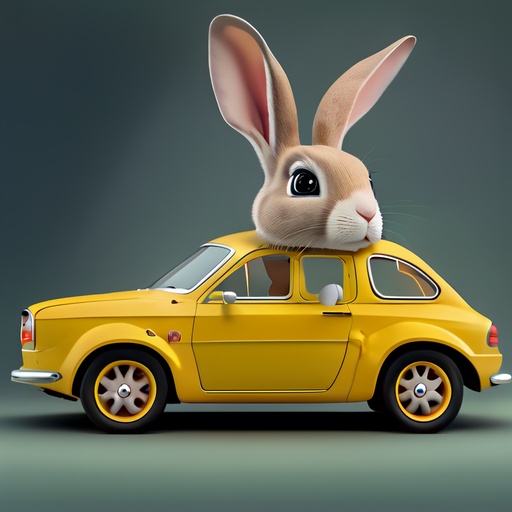} &
    \includegraphics[width=0.1\textwidth]{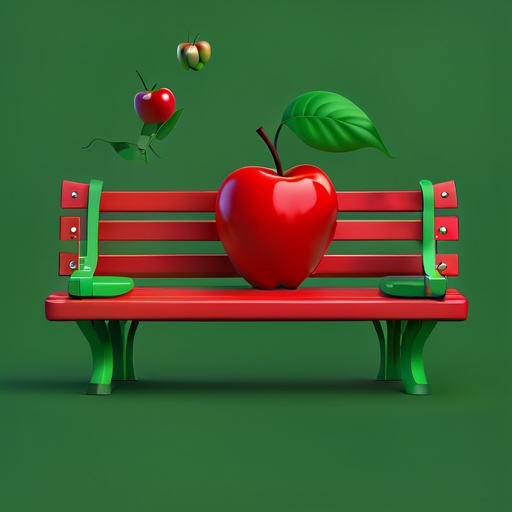} &
    \includegraphics[width=0.1\textwidth]{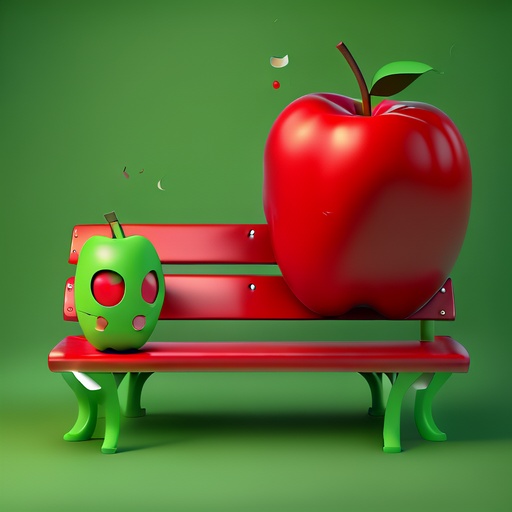} \\

    & \hspace{0.05cm}
    \includegraphics[width=0.1\textwidth]{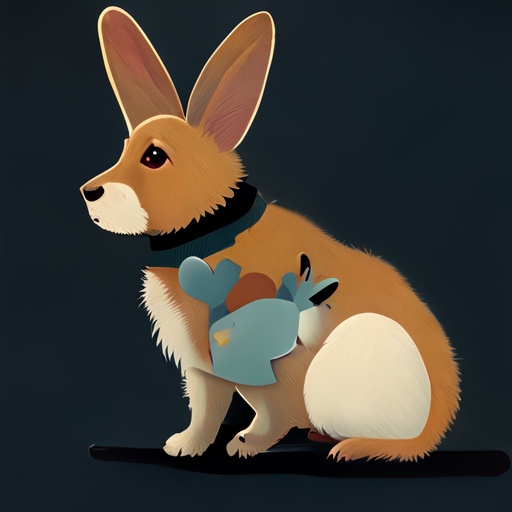} &
    \includegraphics[width=0.1\textwidth]{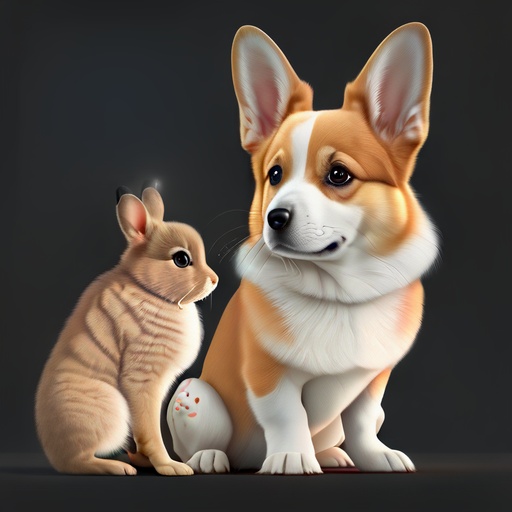} &
    \includegraphics[width=0.1\textwidth]{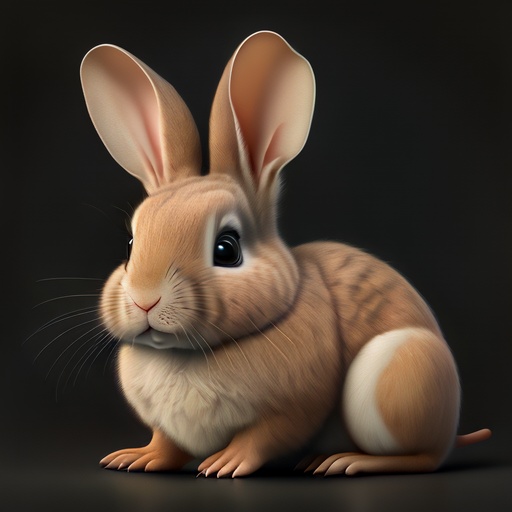} &
    \includegraphics[width=0.1\textwidth]{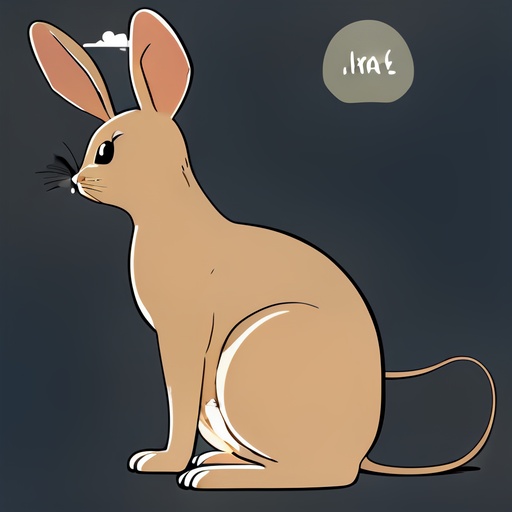} &
    \includegraphics[width=0.1\textwidth]{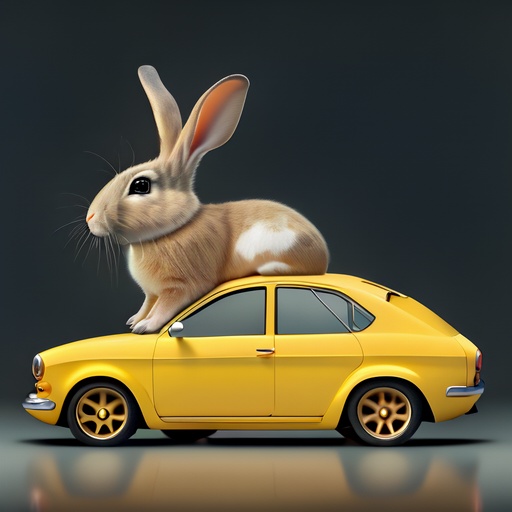} &
    \includegraphics[width=0.1\textwidth]{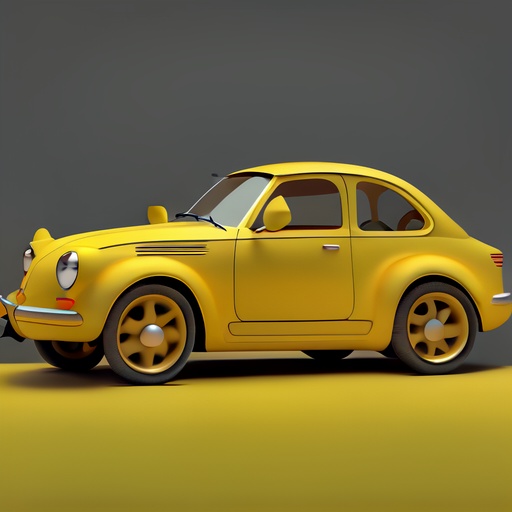} &
    \includegraphics[width=0.1\textwidth]{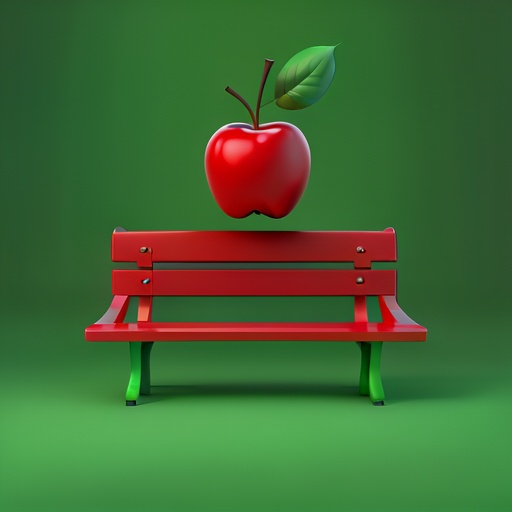} &
    \includegraphics[width=0.1\textwidth]{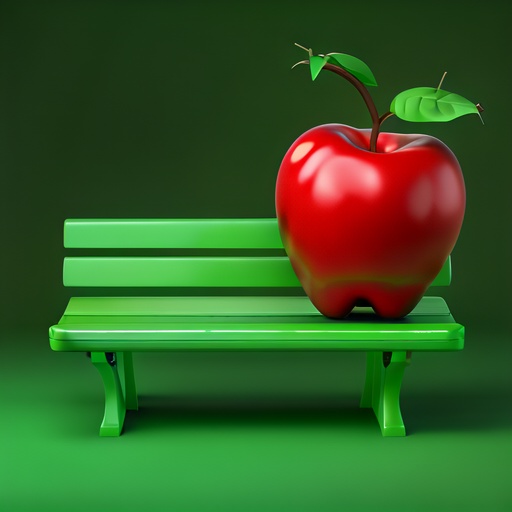} \\ \\ \\

    {\raisebox{0.4in}{\multirow{2}{*}{\rotatebox{90}{\small \textbf{\oursabbr{} (ours)}}}}} & \hspace{0.05cm}
    \includegraphics[width=0.1\textwidth]{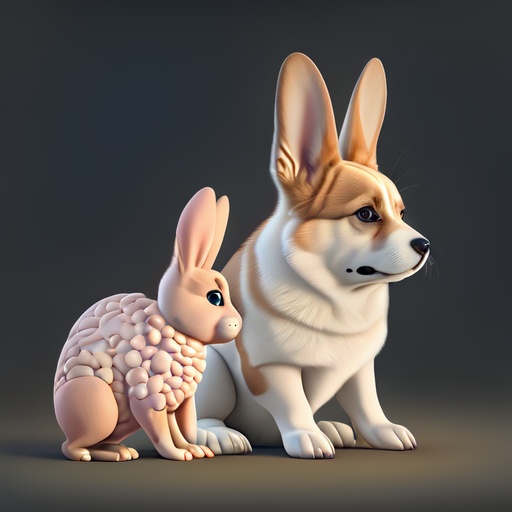} &
    \includegraphics[width=0.1\textwidth]{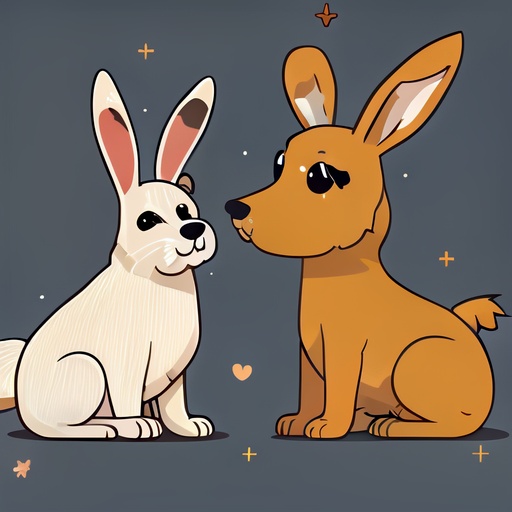} &
    \includegraphics[width=0.1\textwidth]{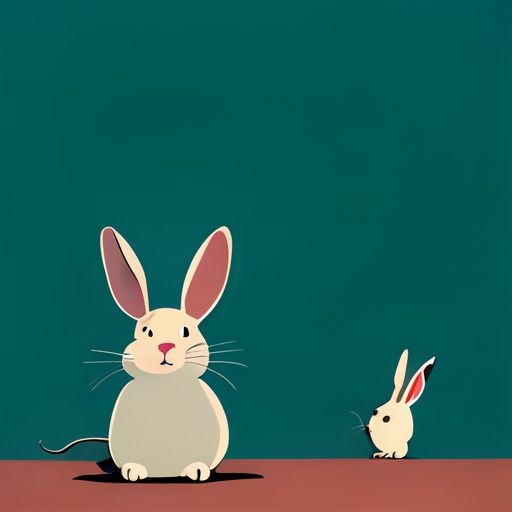} &
    \includegraphics[width=0.1\textwidth]{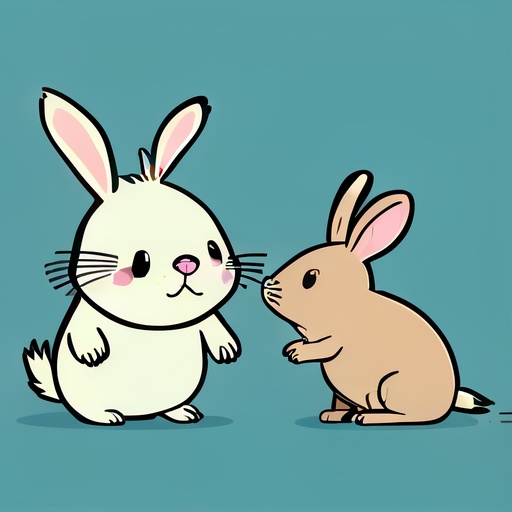} &
    \includegraphics[width=0.1\textwidth]{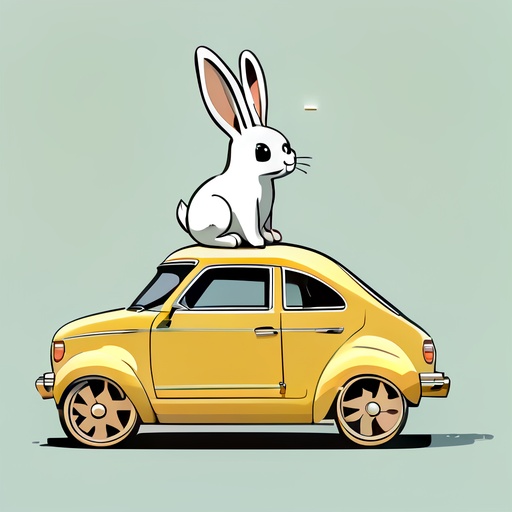} &
    \includegraphics[width=0.1\textwidth]{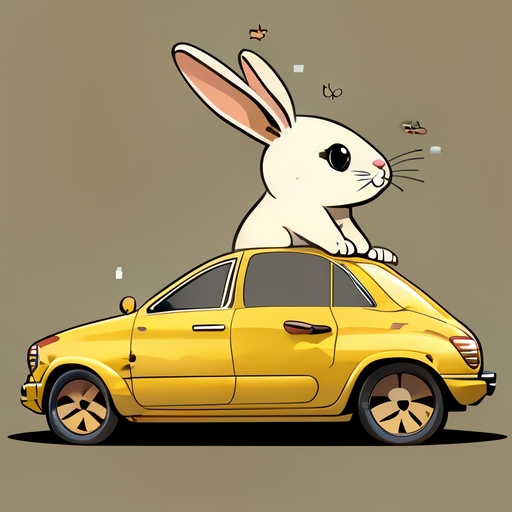} &
    \includegraphics[width=0.1\textwidth]{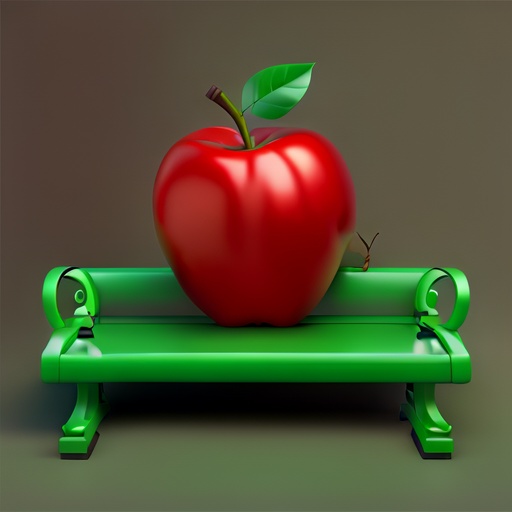} &
    \includegraphics[width=0.1\textwidth]{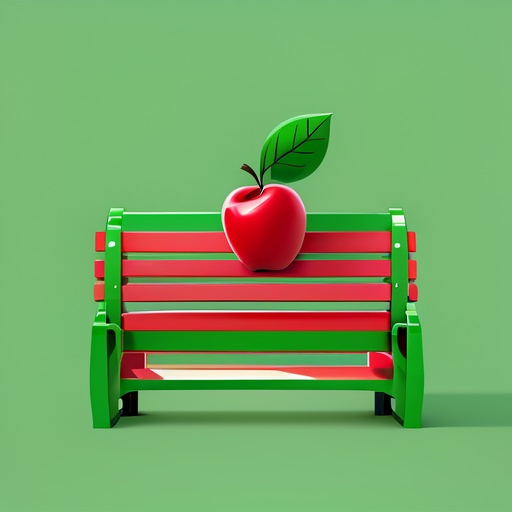} \\

    & \hspace{0.05cm}
    \includegraphics[width=0.1\textwidth]{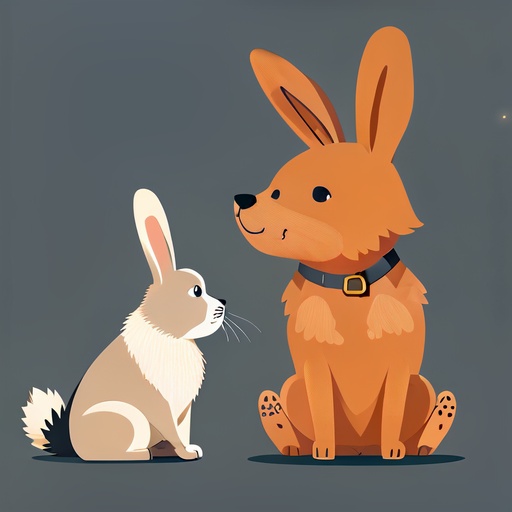} &
    \includegraphics[width=0.1\textwidth]{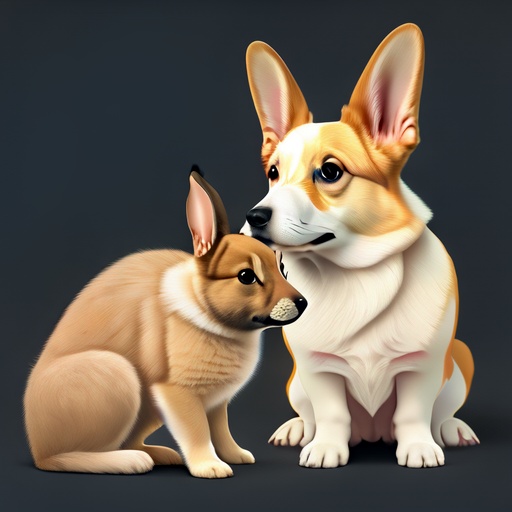} &
    \includegraphics[width=0.1\textwidth]{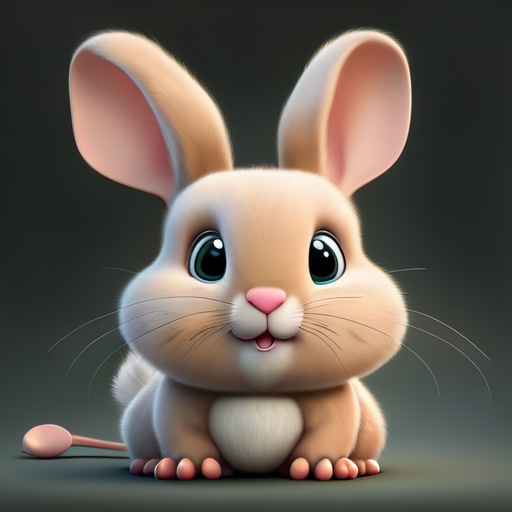} &
    \includegraphics[width=0.1\textwidth]{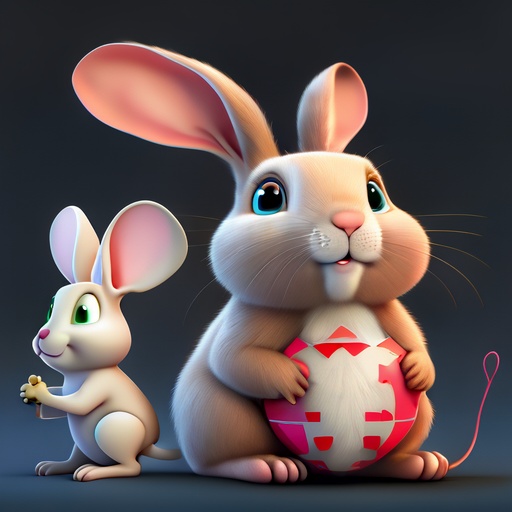} &
    \includegraphics[width=0.1\textwidth]{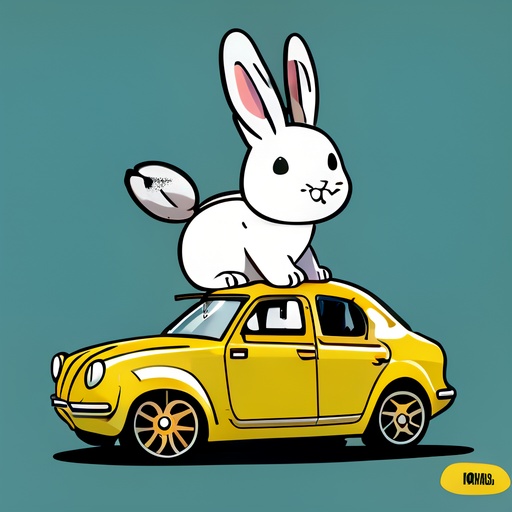} &
    \includegraphics[width=0.1\textwidth]{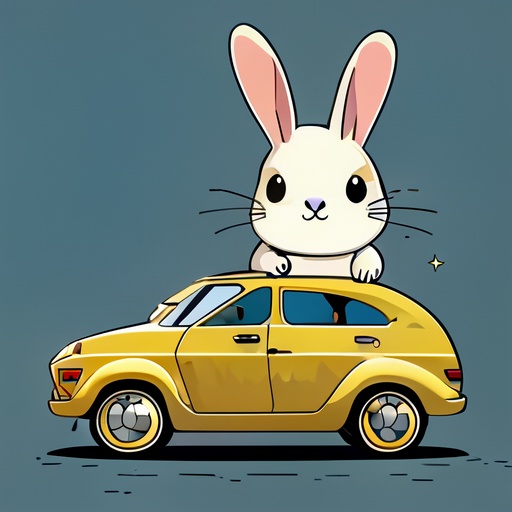} &
    \includegraphics[width=0.1\textwidth]{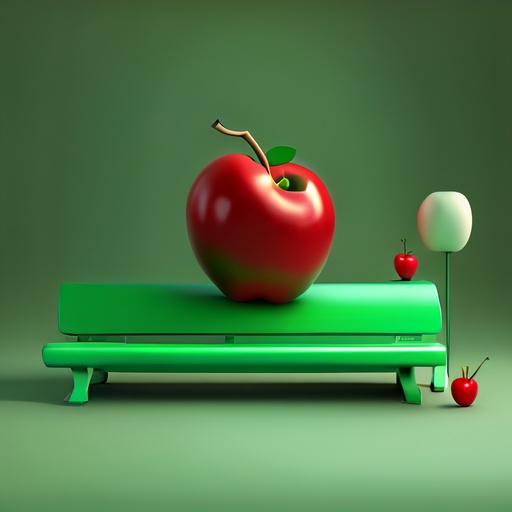} &
    \includegraphics[width=0.1\textwidth]{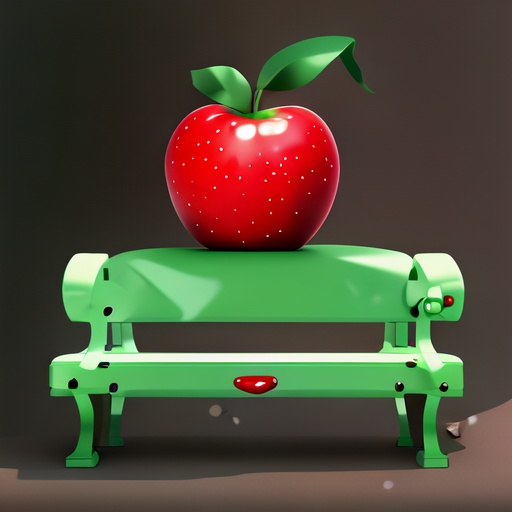} \\

\end{tabular}
    }
    }
\caption{Additional qualitative results on Attend-and-Excite~\citep{attend} dataset (1). The same seeds are applied to each prompt across all methods.}
\label{fig:app_qual_ane_1}
\end{figure*}

\begin{figure*}[!htp]
    \centering
    \setlength{\tabcolsep}{0.5pt}
    \renewcommand{\arraystretch}{0.3}
    \resizebox{1.0\linewidth}{!}{
    {\scriptsize
\begin{tabular}{c c c @{\hspace{0.1cm}} c c @{\hspace{0.1cm}} c c @{\hspace{0.1cm}} c c}

    &
    \multicolumn{2}{c}{\textit{"a bear and a mouse"}} &
    \multicolumn{2}{c}{\textit{"a dog and a lion"}} &
    \multicolumn{2}{c}{\textit{"a monkey and a green bowl"}} &
    \multicolumn{2}{c}{\textit{"a white car and a black bowl"}} \\ \\

    {\raisebox{0.45in}{\multirow{2}{*}{\rotatebox{90}{\small Meissonic (baseline)}}}} & \hspace{0.05cm}
    \includegraphics[width=0.1\textwidth]{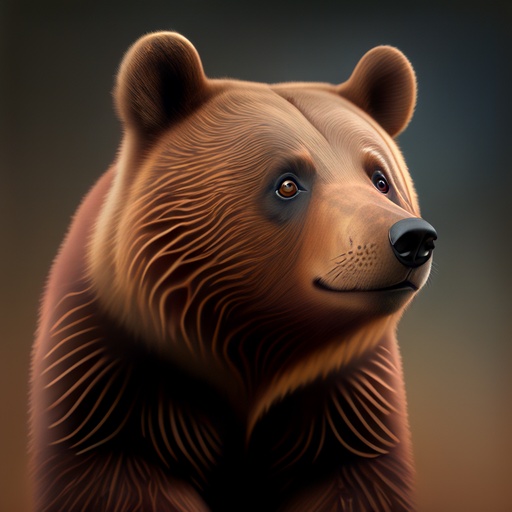} &
    \includegraphics[width=0.1\textwidth]{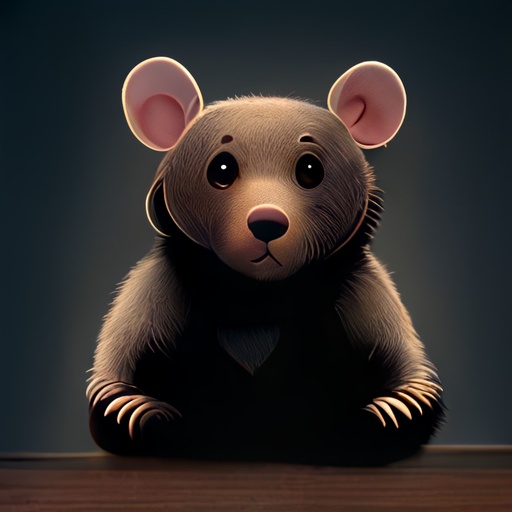} &
    \includegraphics[width=0.1\textwidth]{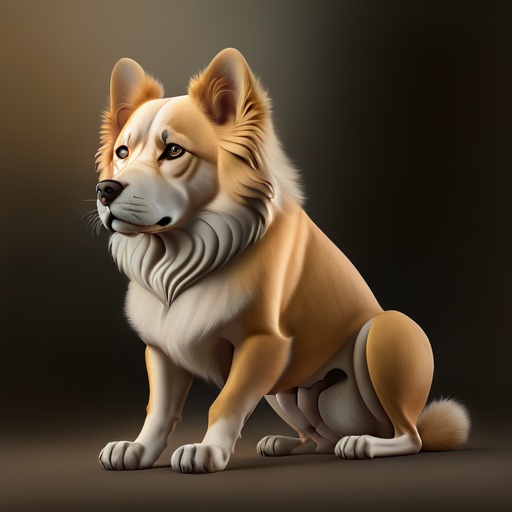} &
    \includegraphics[width=0.1\textwidth]{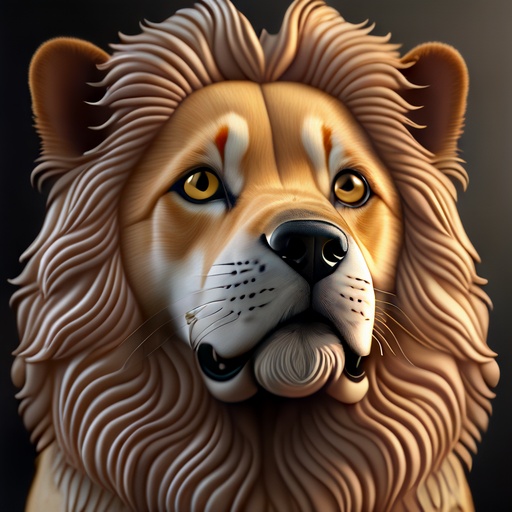} &
    \includegraphics[width=0.1\textwidth]{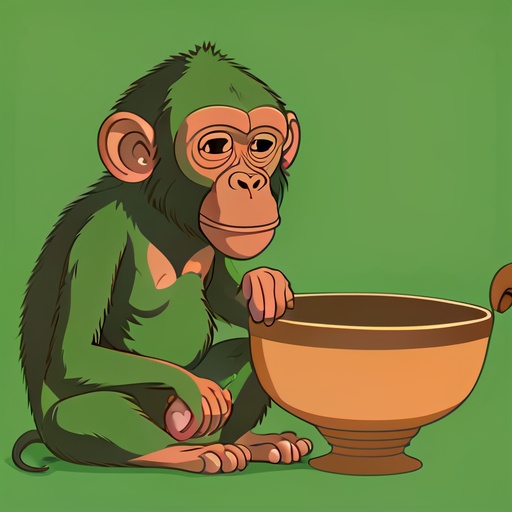} &
    \includegraphics[width=0.1\textwidth]{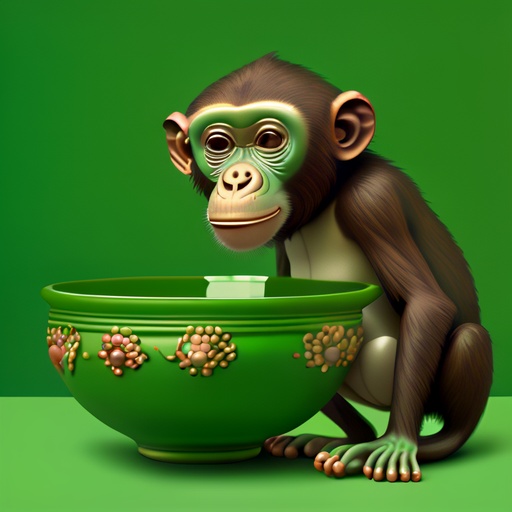} &
    \includegraphics[width=0.1\textwidth]{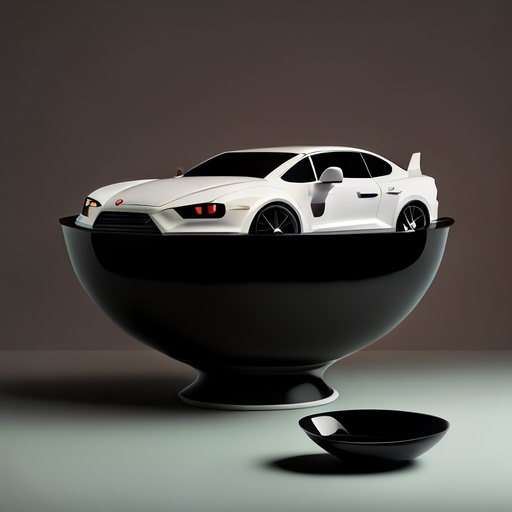} &
    \includegraphics[width=0.1\textwidth]{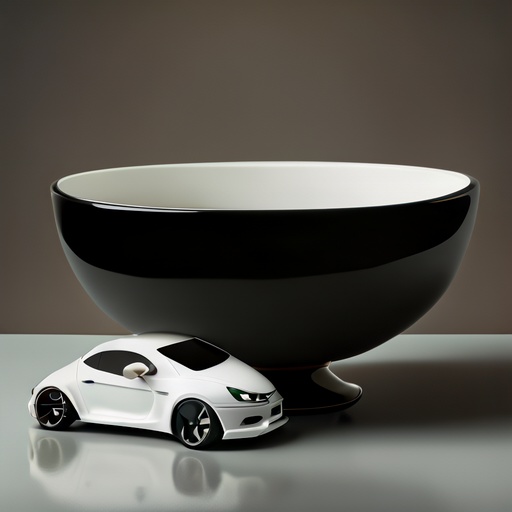} \\

    & \hspace{0.05cm}
    \includegraphics[width=0.1\textwidth]{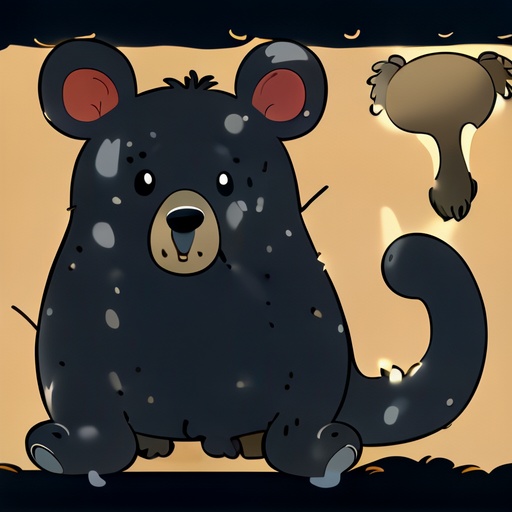} &
    \includegraphics[width=0.1\textwidth]{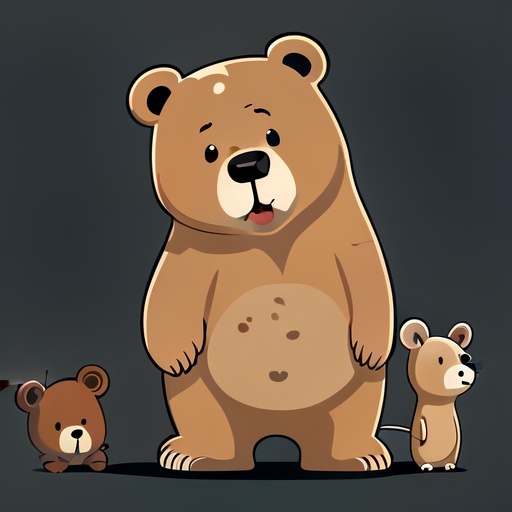} &
    \includegraphics[width=0.1\textwidth]{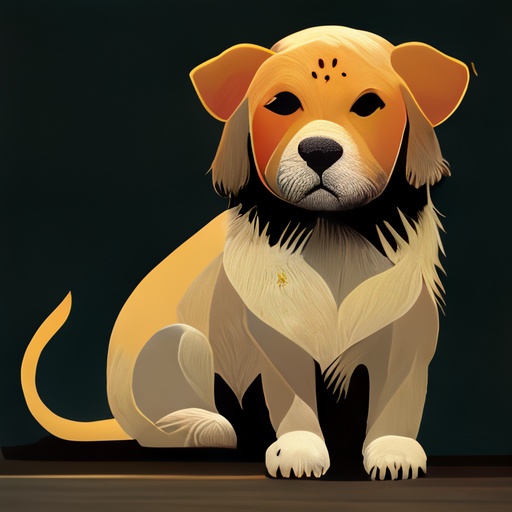} &
    \includegraphics[width=0.1\textwidth]{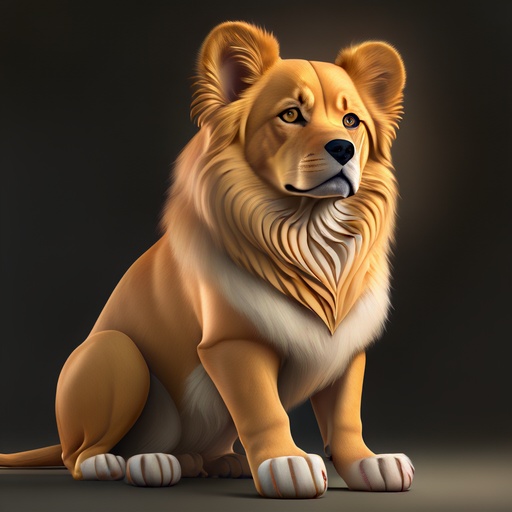} &
    \includegraphics[width=0.1\textwidth]{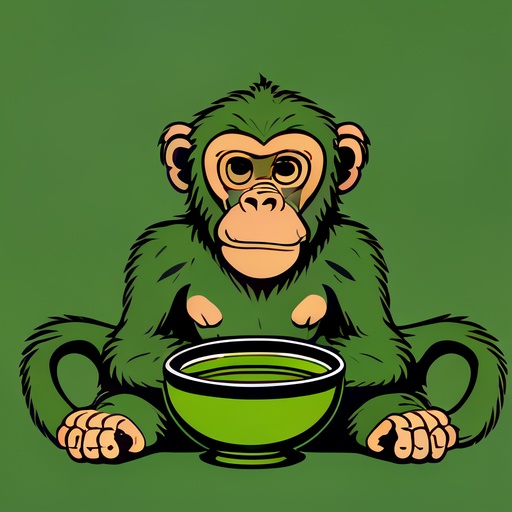} &
    \includegraphics[width=0.1\textwidth]{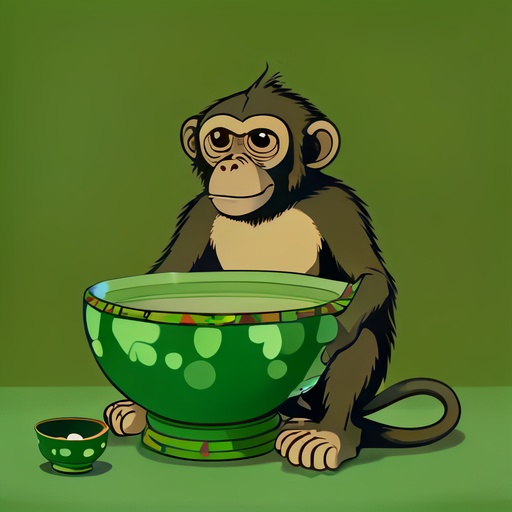} &
    \includegraphics[width=0.1\textwidth]{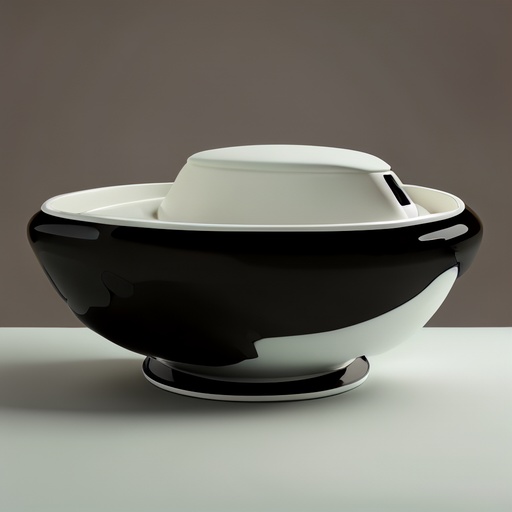} &
    \includegraphics[width=0.1\textwidth]{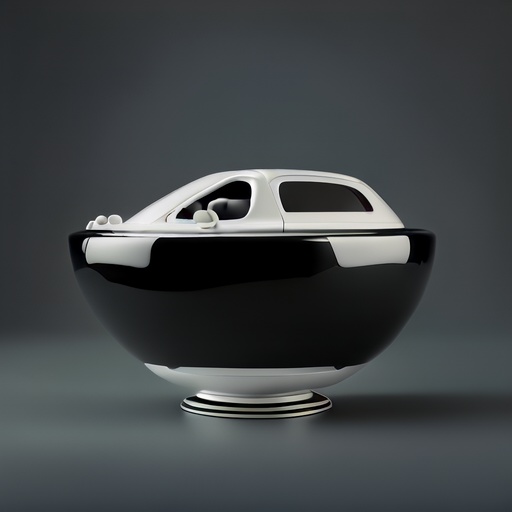} \\ \\ \\

    {\raisebox{0.4in}{\multirow{2}{*}{\rotatebox{90}{\small \textbf{\oursabbr{} (ours)}}}}} & \hspace{0.05cm}
    \includegraphics[width=0.1\textwidth]{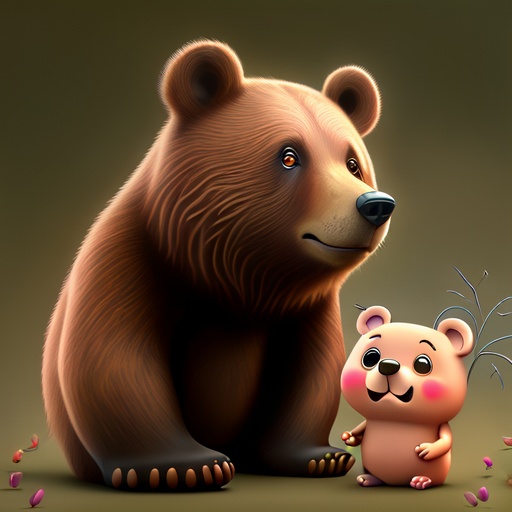} &
    \includegraphics[width=0.1\textwidth]{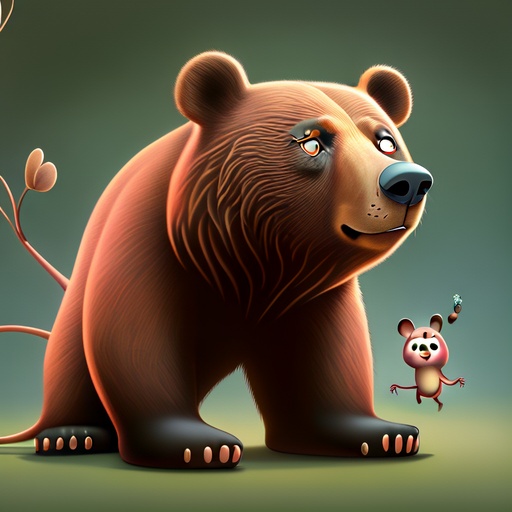} &
    \includegraphics[width=0.1\textwidth]{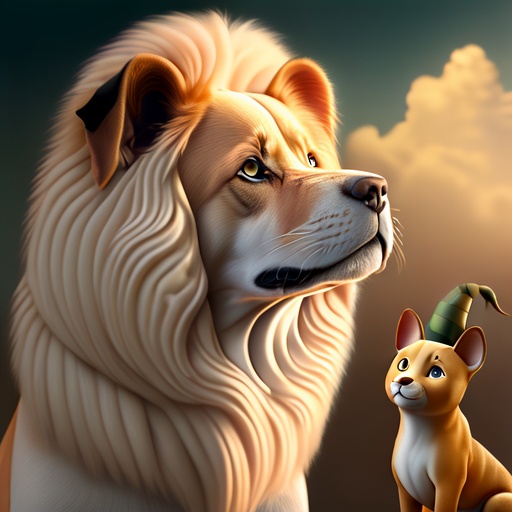} &
    \includegraphics[width=0.1\textwidth]{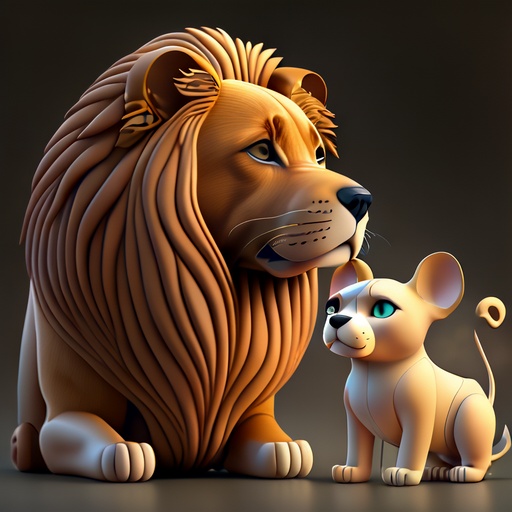} &
    \includegraphics[width=0.1\textwidth]{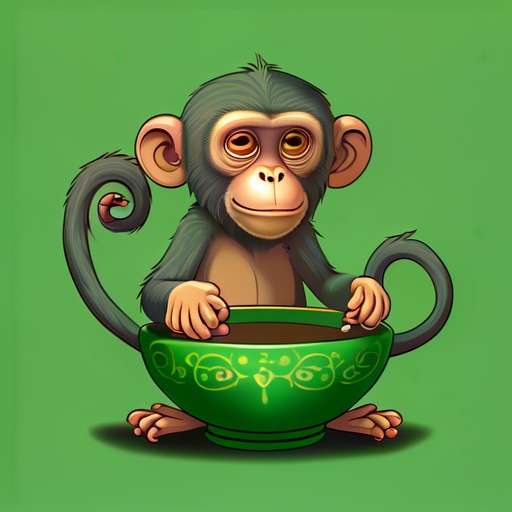} &
    \includegraphics[width=0.1\textwidth]{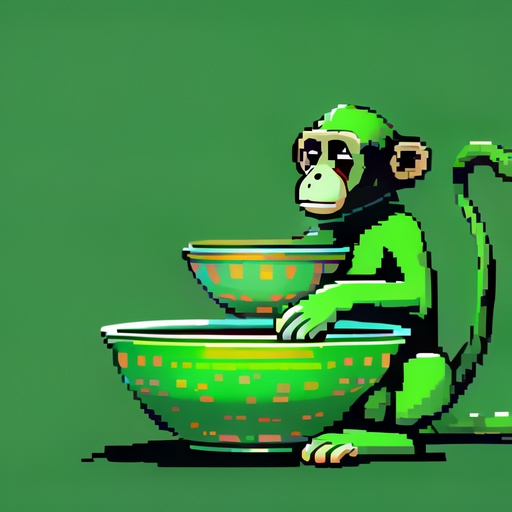} &
    \includegraphics[width=0.1\textwidth]{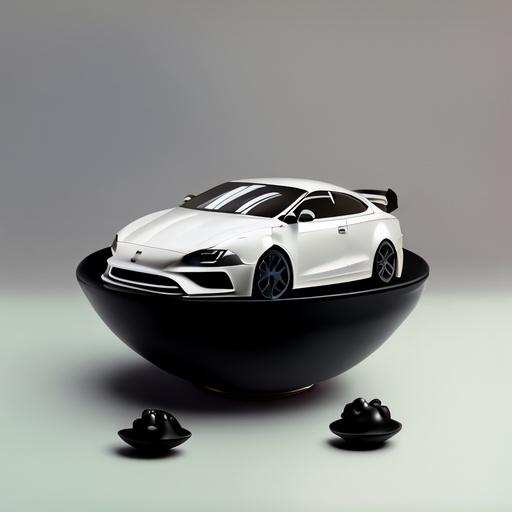} &
    \includegraphics[width=0.1\textwidth]{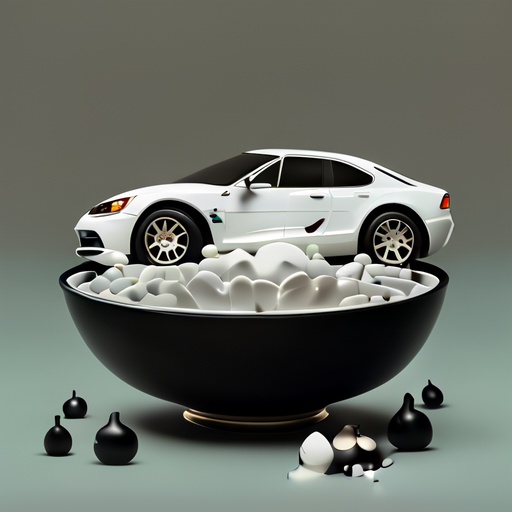} \\

    & \hspace{0.05cm}
    \includegraphics[width=0.1\textwidth]{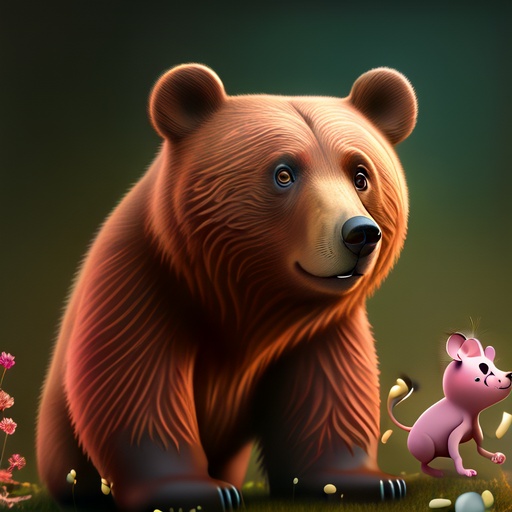} &
    \includegraphics[width=0.1\textwidth]{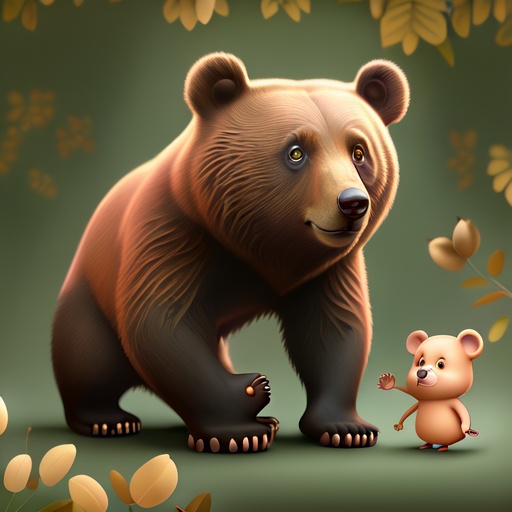} &
    \includegraphics[width=0.1\textwidth]{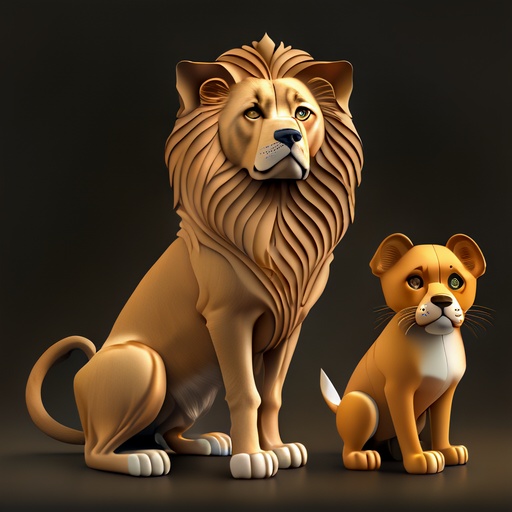} &
    \includegraphics[width=0.1\textwidth]{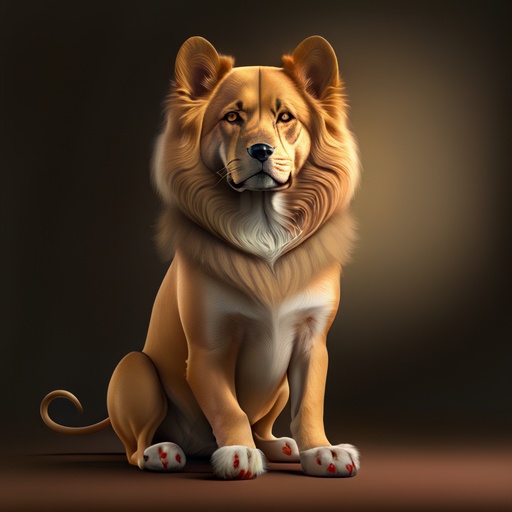} &
    \includegraphics[width=0.1\textwidth]{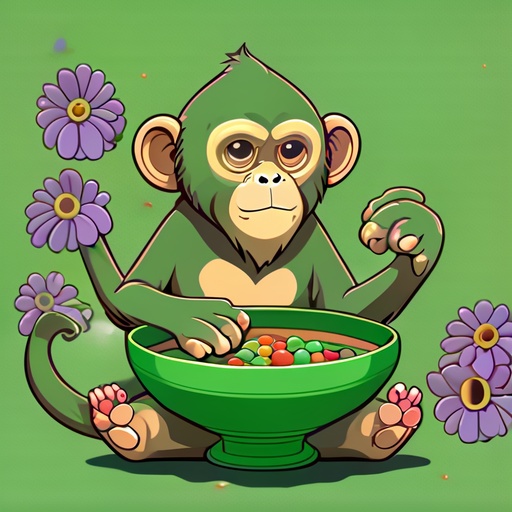} &
    \includegraphics[width=0.1\textwidth]{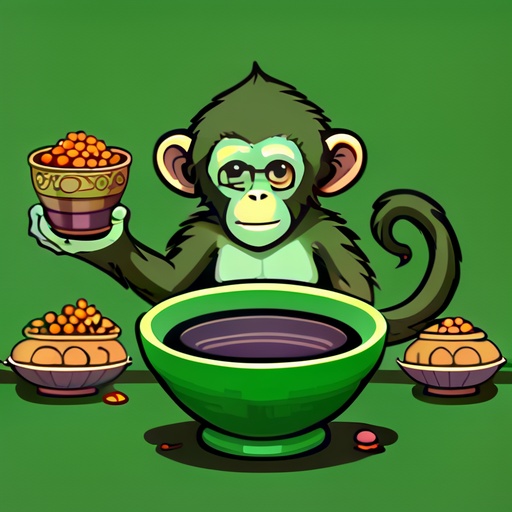} &
    \includegraphics[width=0.1\textwidth]{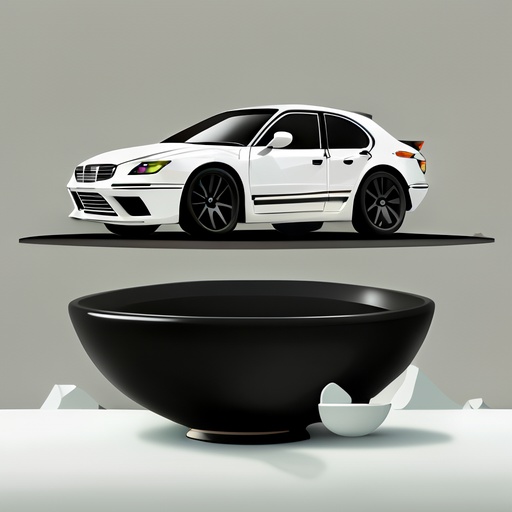} &
    \includegraphics[width=0.1\textwidth]{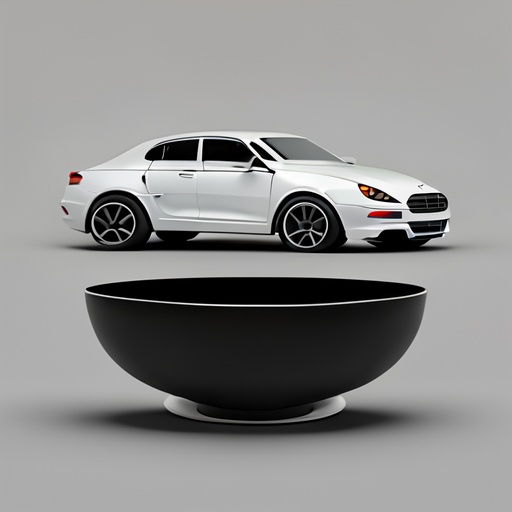} \\ \\ \\ \\ \\ \\ \\ \\ \\

    &
    \multicolumn{2}{c}{\textit{"a horse and a monkey"}} &
    \multicolumn{2}{c}{\textit{"a bear and a rabbit"}} &
    \multicolumn{2}{c}{\textit{"a mouse and a red bench"}} &
    \multicolumn{2}{c}{\textit{"a brown bowl and a green clock"}} \\ \\

    {\raisebox{0.45in}{\multirow{2}{*}{\rotatebox{90}{\small Meissonic (baseline)}}}} & \hspace{0.05cm}
    \includegraphics[width=0.1\textwidth]{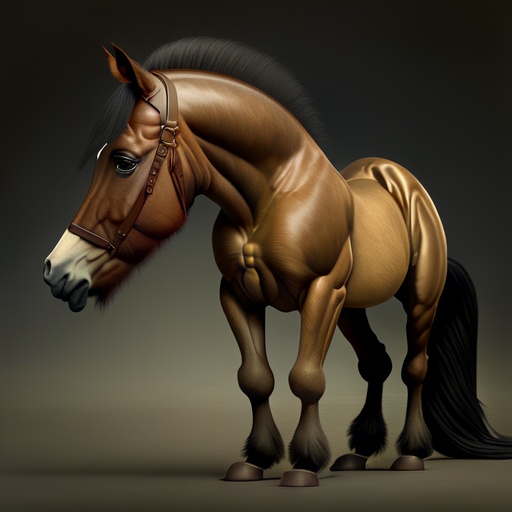} &
    \includegraphics[width=0.1\textwidth]{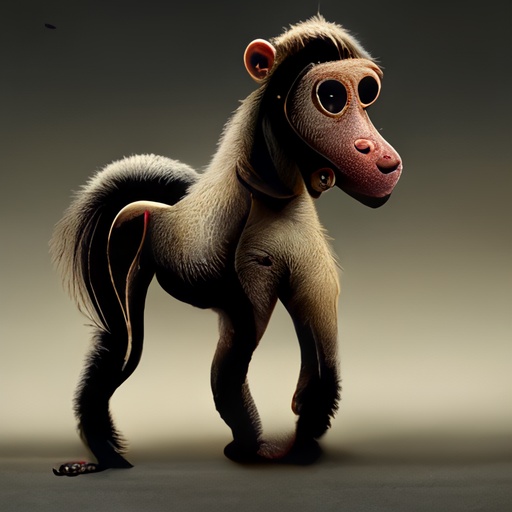} &
    \includegraphics[width=0.1\textwidth]{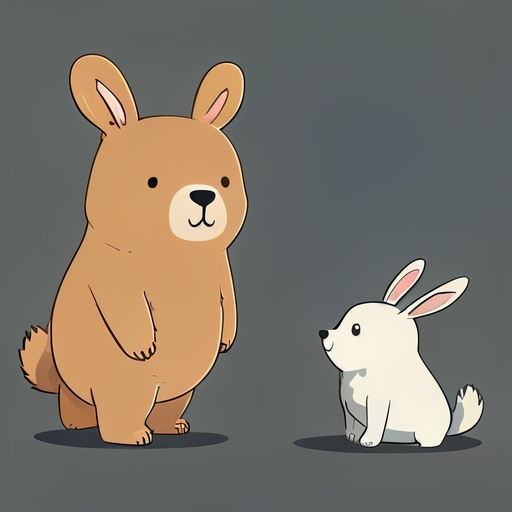} &
    \includegraphics[width=0.1\textwidth]{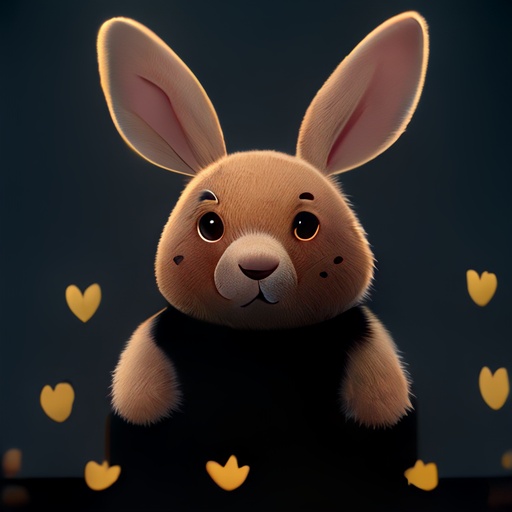} &
    \includegraphics[width=0.1\textwidth]{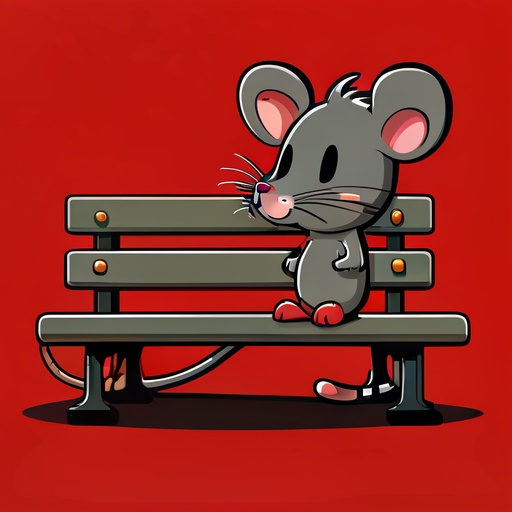} &
    \includegraphics[width=0.1\textwidth]{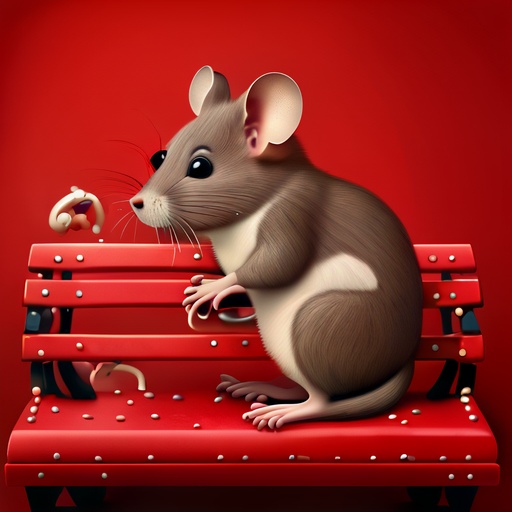} &
    \includegraphics[width=0.1\textwidth]{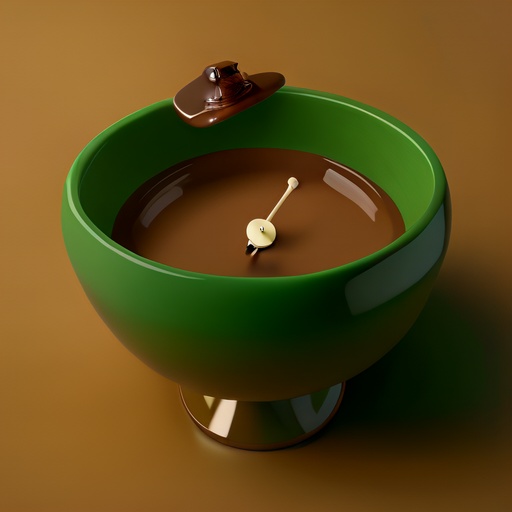} &
    \includegraphics[width=0.1\textwidth]{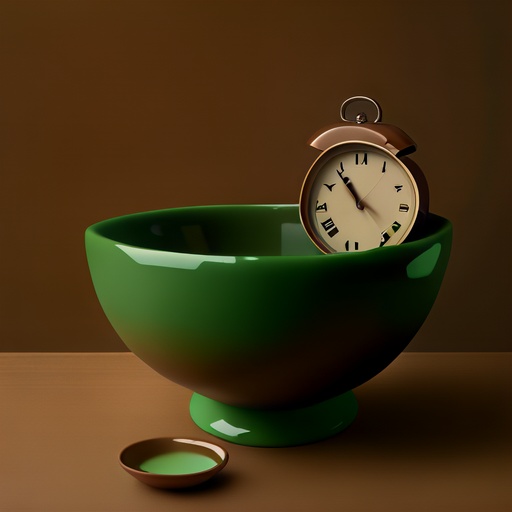} \\

    & \hspace{0.05cm}
    \includegraphics[width=0.1\textwidth]{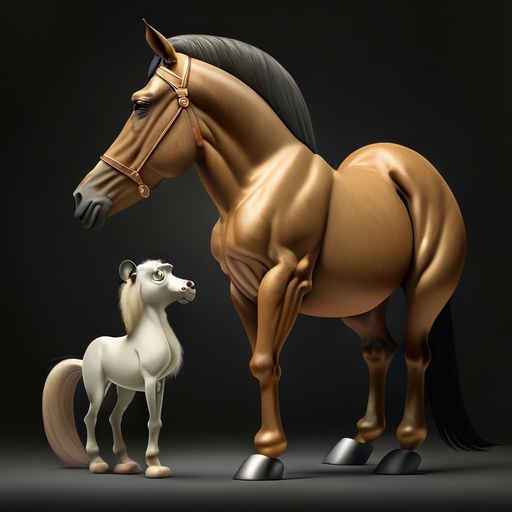} &
    \includegraphics[width=0.1\textwidth]{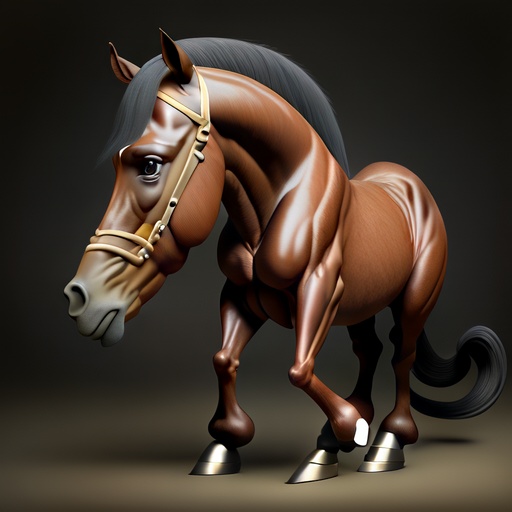} &
    \includegraphics[width=0.1\textwidth]{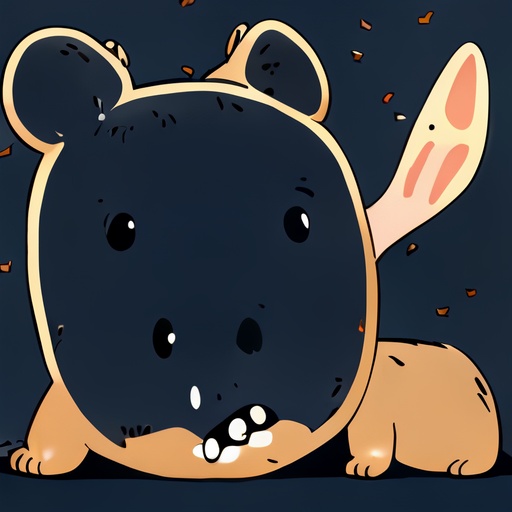} &
    \includegraphics[width=0.1\textwidth]{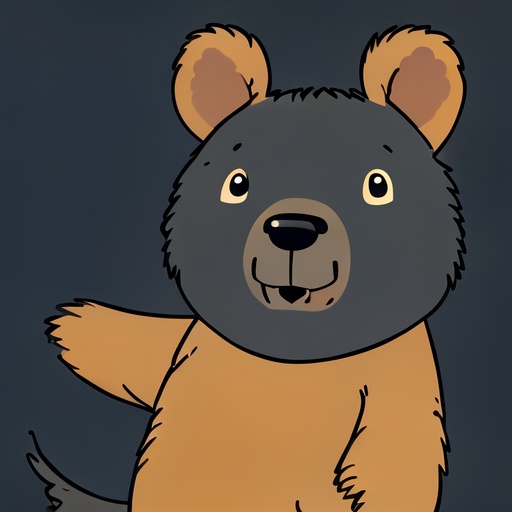} &
    \includegraphics[width=0.1\textwidth]{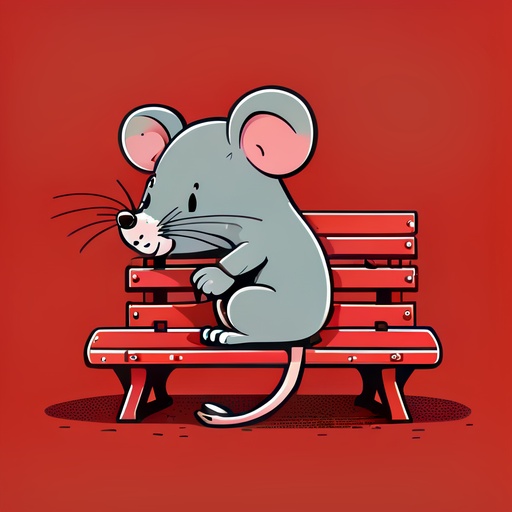} &
    \includegraphics[width=0.1\textwidth]{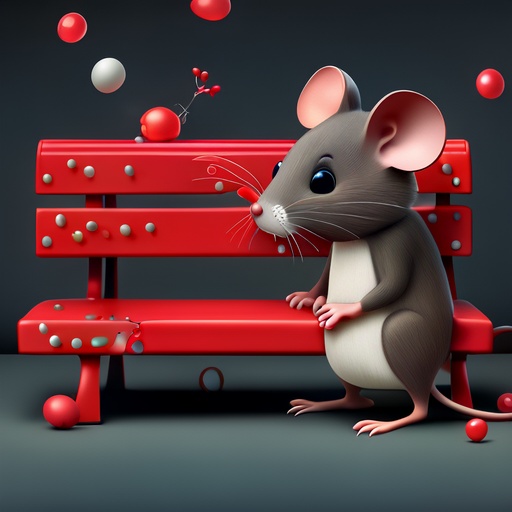} &
    \includegraphics[width=0.1\textwidth]{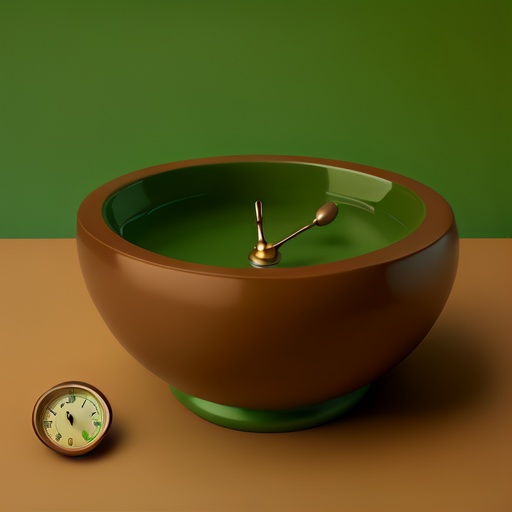} &
    \includegraphics[width=0.1\textwidth]{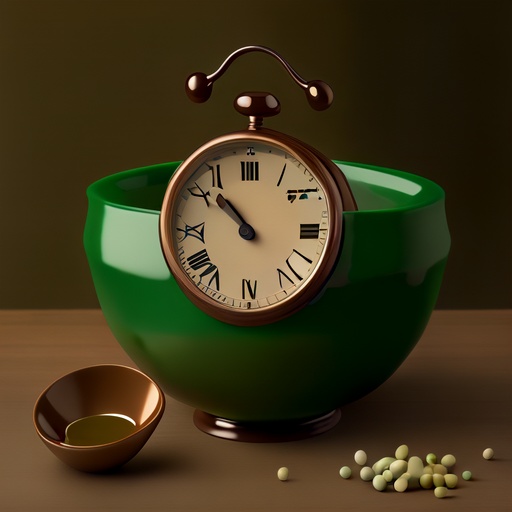} \\ \\ \\

    {\raisebox{0.4in}{\multirow{2}{*}{\rotatebox{90}{\small \textbf{\oursabbr{} (ours)}}}}} & \hspace{0.05cm}
    \includegraphics[width=0.1\textwidth]{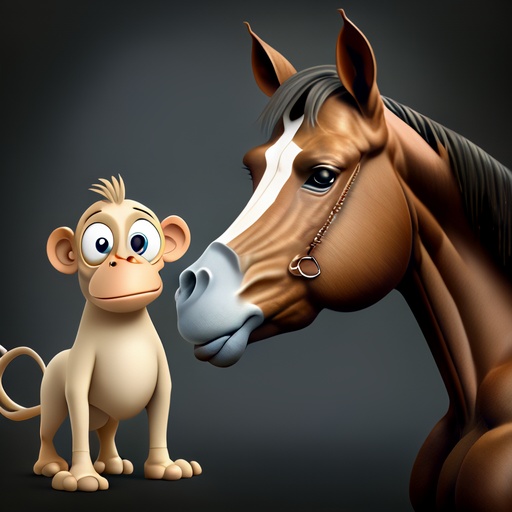} &
    \includegraphics[width=0.1\textwidth]{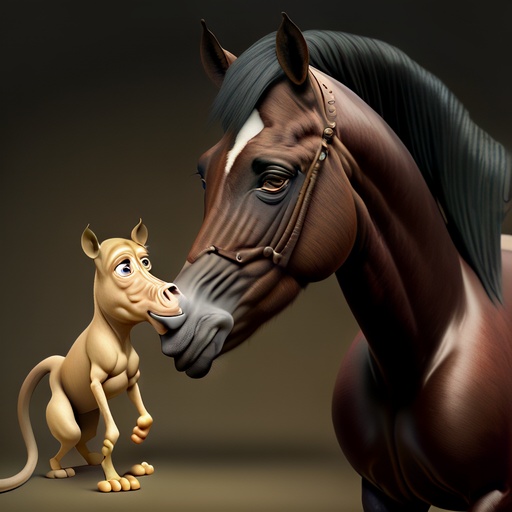} &
    \includegraphics[width=0.1\textwidth]{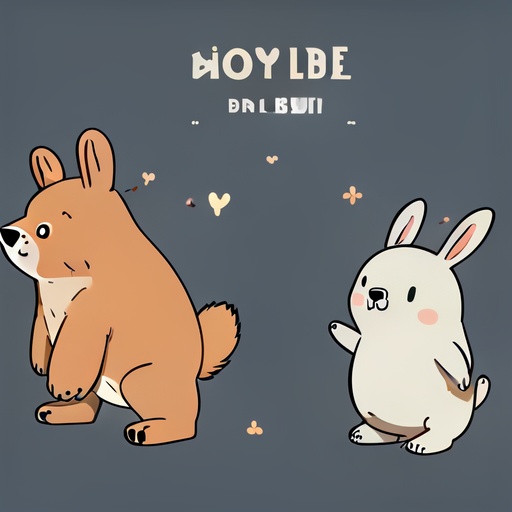} &
    \includegraphics[width=0.1\textwidth]{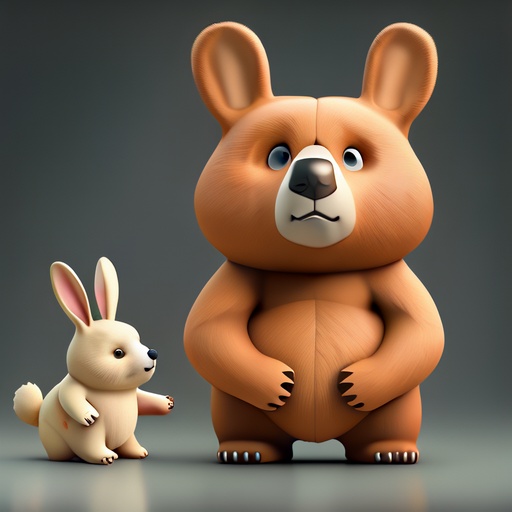} &
    \includegraphics[width=0.1\textwidth]{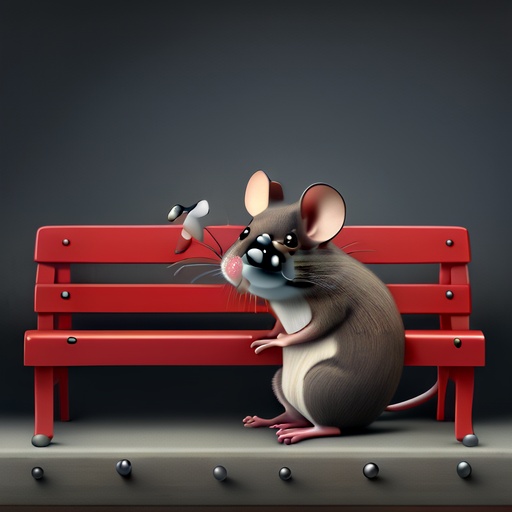} &
    \includegraphics[width=0.1\textwidth]{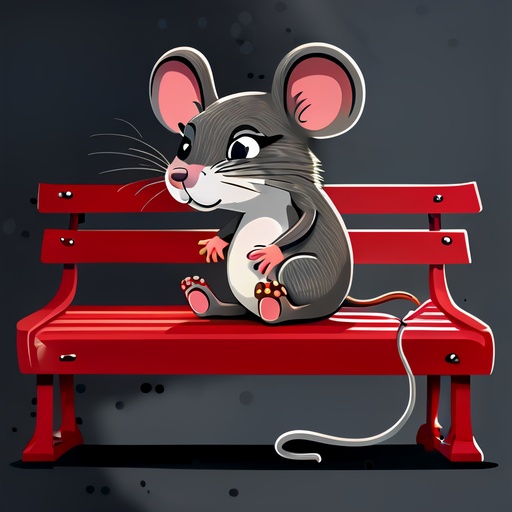} &
    \includegraphics[width=0.1\textwidth]{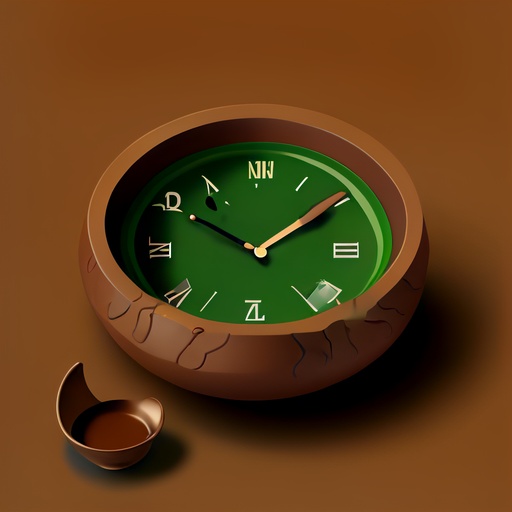} &
    \includegraphics[width=0.1\textwidth]{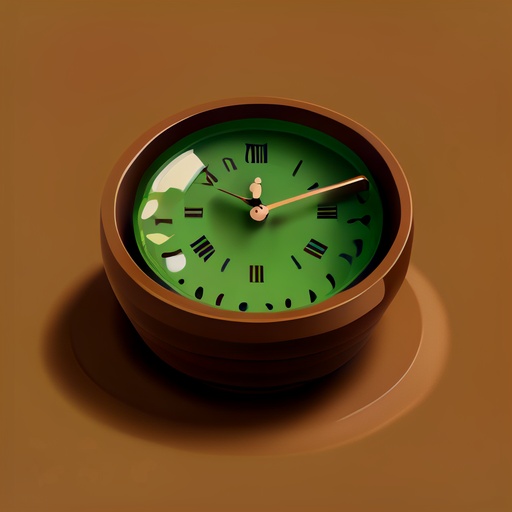} \\

    & \hspace{0.05cm}
    \includegraphics[width=0.1\textwidth]{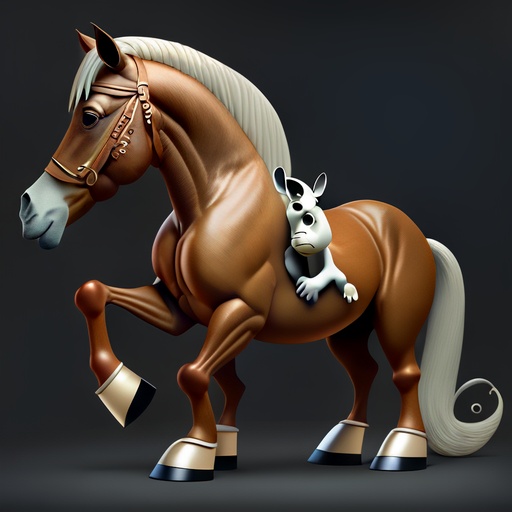} &
    \includegraphics[width=0.1\textwidth]{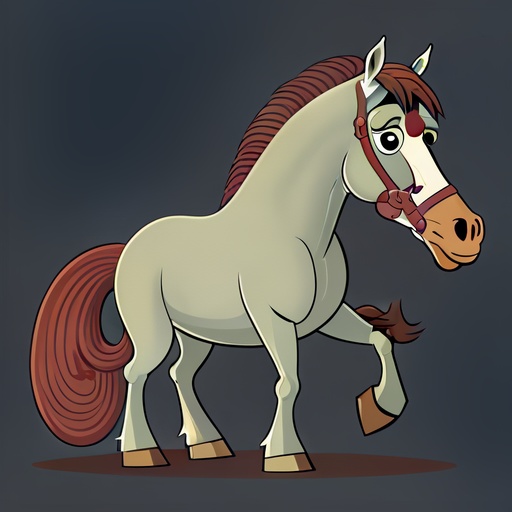} &
    \includegraphics[width=0.1\textwidth]{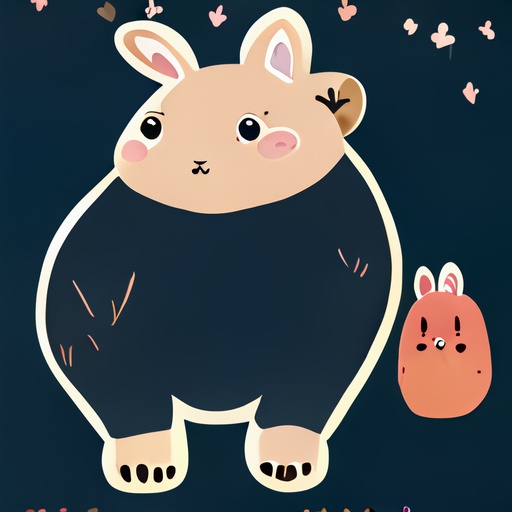} &
    \includegraphics[width=0.1\textwidth]{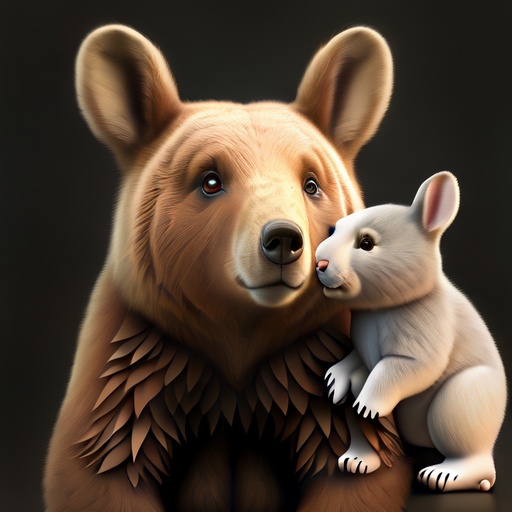} &
    \includegraphics[width=0.1\textwidth]{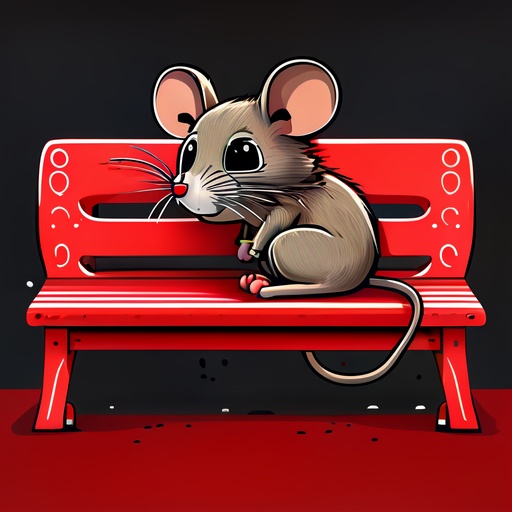} &
    \includegraphics[width=0.1\textwidth]{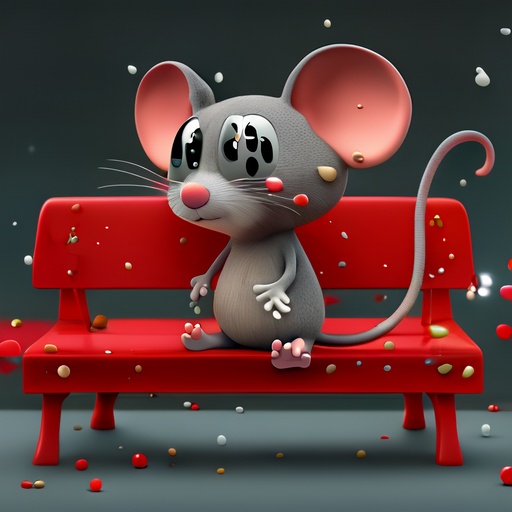} &
    \includegraphics[width=0.1\textwidth]{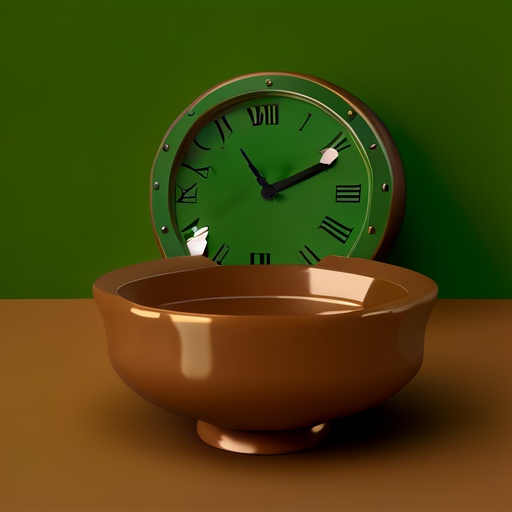} &
    \includegraphics[width=0.1\textwidth]{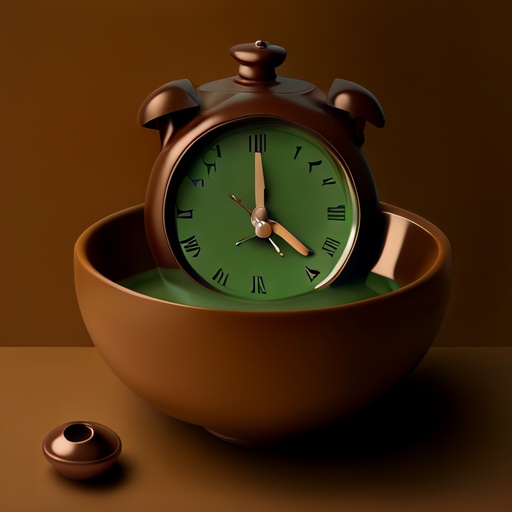} \\

\end{tabular}
    }
    }
\caption{Additional qualitative results on Attend-and-Excite~\citep{attend} dataset (2). The same seeds are applied to each prompt across all methods.}
\label{fig:app_qual_ane_2}
\end{figure*}

Additional qualitative results on the Attend-and-Excite dataset are provided in \cref{fig:app_qual_ane_1,fig:app_qual_ane_2}. The same seeds are applied to each prompt across all methods.

\subsection{SSD Dataset}

\begin{figure*}[!htp]
    \centering
    \setlength{\tabcolsep}{0.5pt}
    \renewcommand{\arraystretch}{0.3}
    \resizebox{1.0\linewidth}{!}{
    {\footnotesize
\begin{tabular}{c c c c c @{\hspace{0.1cm}} c c c c}

    & \multicolumn{4}{c}{\textit{“a beagle and a collie”}} & \multicolumn{4}{c}{\textit{“a eagle and a condor”}} \\ \\

     & \hspace{0.1cm}
    \includegraphics[width=0.1\textwidth]{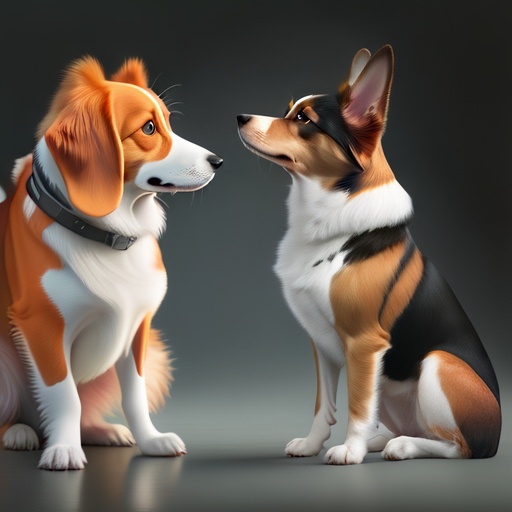} &     \includegraphics[width=0.1\textwidth]{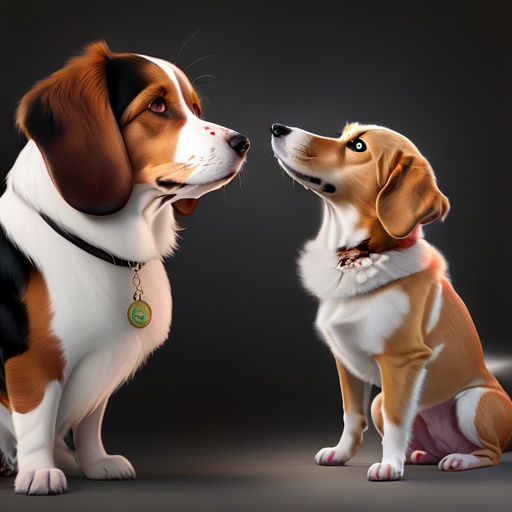} &     \includegraphics[width=0.1\textwidth]{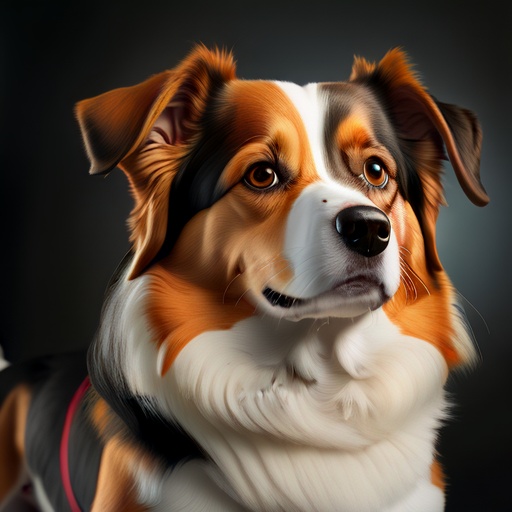} &     \includegraphics[width=0.1\textwidth]{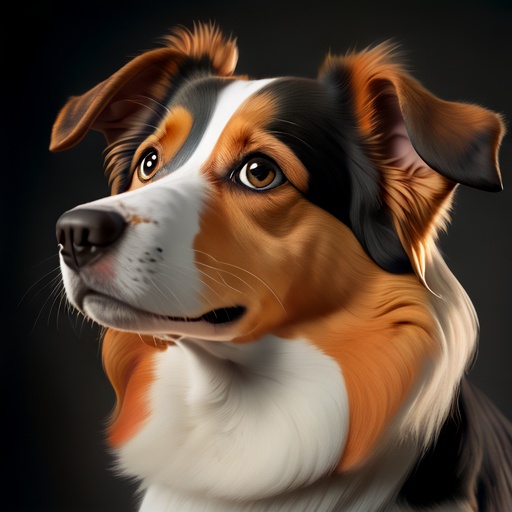} &     \includegraphics[width=0.1\textwidth]{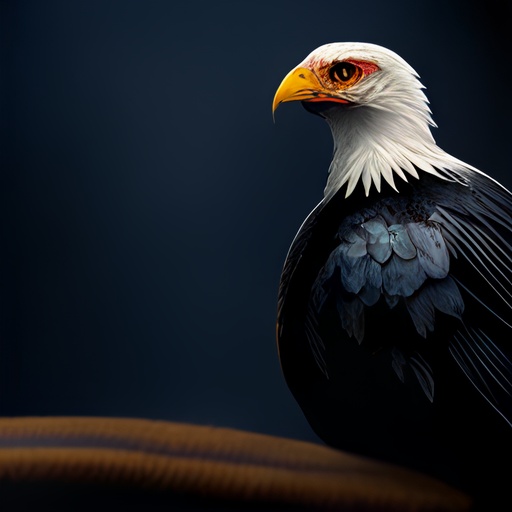} &     \includegraphics[width=0.1\textwidth]{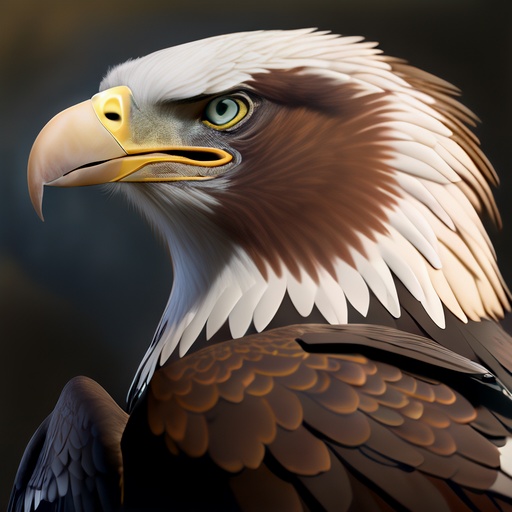} &     \includegraphics[width=0.1\textwidth]{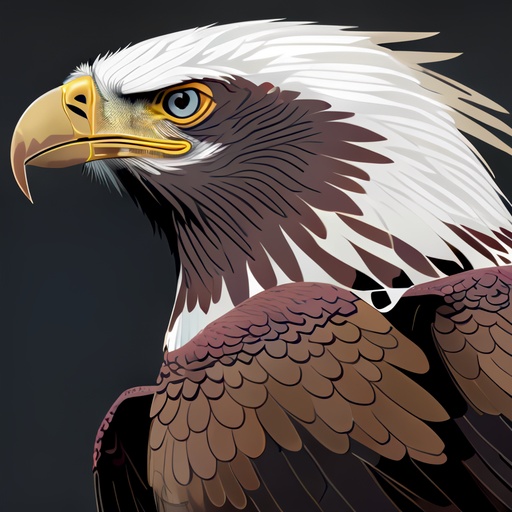} &     \includegraphics[width=0.1\textwidth]{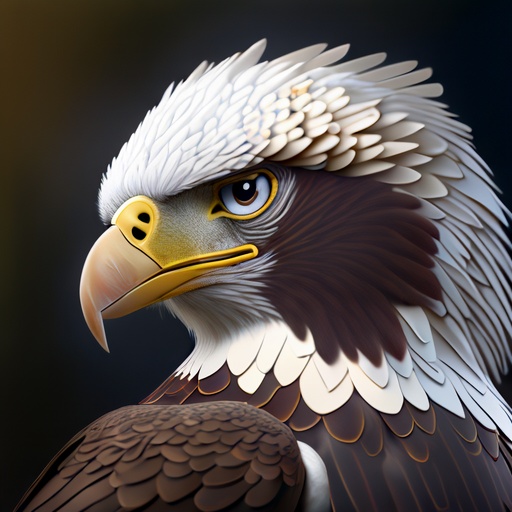} \\
{\raisebox{0.45in}{\multirow{2}{*}{\rotatebox{90}{\normalsize Meissonic (baseline)}}}}& \hspace{0.1cm}
    \includegraphics[width=0.1\textwidth]{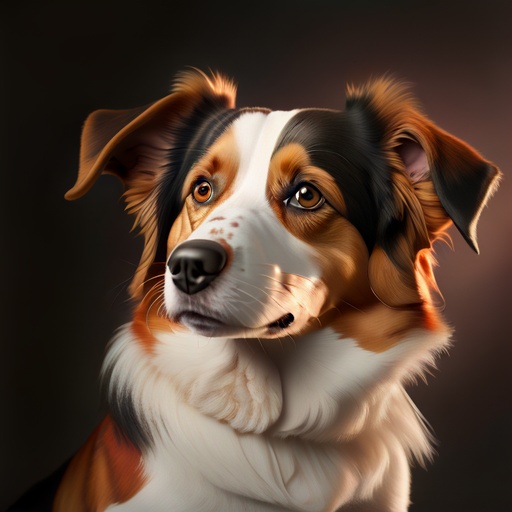} &     \includegraphics[width=0.1\textwidth]{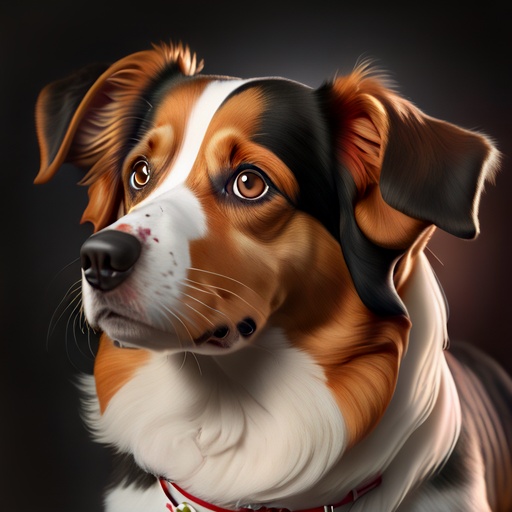} &     \includegraphics[width=0.1\textwidth]{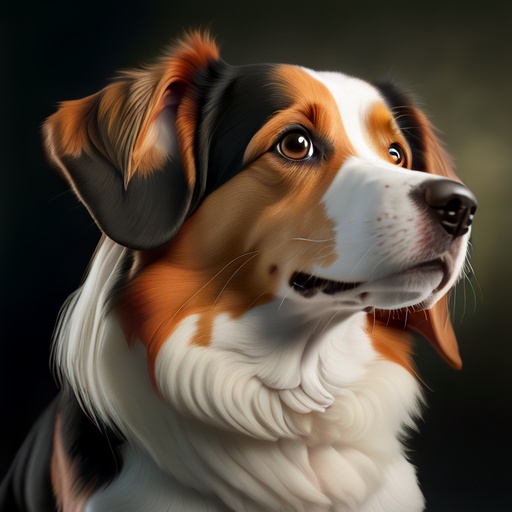} &     \includegraphics[width=0.1\textwidth]{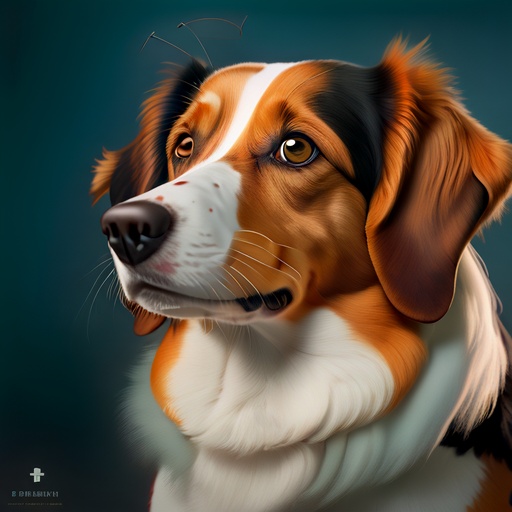} &     \includegraphics[width=0.1\textwidth]{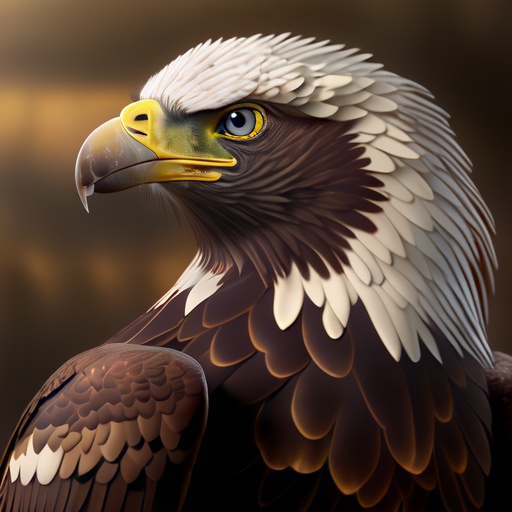} &     \includegraphics[width=0.1\textwidth]{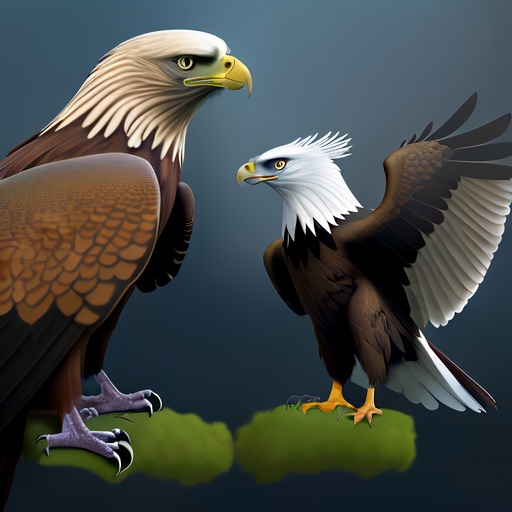} &     \includegraphics[width=0.1\textwidth]{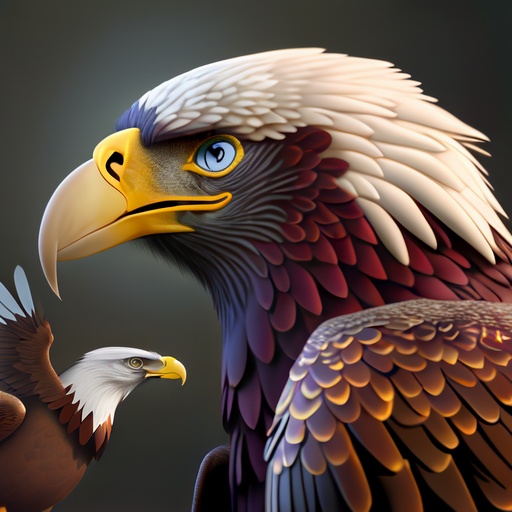} &     \includegraphics[width=0.1\textwidth]{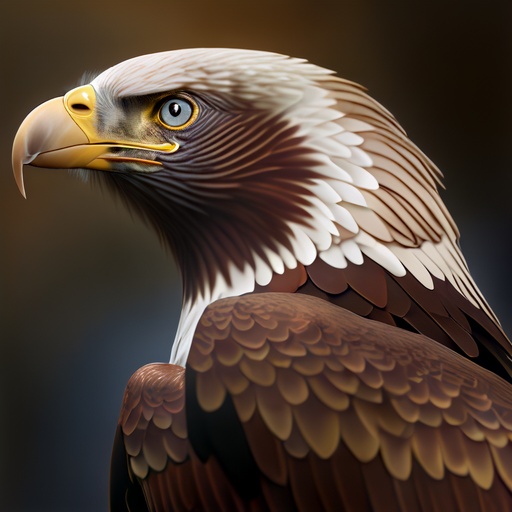} \\
 & \hspace{0.1cm}
    \includegraphics[width=0.1\textwidth]{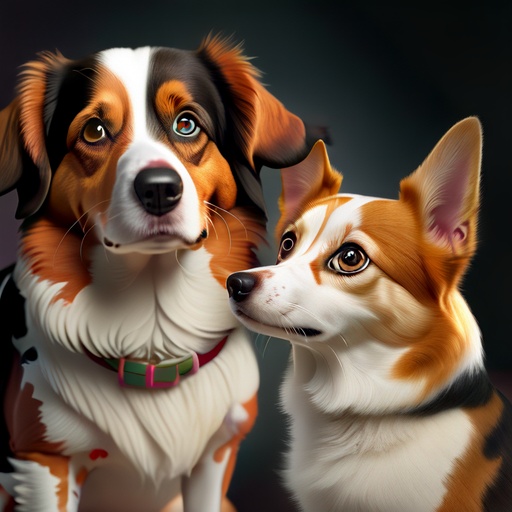} &     \includegraphics[width=0.1\textwidth]{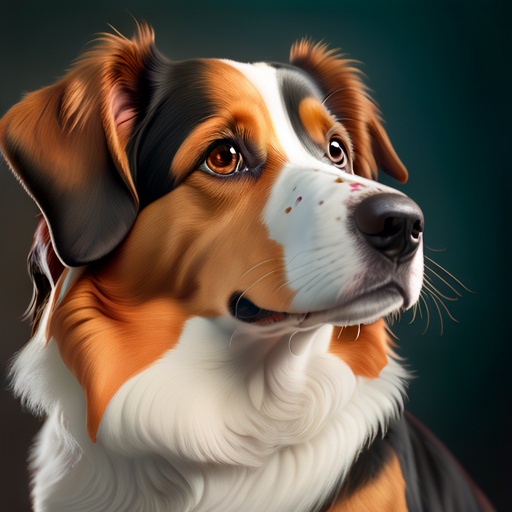} &     \includegraphics[width=0.1\textwidth]{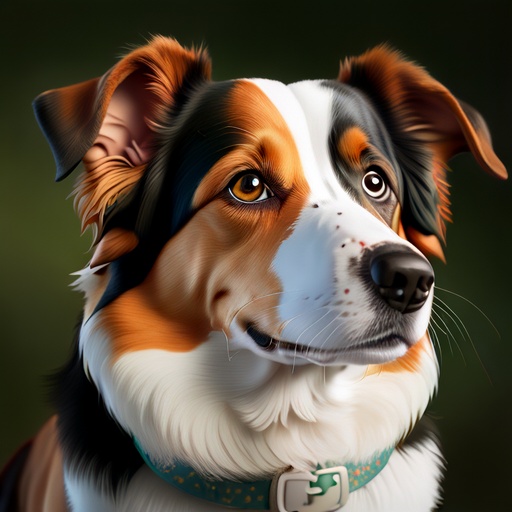} &     \includegraphics[width=0.1\textwidth]{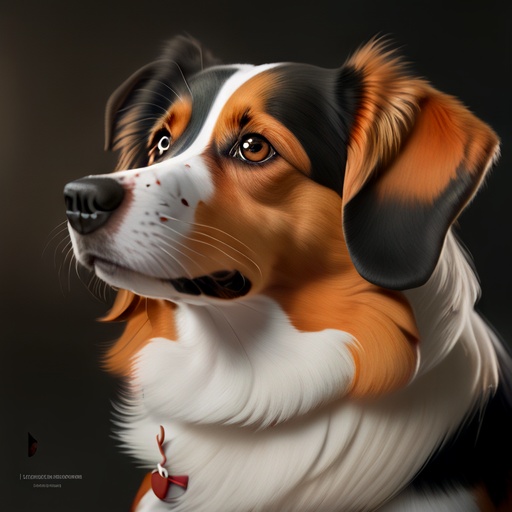} &     \includegraphics[width=0.1\textwidth]{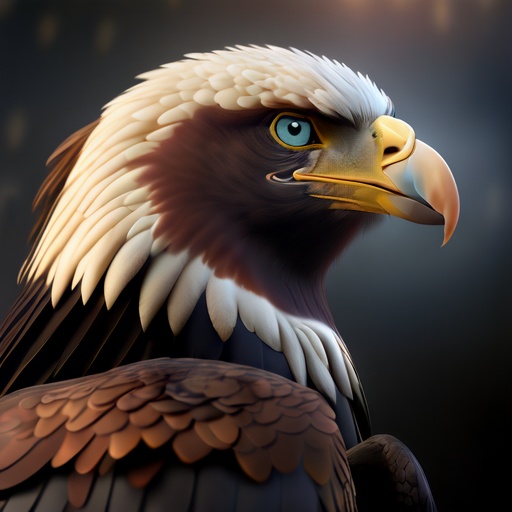} &     \includegraphics[width=0.1\textwidth]{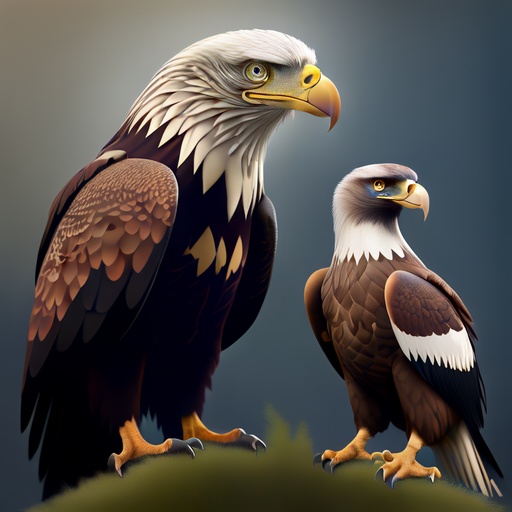} &     \includegraphics[width=0.1\textwidth]{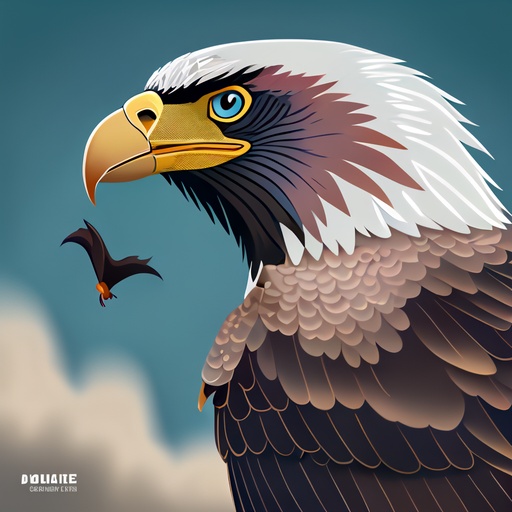} &     \includegraphics[width=0.1\textwidth]{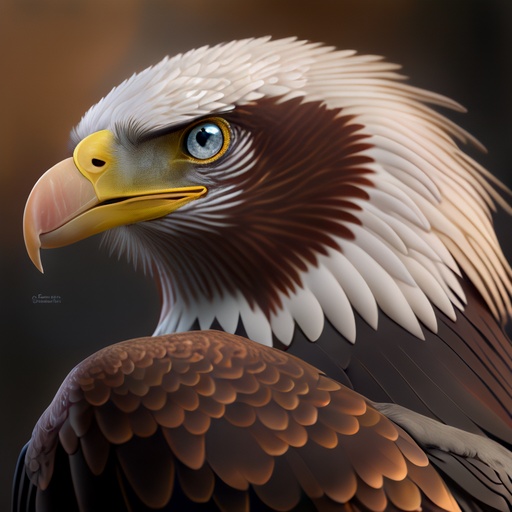} \\
 & \hspace{0.1cm}
    \includegraphics[width=0.1\textwidth]{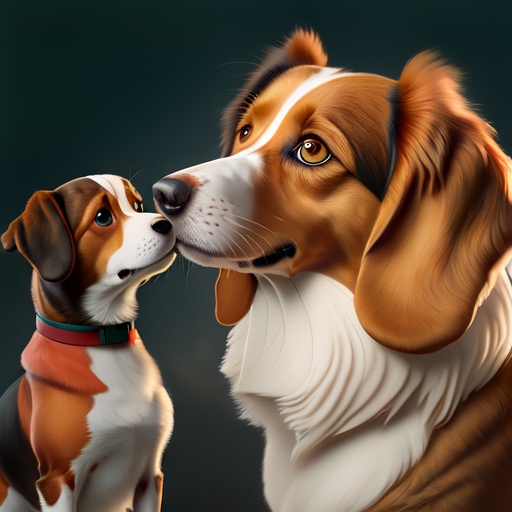} &     \includegraphics[width=0.1\textwidth]{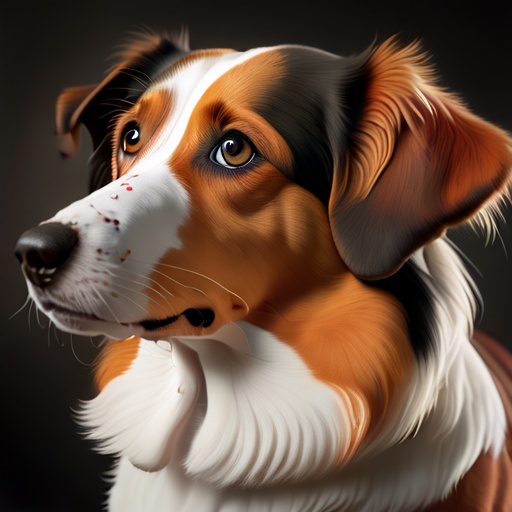} &     \includegraphics[width=0.1\textwidth]{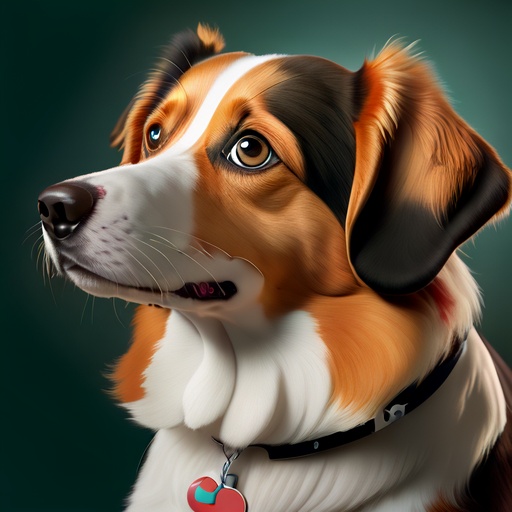} &     \includegraphics[width=0.1\textwidth]{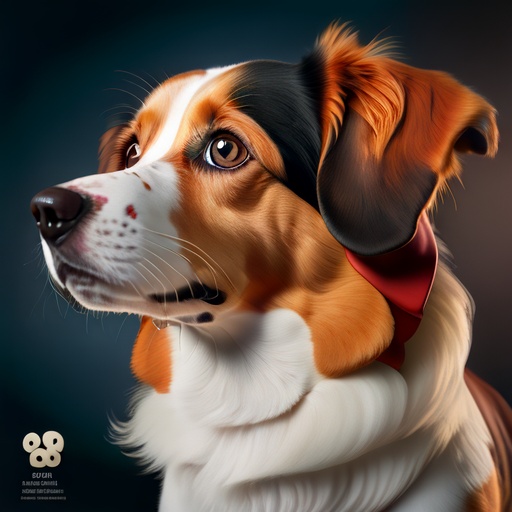} &     \includegraphics[width=0.1\textwidth]{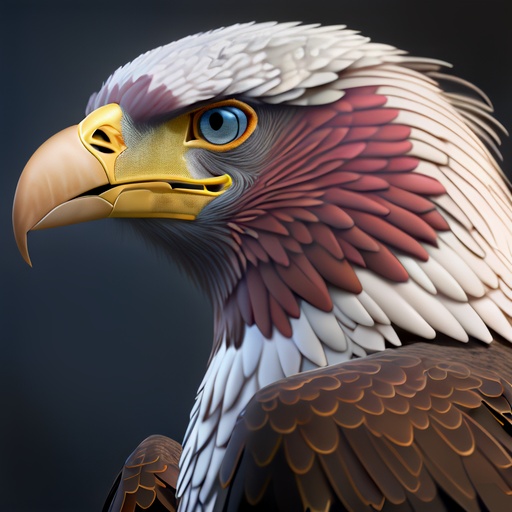} &     \includegraphics[width=0.1\textwidth]{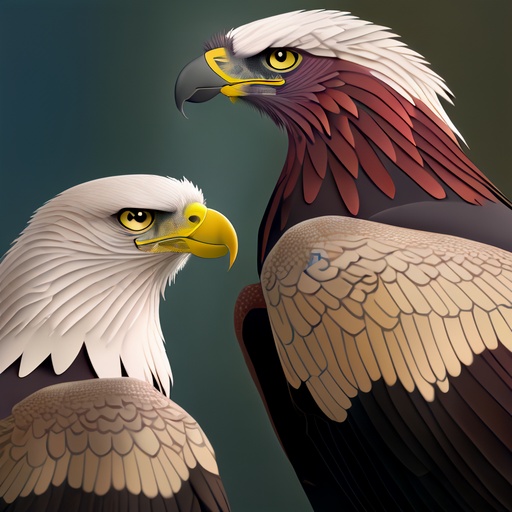} &     \includegraphics[width=0.1\textwidth]{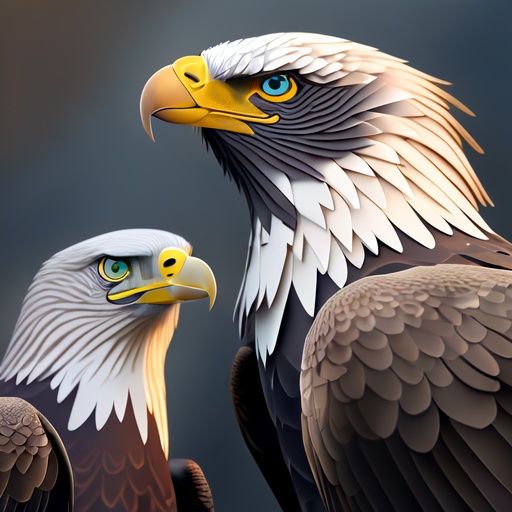} &     \includegraphics[width=0.1\textwidth]{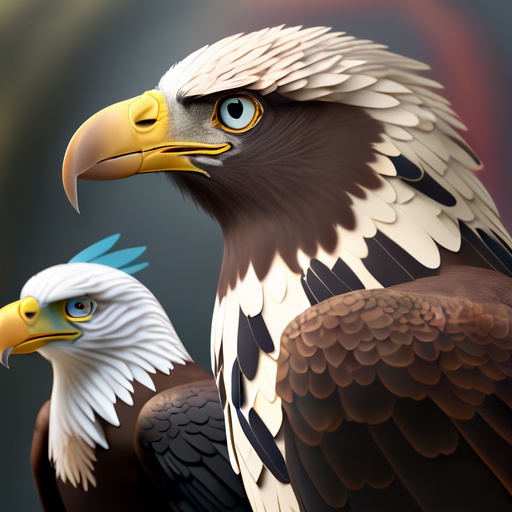} \\
    \\ \\ \\

 & \hspace{0.1cm}
    \includegraphics[width=0.1\textwidth]{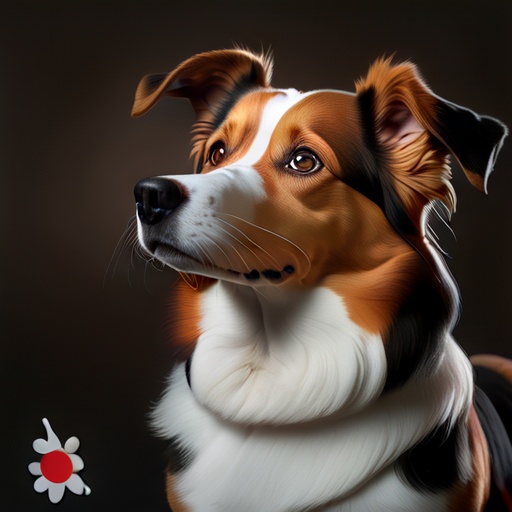} &     \includegraphics[width=0.1\textwidth]{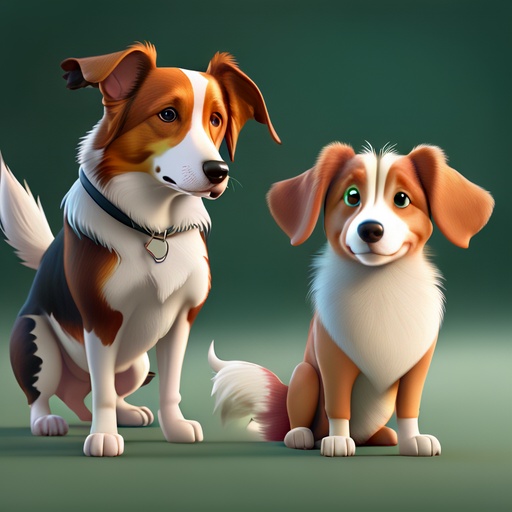} &     \includegraphics[width=0.1\textwidth]{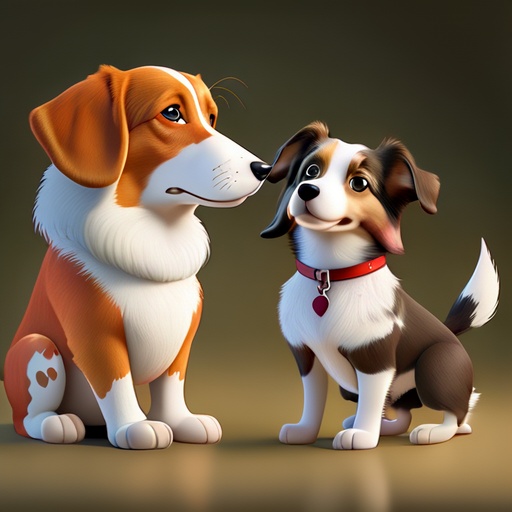} &     \includegraphics[width=0.1\textwidth]{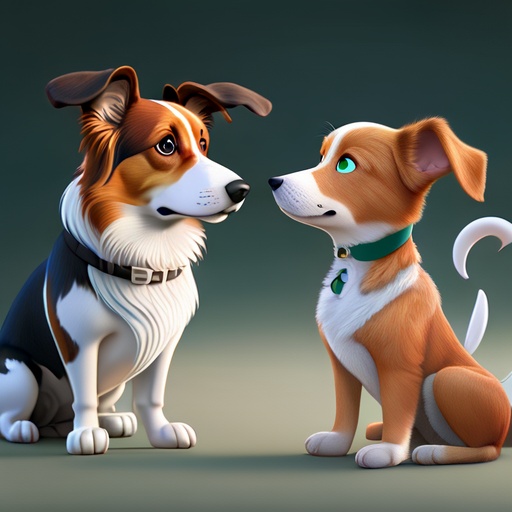} &     \includegraphics[width=0.1\textwidth]{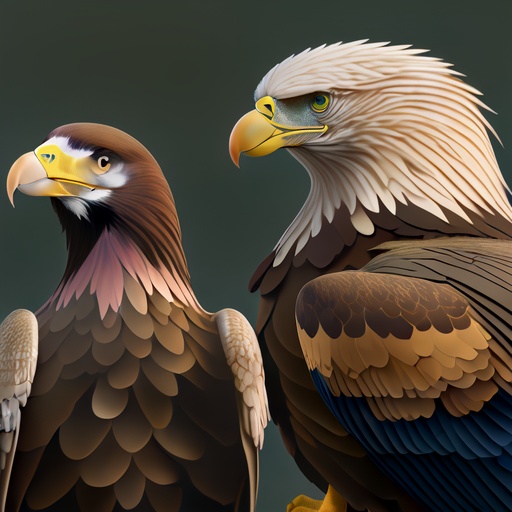} &     \includegraphics[width=0.1\textwidth]{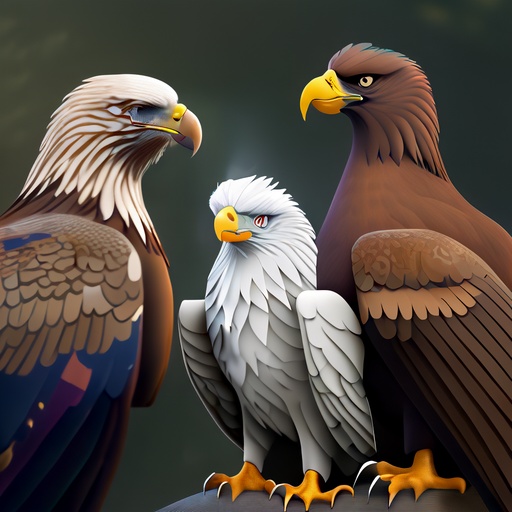} &     \includegraphics[width=0.1\textwidth]{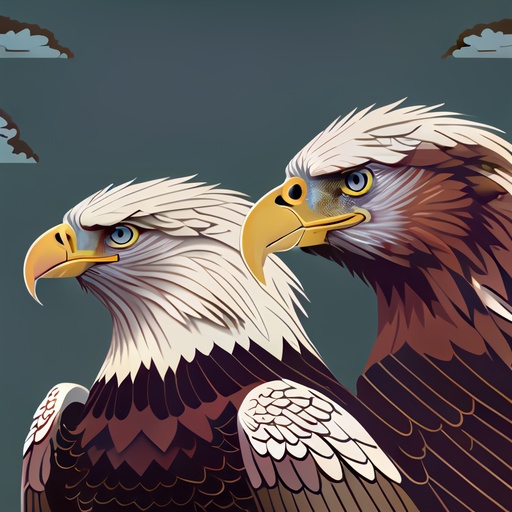} &     \includegraphics[width=0.1\textwidth]{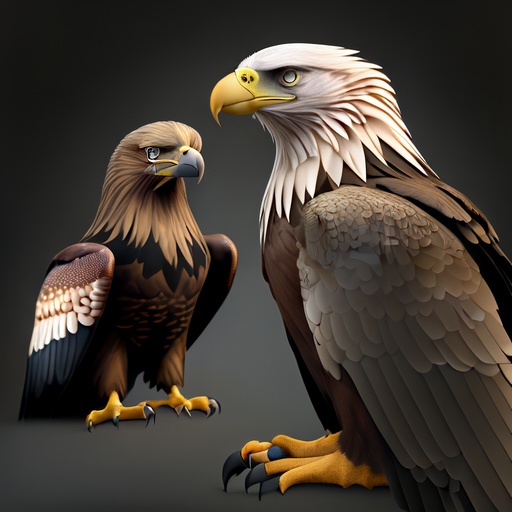} \\
{\raisebox{0.4in}{\multirow{4}{*}{\rotatebox{90}{\normalsize \textbf{\oursabbr{} (ours)}}}}} & \hspace{0.1cm}
    \includegraphics[width=0.1\textwidth]{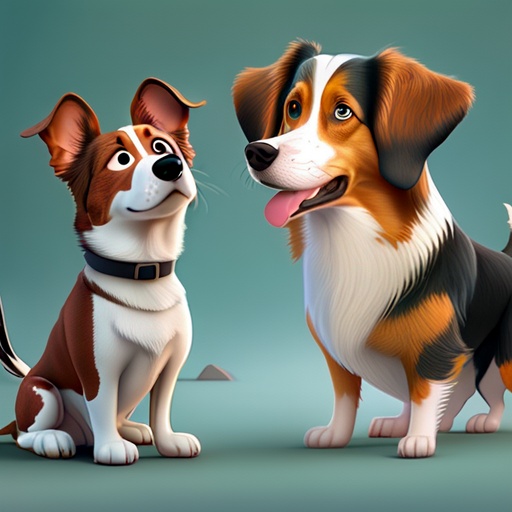} &     \includegraphics[width=0.1\textwidth]{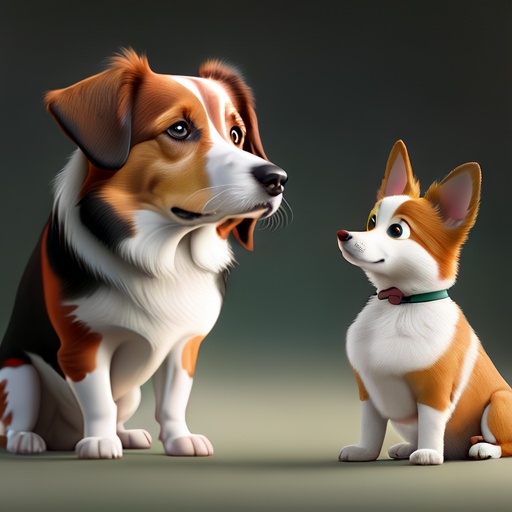} &     \includegraphics[width=0.1\textwidth]{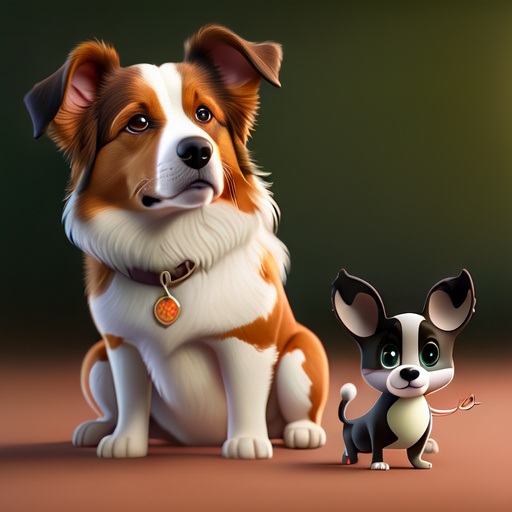} &     \includegraphics[width=0.1\textwidth]{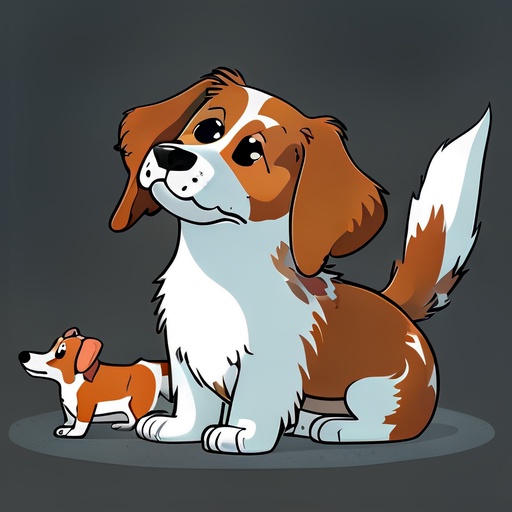} &     \includegraphics[width=0.1\textwidth]{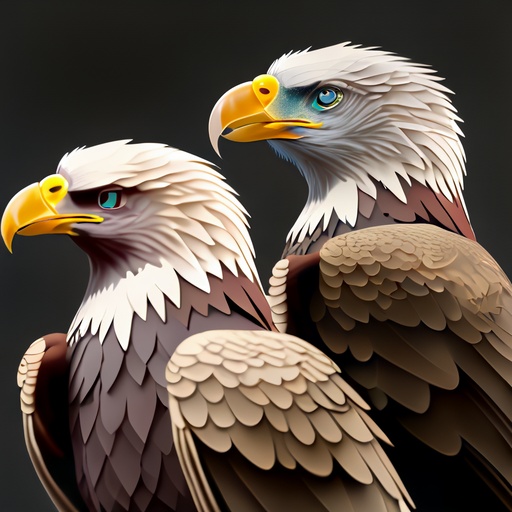} &     \includegraphics[width=0.1\textwidth]{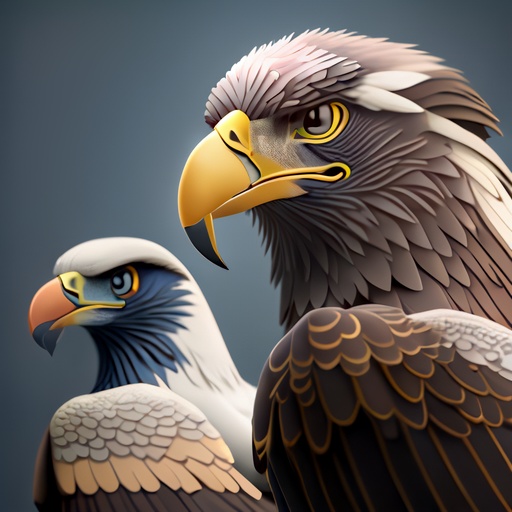} &     \includegraphics[width=0.1\textwidth]{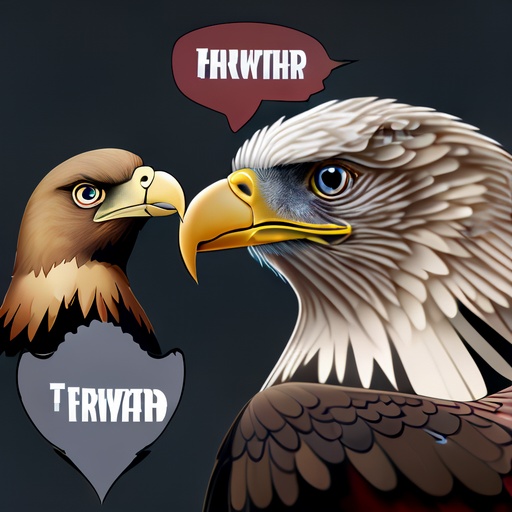} &     \includegraphics[width=0.1\textwidth]{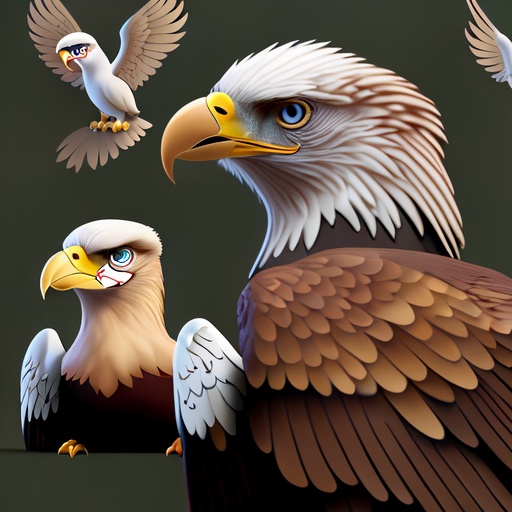} \\
 & \hspace{0.1cm}
    \includegraphics[width=0.1\textwidth]{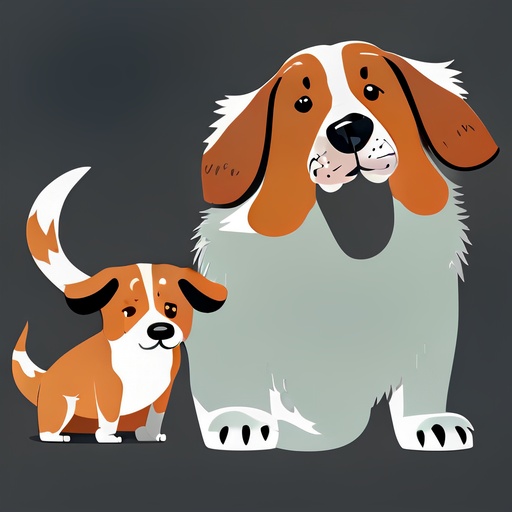} &     \includegraphics[width=0.1\textwidth]{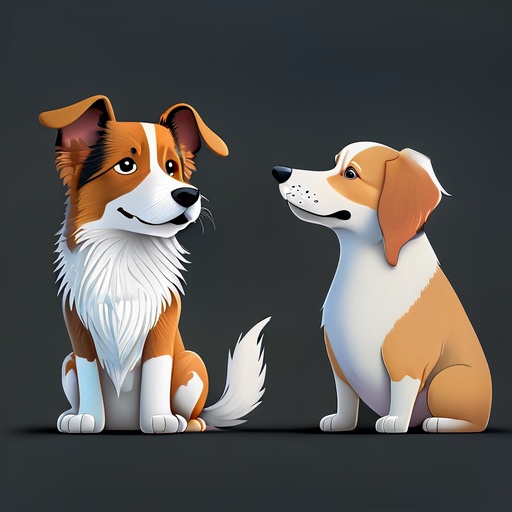} &     \includegraphics[width=0.1\textwidth]{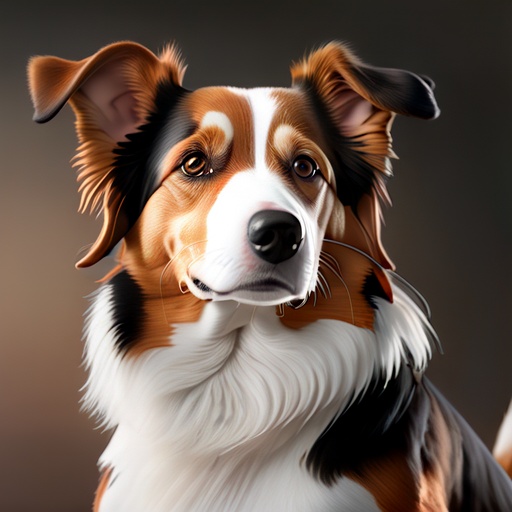} &     \includegraphics[width=0.1\textwidth]{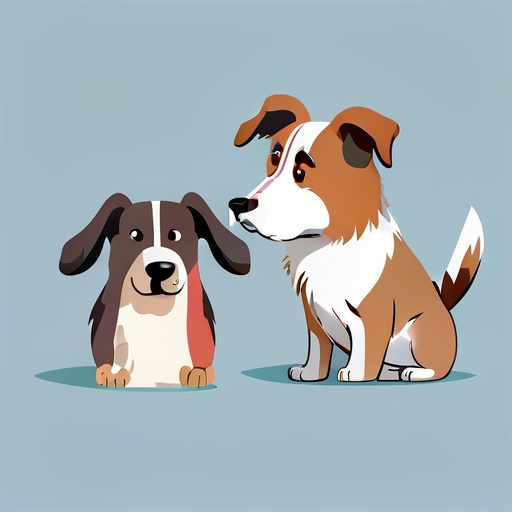} &     \includegraphics[width=0.1\textwidth]{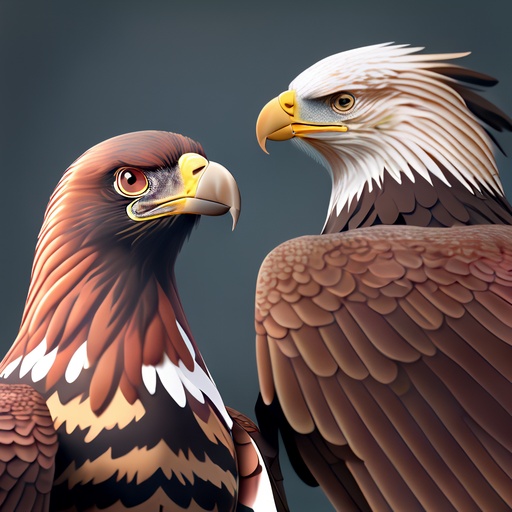} &     \includegraphics[width=0.1\textwidth]{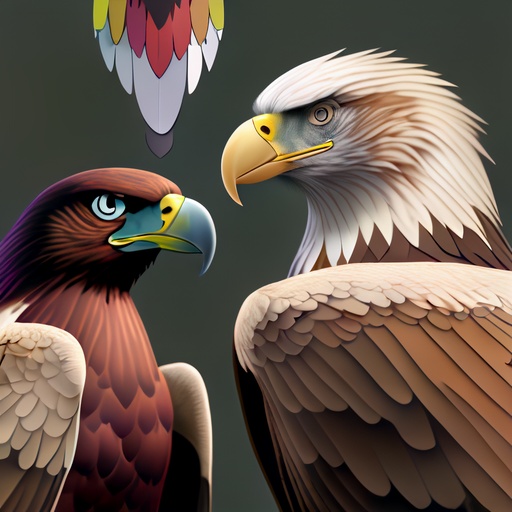} &     \includegraphics[width=0.1\textwidth]{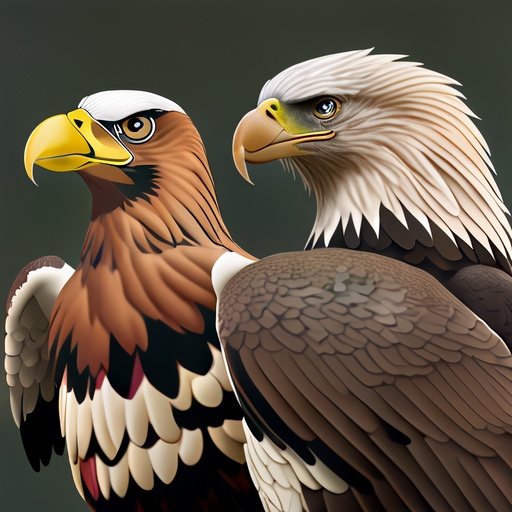} &     \includegraphics[width=0.1\textwidth]{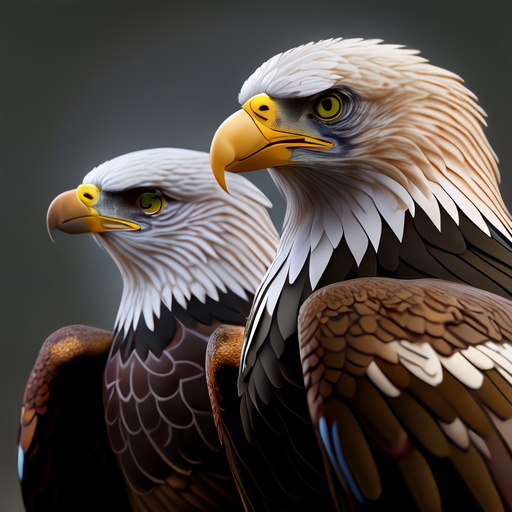} \\
 & \hspace{0.1cm}
    \includegraphics[width=0.1\textwidth]{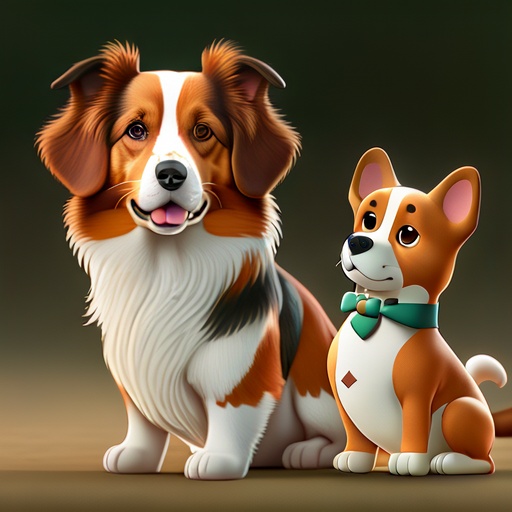} &     \includegraphics[width=0.1\textwidth]{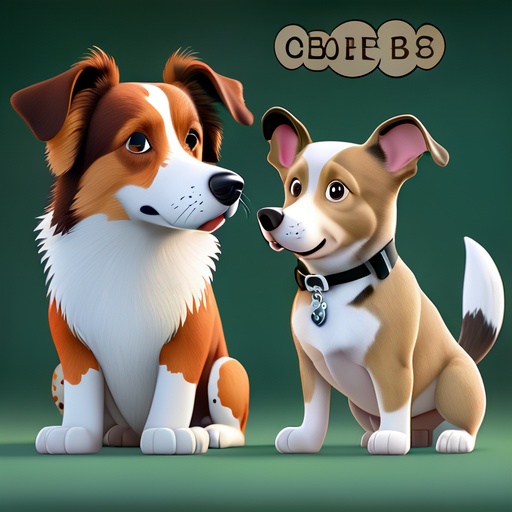} &     \includegraphics[width=0.1\textwidth]{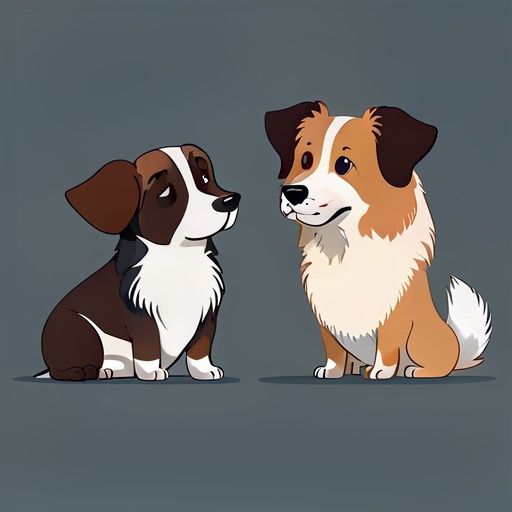} &     \includegraphics[width=0.1\textwidth]{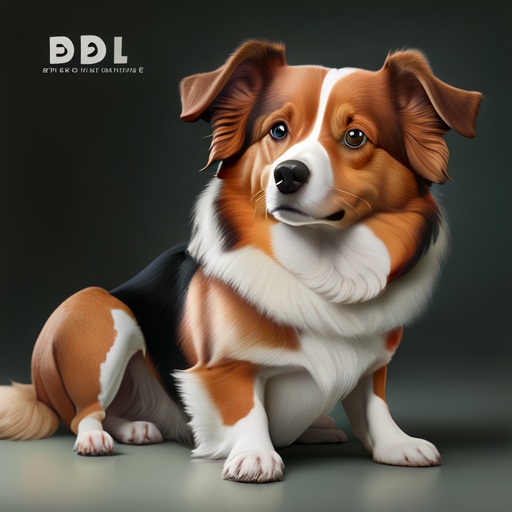} &     \includegraphics[width=0.1\textwidth]{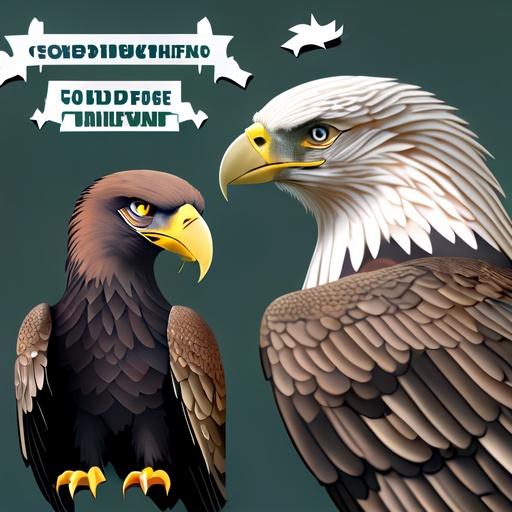} &     \includegraphics[width=0.1\textwidth]{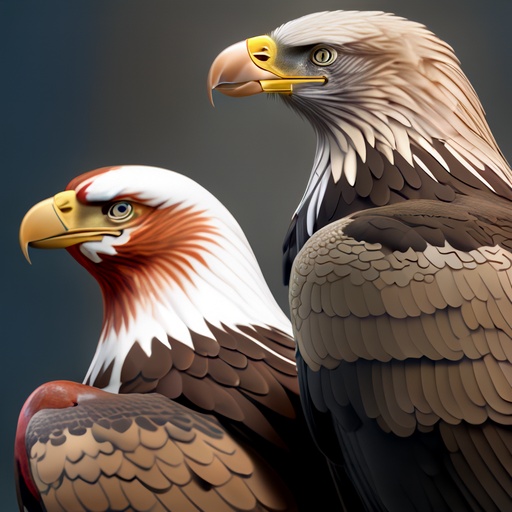} &     \includegraphics[width=0.1\textwidth]{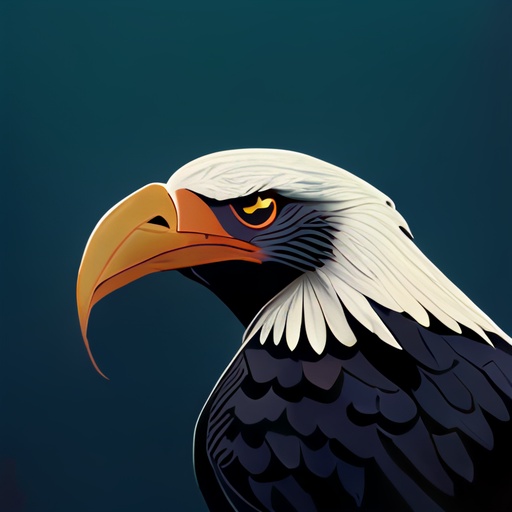} &     \includegraphics[width=0.1\textwidth]{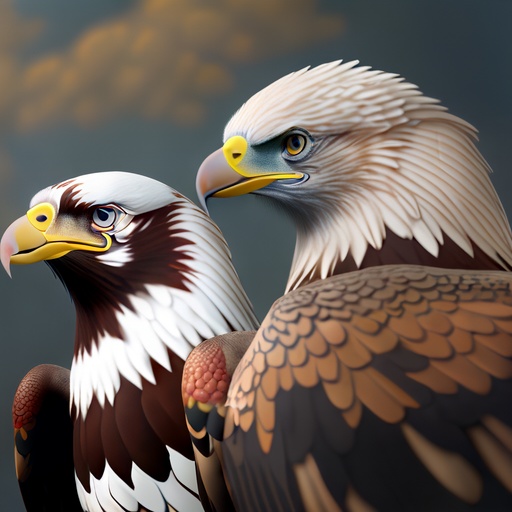} \\

\end{tabular}
    }
    }
\caption{Additional qualitative results on SSD~\citep{ssd} dataset (1). The same seeds are applied to each prompt across all methods.}
\label{fig:app_qual_ssd_1}
\end{figure*}

\begin{figure*}[!htp]
    \centering
    \setlength{\tabcolsep}{0.5pt}
    \renewcommand{\arraystretch}{0.3}
    \resizebox{1.0\linewidth}{!}{
    {\footnotesize
\begin{tabular}{c c c c c @{\hspace{0.1cm}} c c c c}

    & \multicolumn{4}{c}{\textit{“a eagle and a owl”}} & \multicolumn{4}{c}{\textit{“a seal and a manatee”}} \\ \\

& \hspace{0.1cm}
    \includegraphics[width=0.1\textwidth]{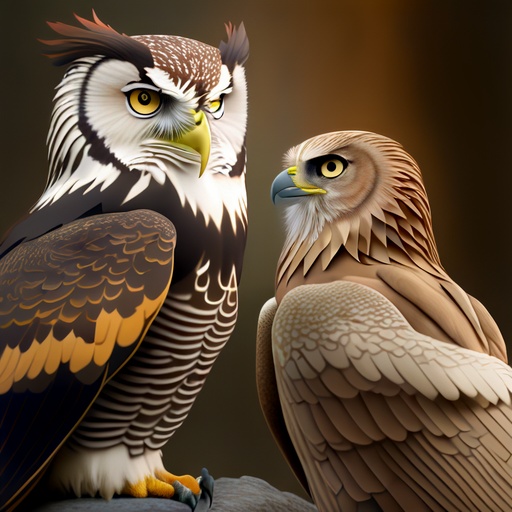} &     \includegraphics[width=0.1\textwidth]{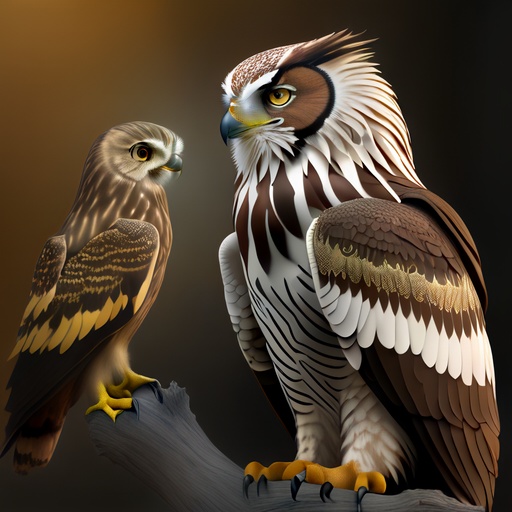} &     \includegraphics[width=0.1\textwidth]{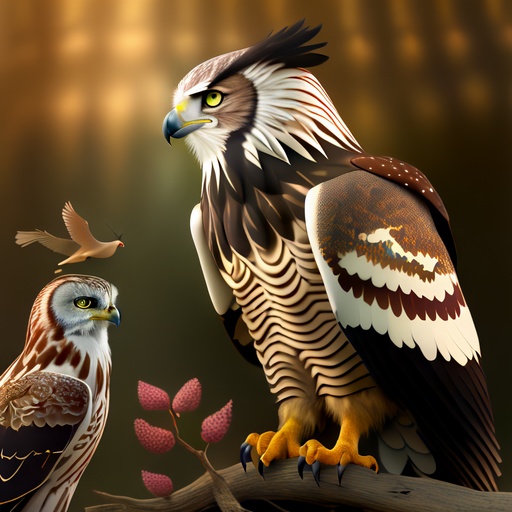} &     \includegraphics[width=0.1\textwidth]{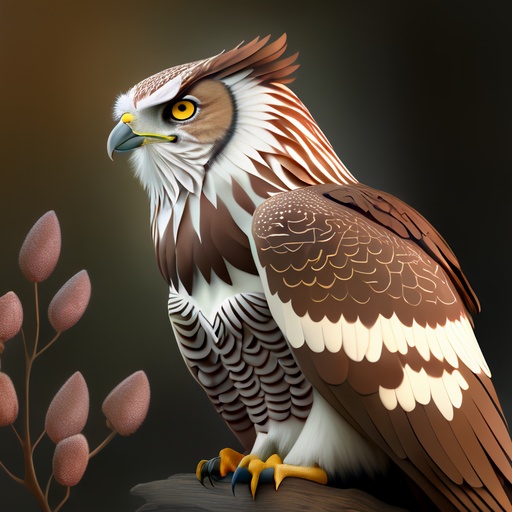} &     \includegraphics[width=0.1\textwidth]{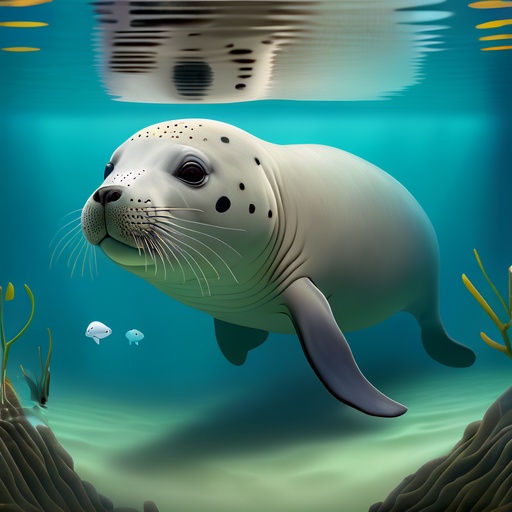} &     \includegraphics[width=0.1\textwidth]{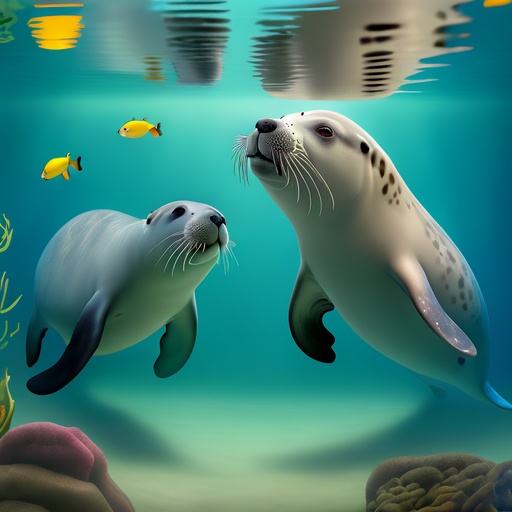} &     \includegraphics[width=0.1\textwidth]{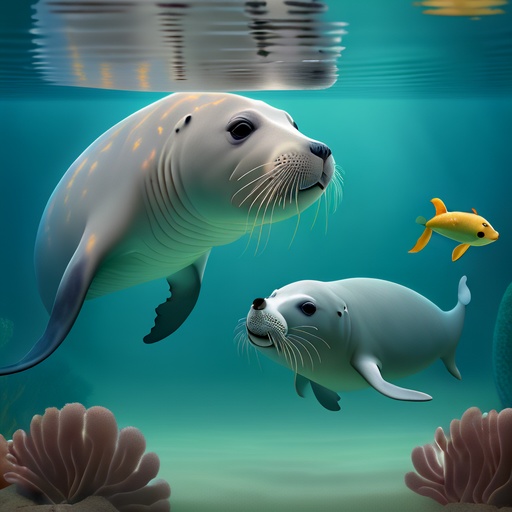} &     \includegraphics[width=0.1\textwidth]{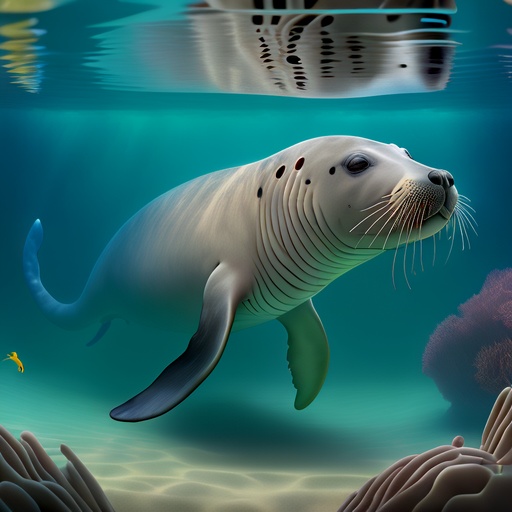} \\
{\raisebox{0.45in}{\multirow{4}{*}{\rotatebox{90}{\normalsize Meissonic (baseline)}}}} & \hspace{0.1cm}
    \includegraphics[width=0.1\textwidth]{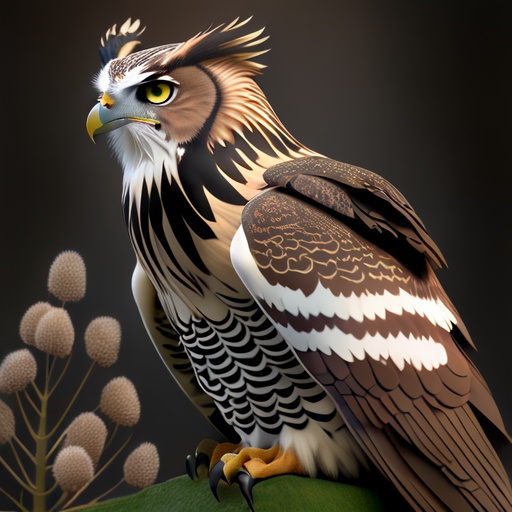} &     \includegraphics[width=0.1\textwidth]{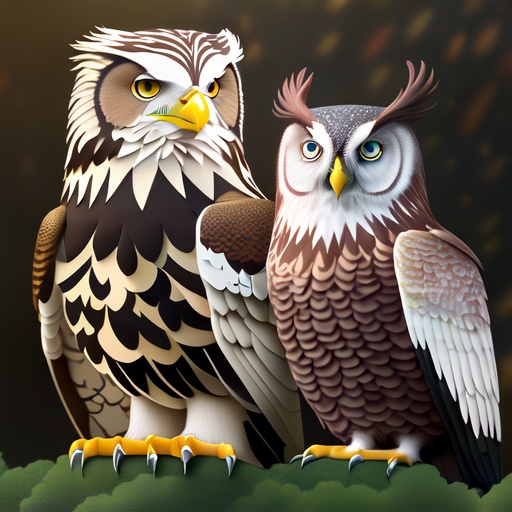} &     \includegraphics[width=0.1\textwidth]{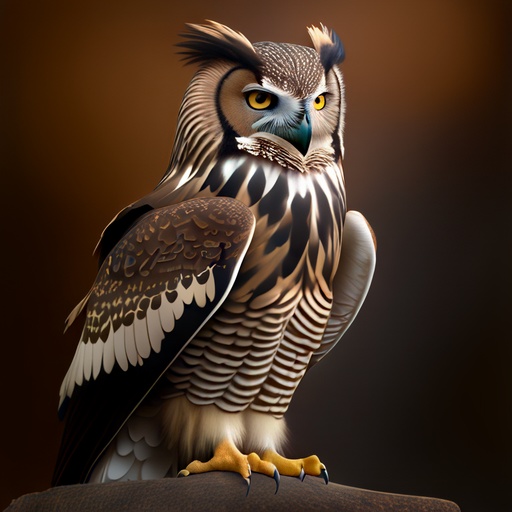} &     \includegraphics[width=0.1\textwidth]{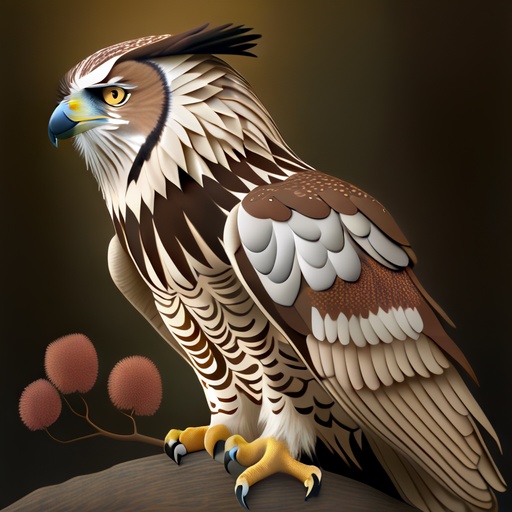} &     \includegraphics[width=0.1\textwidth]{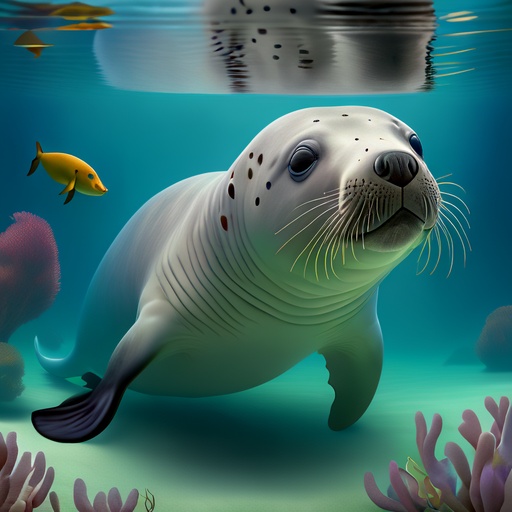} &     \includegraphics[width=0.1\textwidth]{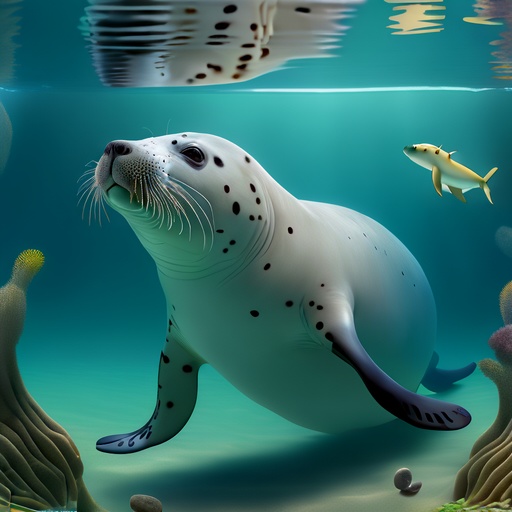} &     \includegraphics[width=0.1\textwidth]{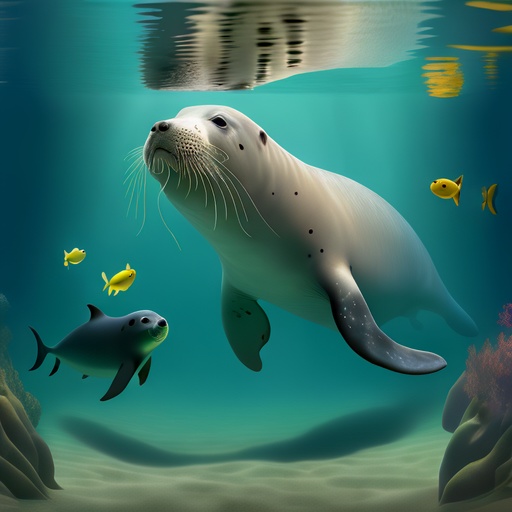} &     \includegraphics[width=0.1\textwidth]{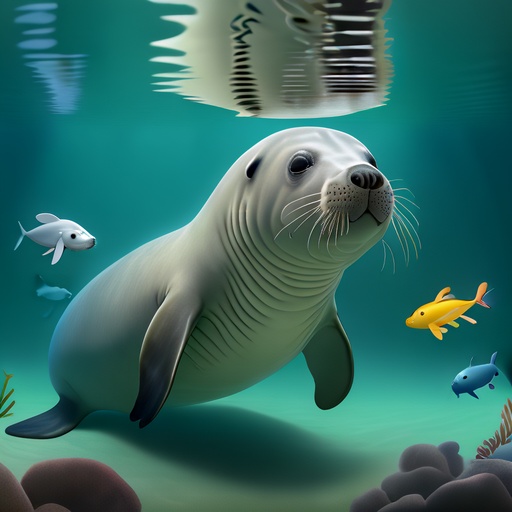} \\
 & \hspace{0.1cm}
    \includegraphics[width=0.1\textwidth]{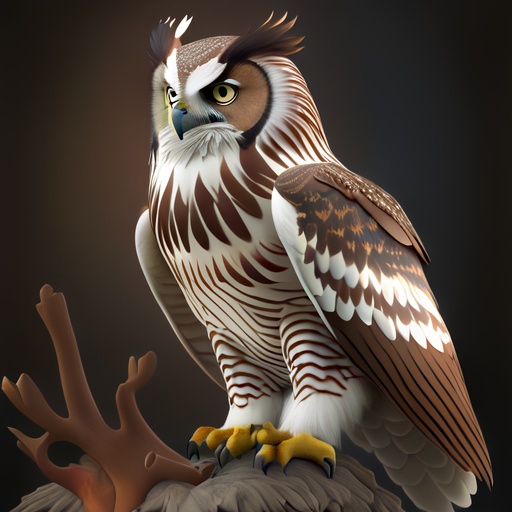} &     \includegraphics[width=0.1\textwidth]{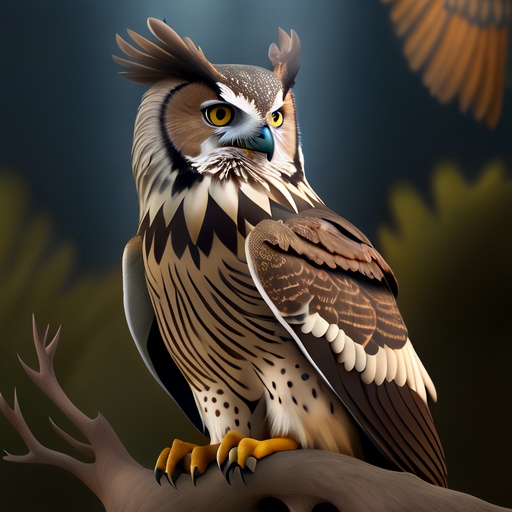} &     \includegraphics[width=0.1\textwidth]{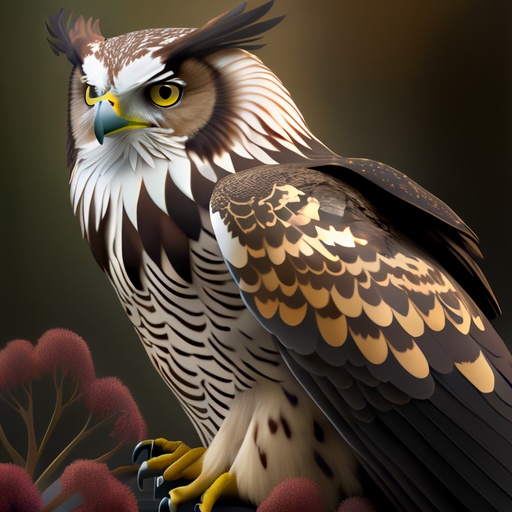} &     \includegraphics[width=0.1\textwidth]{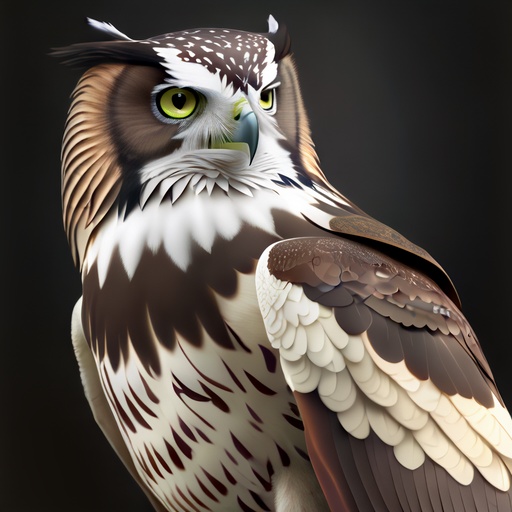} &     \includegraphics[width=0.1\textwidth]{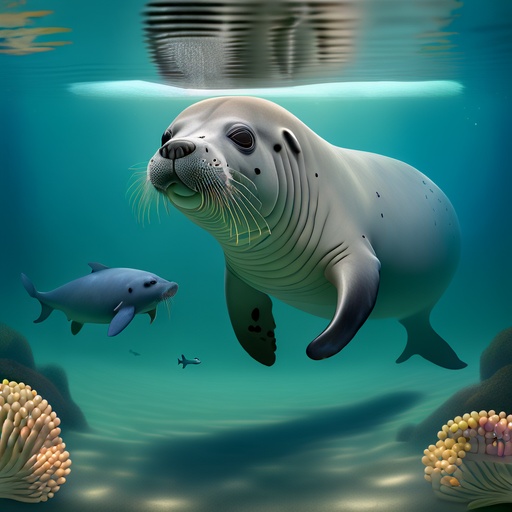} &     \includegraphics[width=0.1\textwidth]{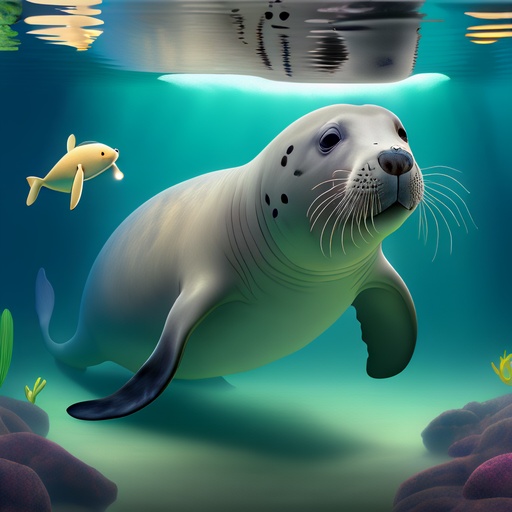} &     \includegraphics[width=0.1\textwidth]{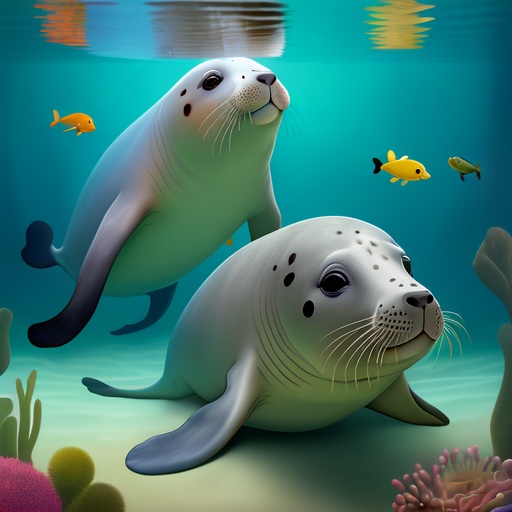} &     \includegraphics[width=0.1\textwidth]{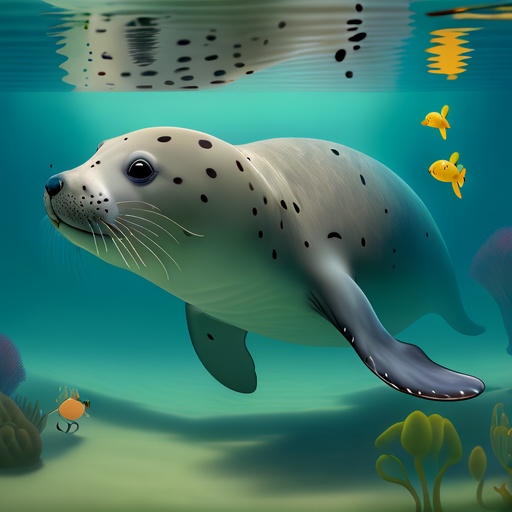} \\
 & \hspace{0.1cm}
    \includegraphics[width=0.1\textwidth]{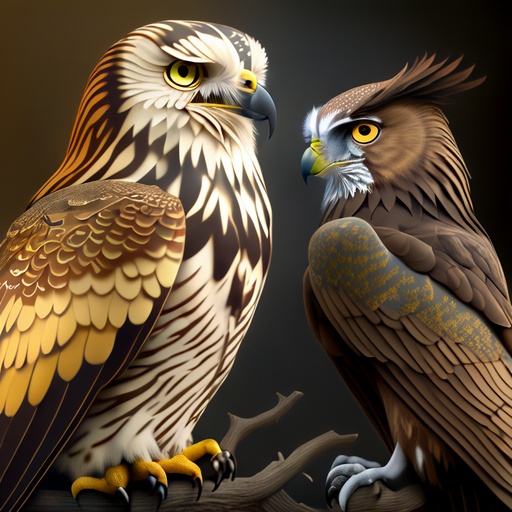} &     \includegraphics[width=0.1\textwidth]{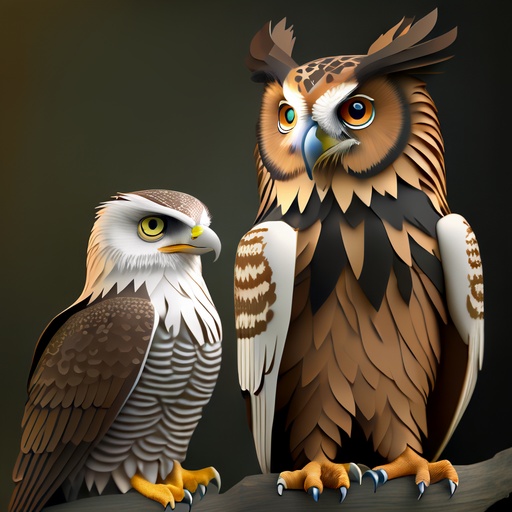} &     \includegraphics[width=0.1\textwidth]{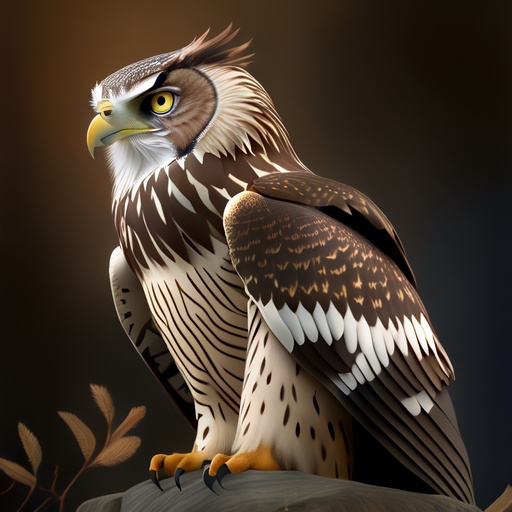} &     \includegraphics[width=0.1\textwidth]{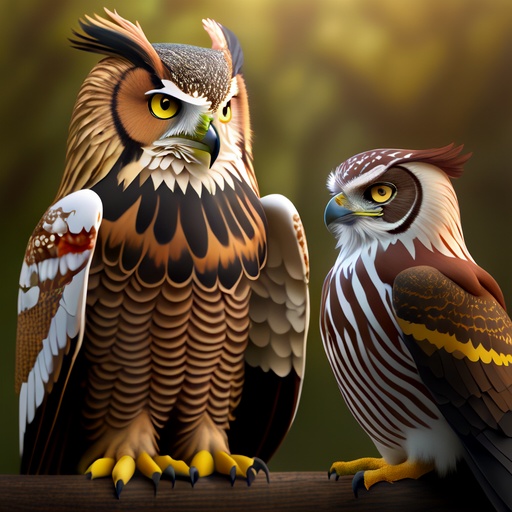} &     \includegraphics[width=0.1\textwidth]{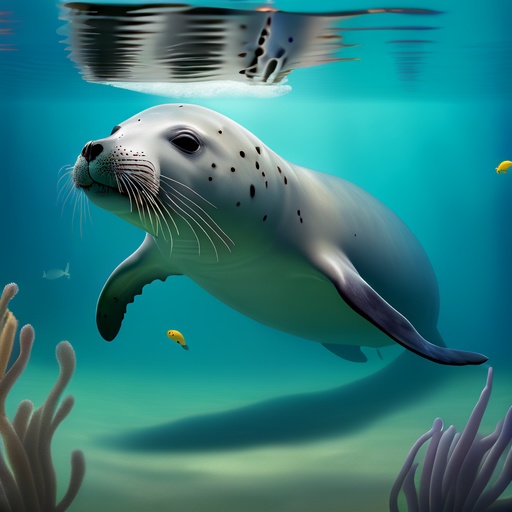} &     \includegraphics[width=0.1\textwidth]{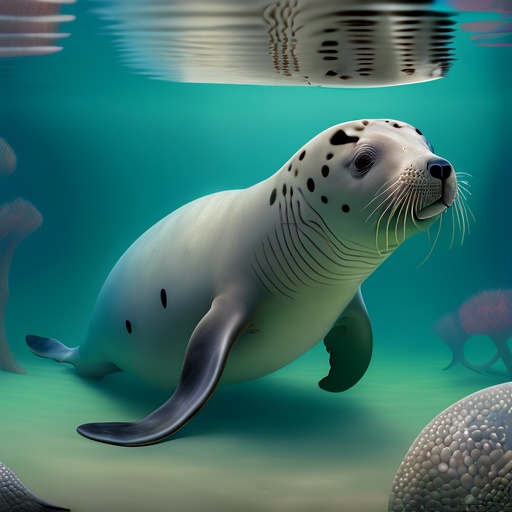} &     \includegraphics[width=0.1\textwidth]{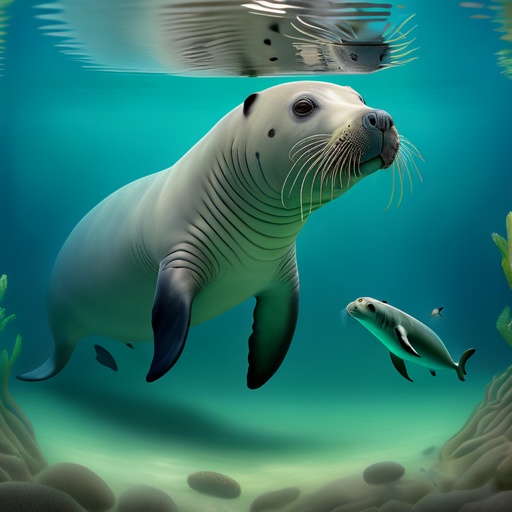} &     \includegraphics[width=0.1\textwidth]{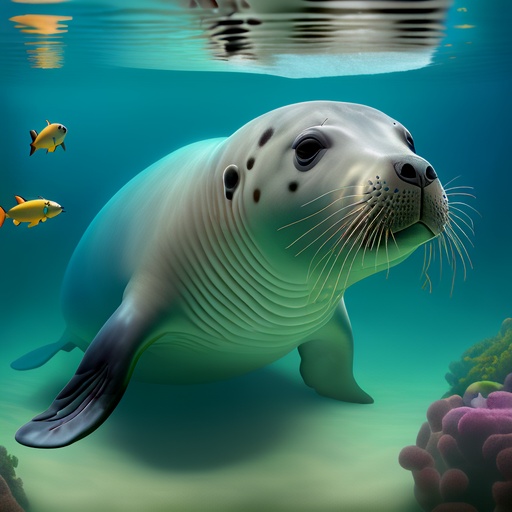} \\
    \\ \\ \\

& \hspace{0.1cm}
    \includegraphics[width=0.1\textwidth]{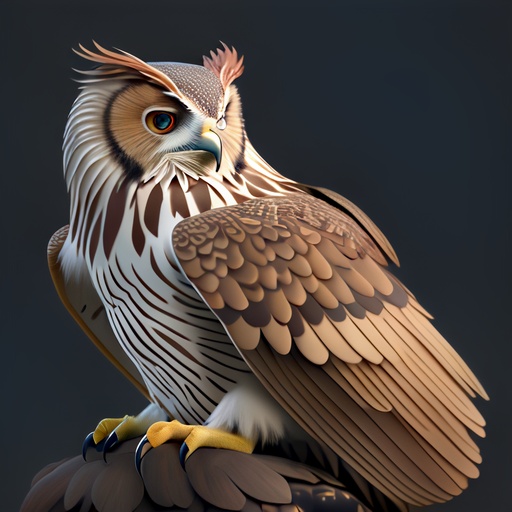} &     \includegraphics[width=0.1\textwidth]{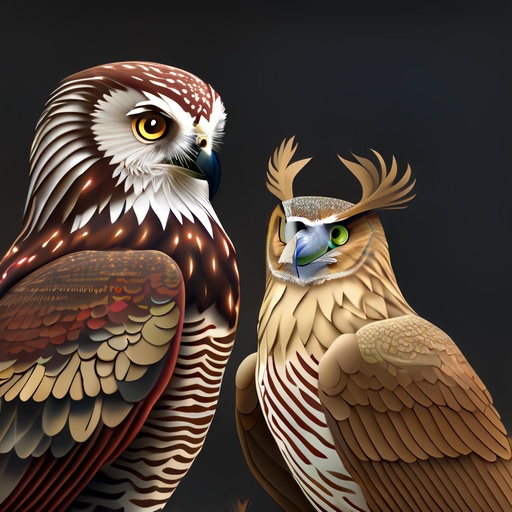} &     \includegraphics[width=0.1\textwidth]{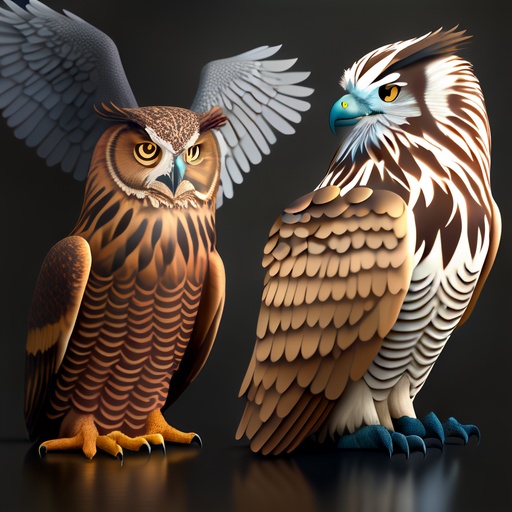} &     \includegraphics[width=0.1\textwidth]{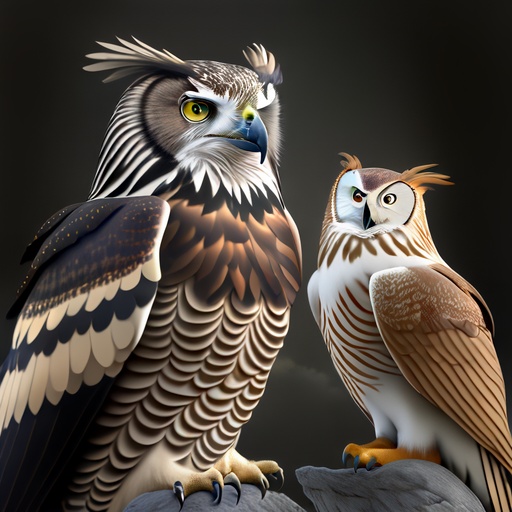} &     \includegraphics[width=0.1\textwidth]{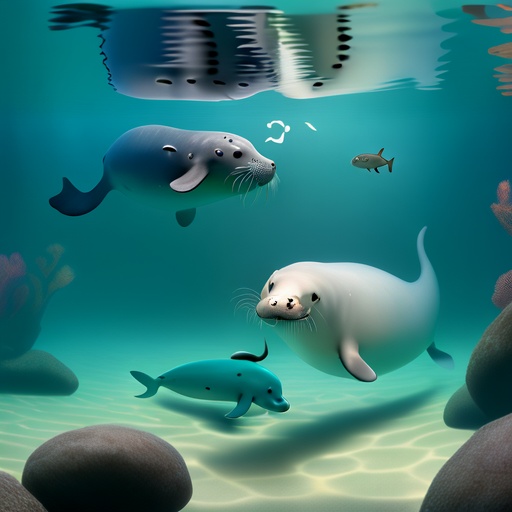} &     \includegraphics[width=0.1\textwidth]{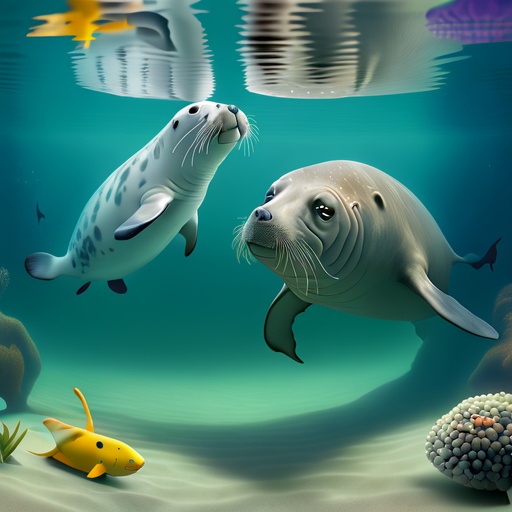} &     \includegraphics[width=0.1\textwidth]{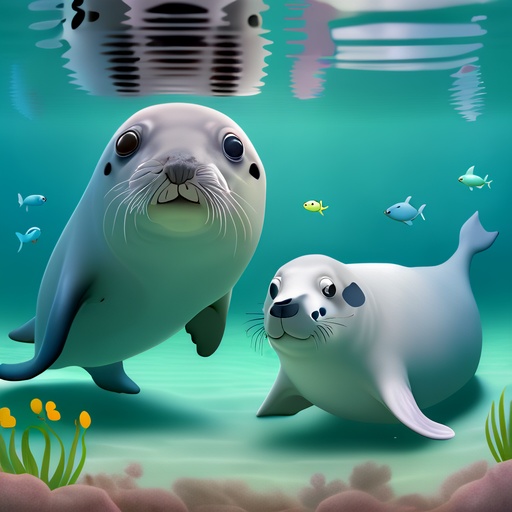} &     \includegraphics[width=0.1\textwidth]{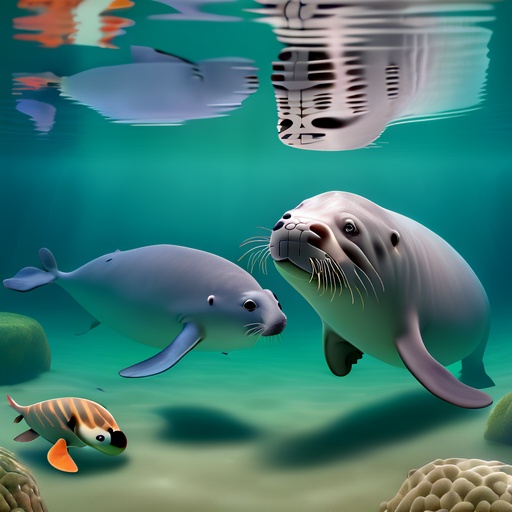} \\
{\raisebox{0.4in}{\multirow{4}{*}{\rotatebox{90}{\normalsize \textbf{\oursabbr{} (ours)}}}}} & \hspace{0.1cm}
    \includegraphics[width=0.1\textwidth]{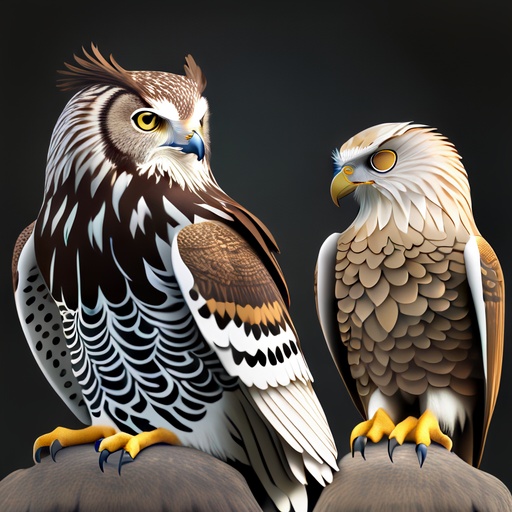} &     \includegraphics[width=0.1\textwidth]{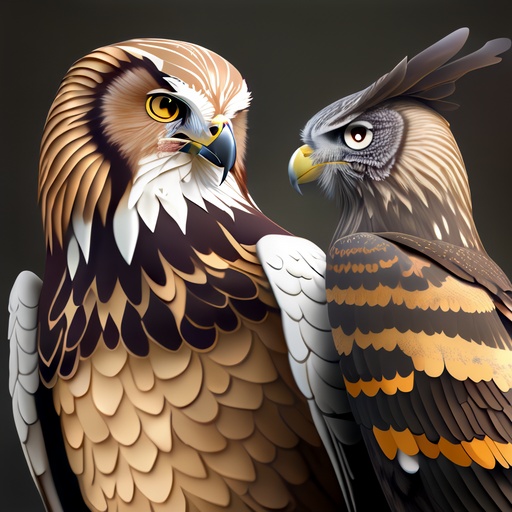} &     \includegraphics[width=0.1\textwidth]{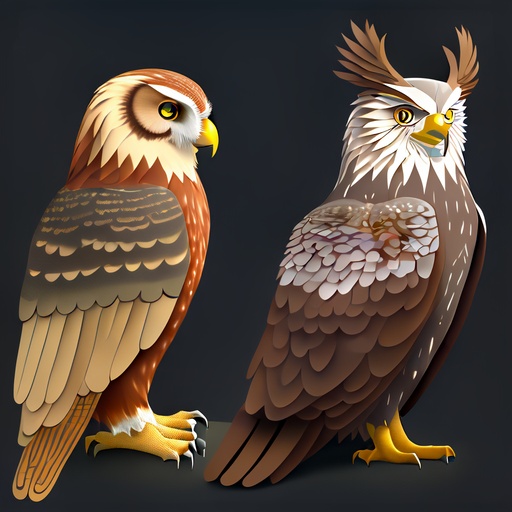} &     \includegraphics[width=0.1\textwidth]{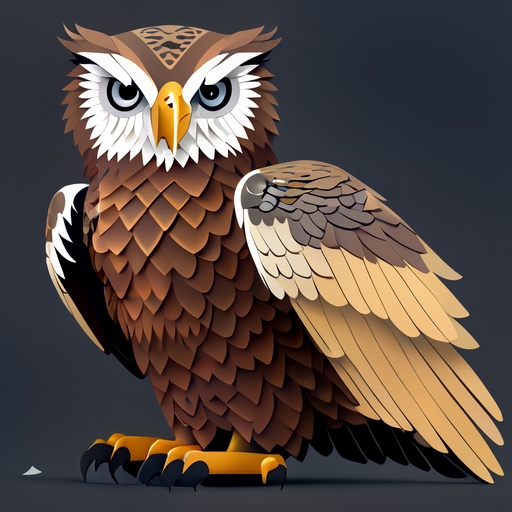} &     \includegraphics[width=0.1\textwidth]{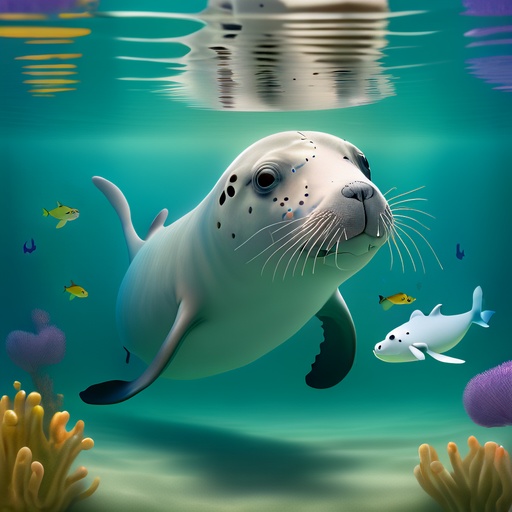} &     \includegraphics[width=0.1\textwidth]{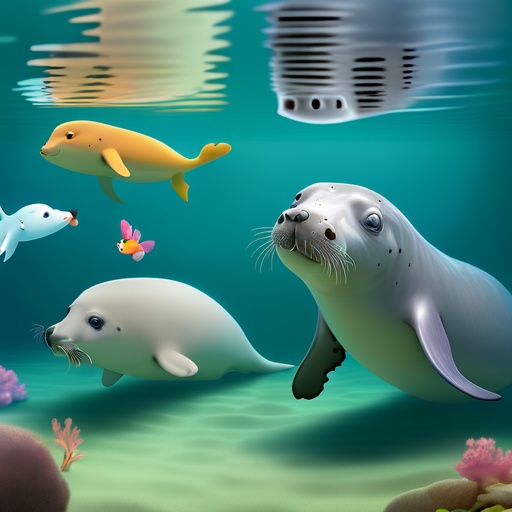} &     \includegraphics[width=0.1\textwidth]{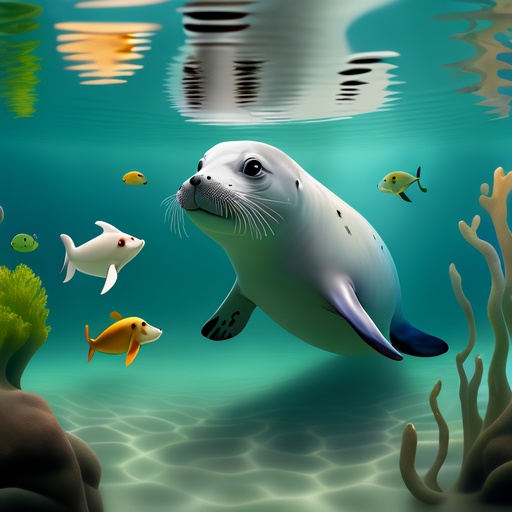} &     \includegraphics[width=0.1\textwidth]{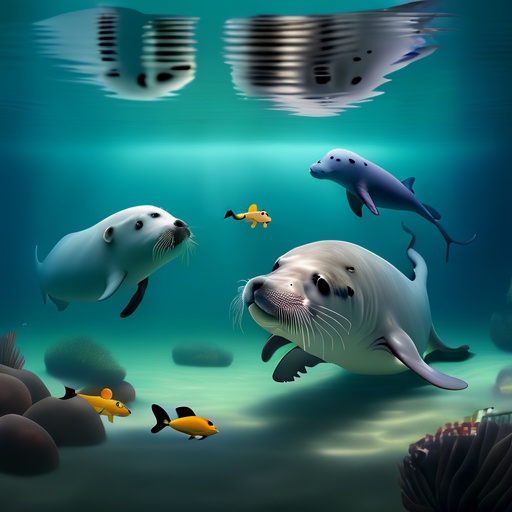} \\
 & \hspace{0.1cm}
    \includegraphics[width=0.1\textwidth]{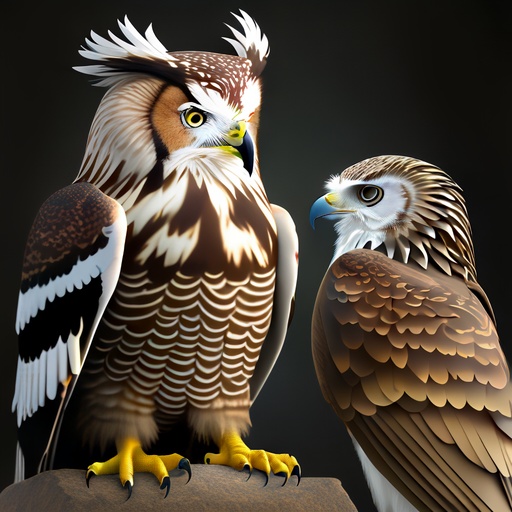} &     \includegraphics[width=0.1\textwidth]{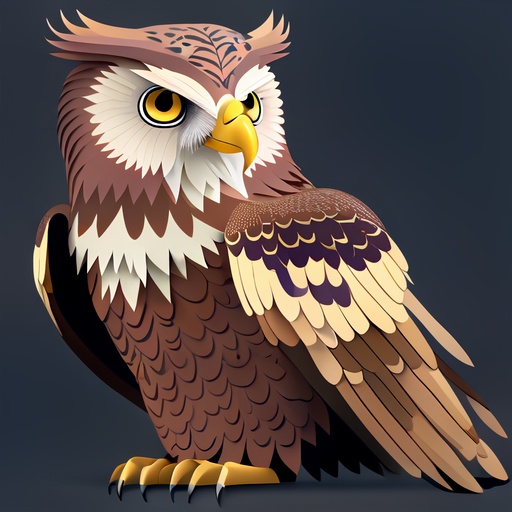} &     \includegraphics[width=0.1\textwidth]{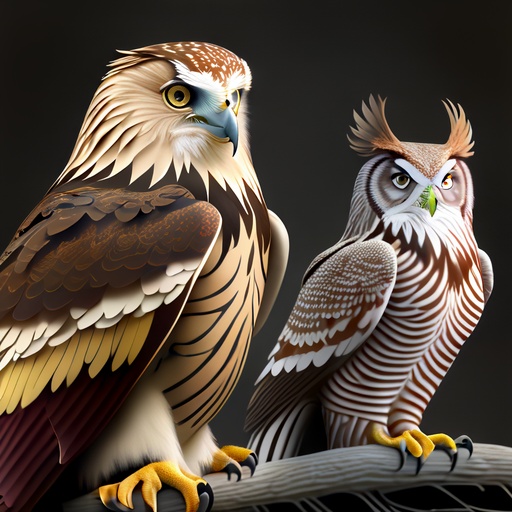} &     \includegraphics[width=0.1\textwidth]{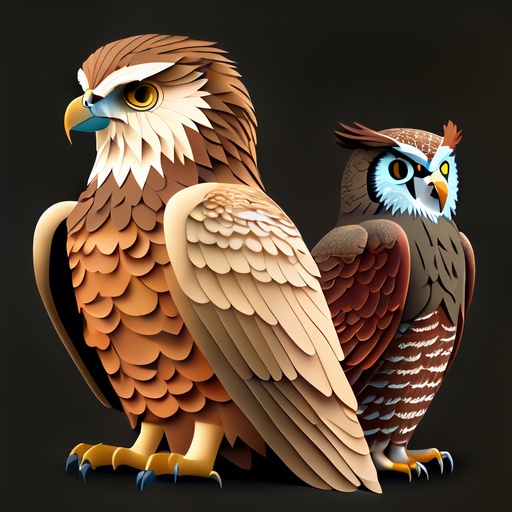} &     \includegraphics[width=0.1\textwidth]{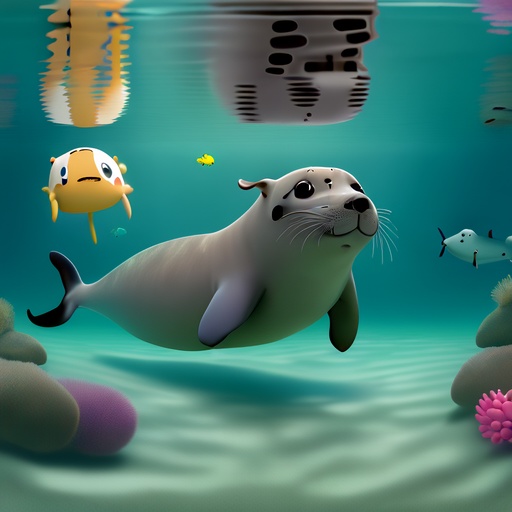} &     \includegraphics[width=0.1\textwidth]{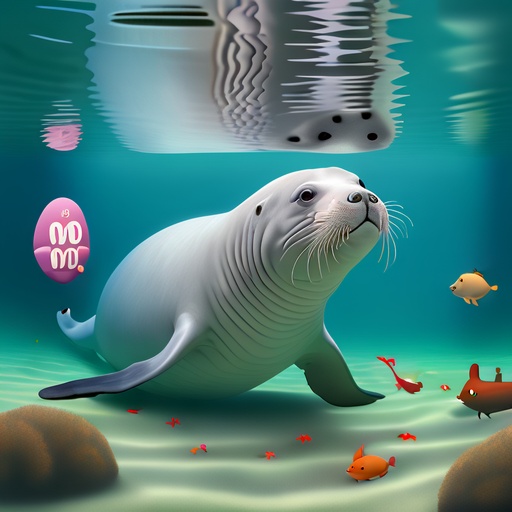} &     \includegraphics[width=0.1\textwidth]{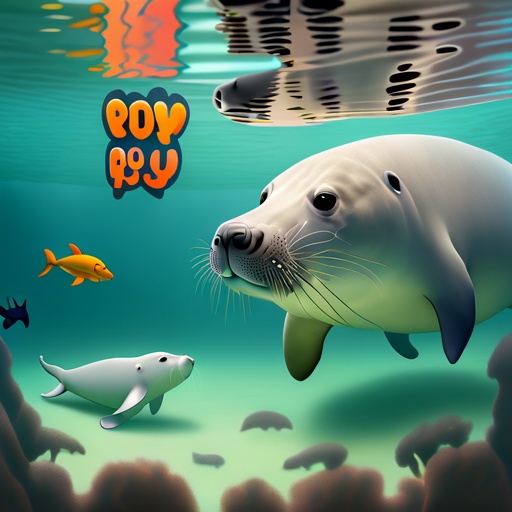} &     \includegraphics[width=0.1\textwidth]{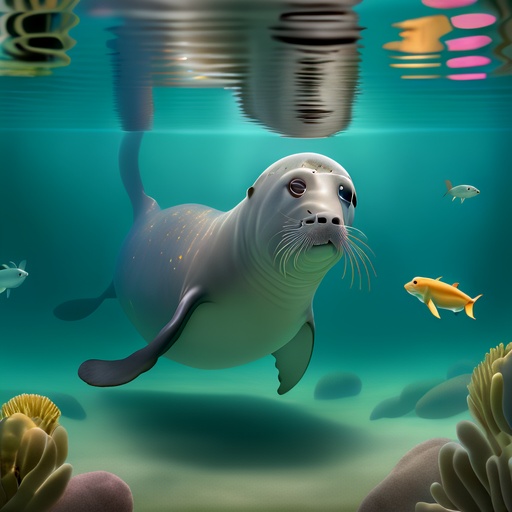} \\
 & \hspace{0.1cm}
    \includegraphics[width=0.1\textwidth]{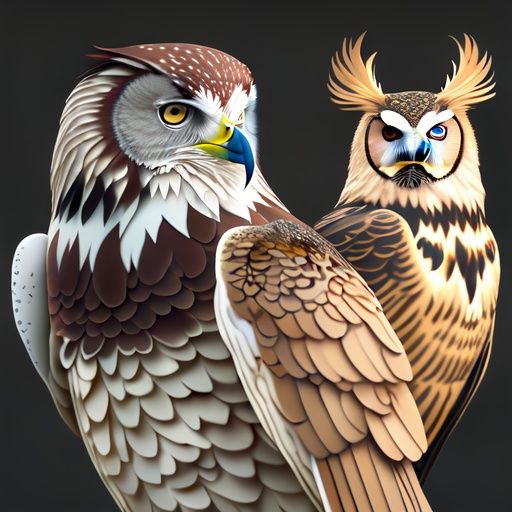} &     \includegraphics[width=0.1\textwidth]{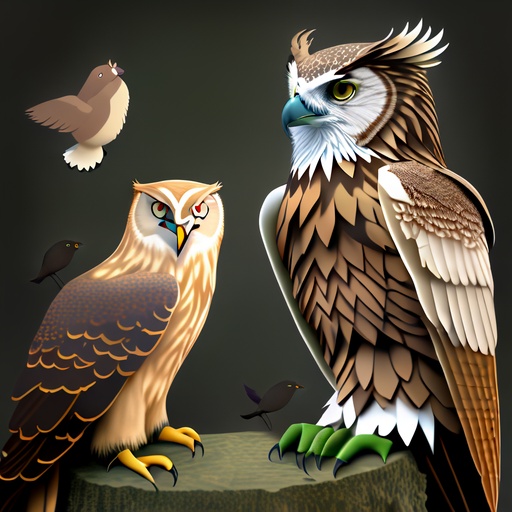} &     \includegraphics[width=0.1\textwidth]{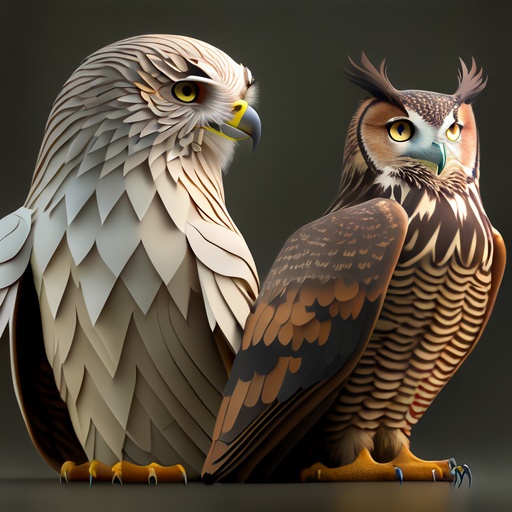} &     \includegraphics[width=0.1\textwidth]{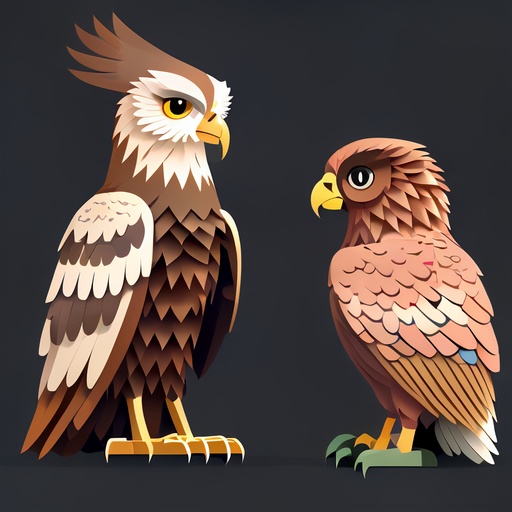} &     \includegraphics[width=0.1\textwidth]{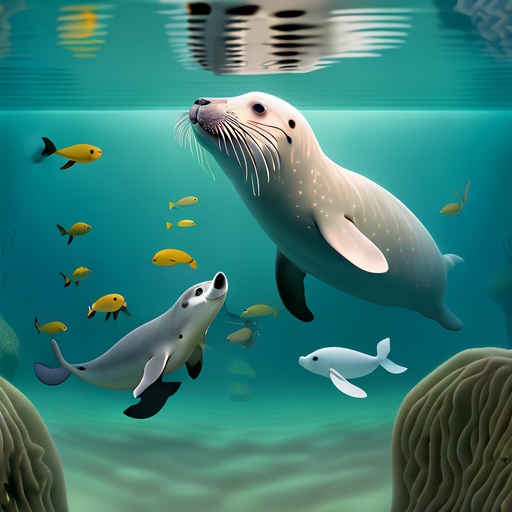} &     \includegraphics[width=0.1\textwidth]{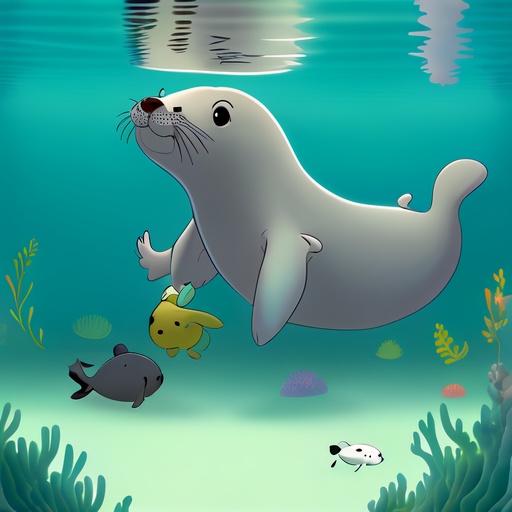} &     \includegraphics[width=0.1\textwidth]{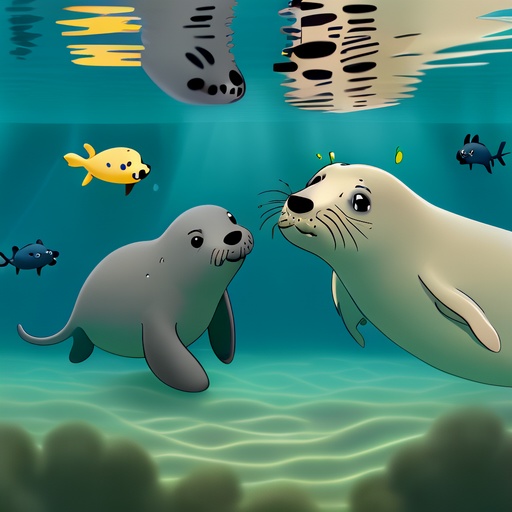} &     \includegraphics[width=0.1\textwidth]{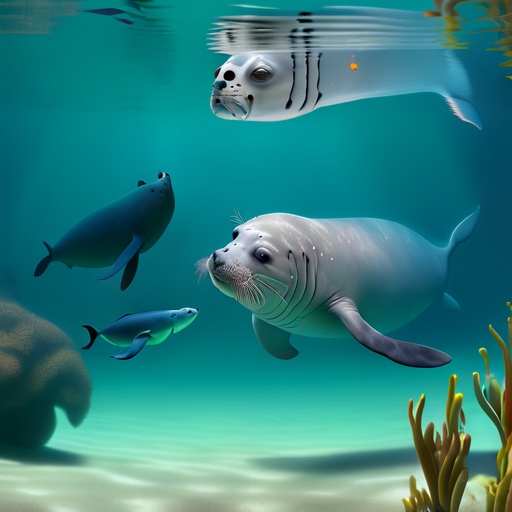} \\

\end{tabular}
    }
    }
\caption{Additional qualitative results on SSD~\citep{ssd} dataset (2). The same seeds are applied to each prompt across all methods.}
\label{fig:app_qual_ssd_2}
\end{figure*}

\begin{figure*}[!htp]
    \centering
    \setlength{\tabcolsep}{0.5pt}
    \renewcommand{\arraystretch}{0.3}
    \resizebox{1.0\linewidth}{!}{
    {\footnotesize
\begin{tabular}{c c c c c @{\hspace{0.1cm}} c c c c}

    & \multicolumn{4}{c}{\textit{“a parrot and a pigeon”}} & \multicolumn{4}{c}{\textit{“a duck and a penguin”}} \\ \\

 & \hspace{0.1cm}
    \includegraphics[width=0.1\textwidth]{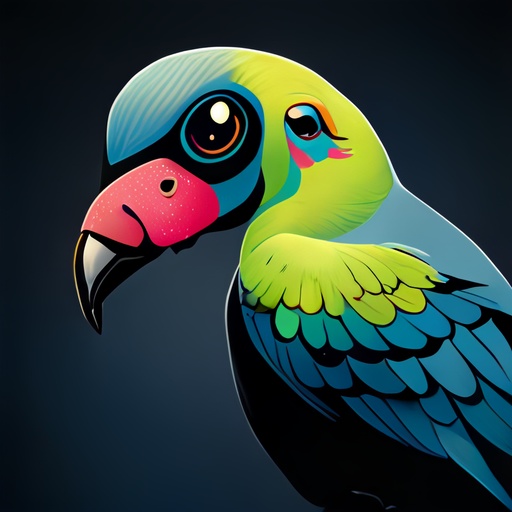} &     \includegraphics[width=0.1\textwidth]{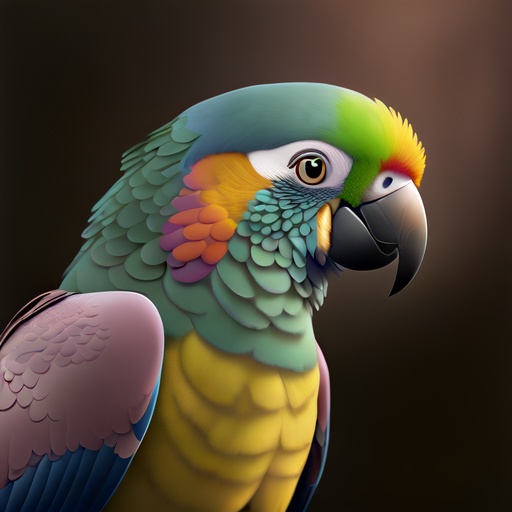} &     \includegraphics[width=0.1\textwidth]{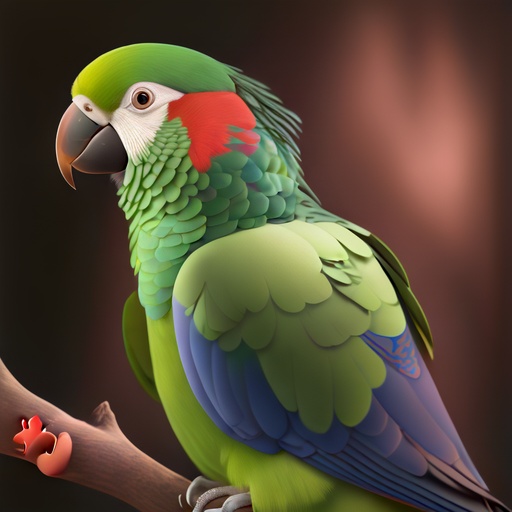} &     \includegraphics[width=0.1\textwidth]{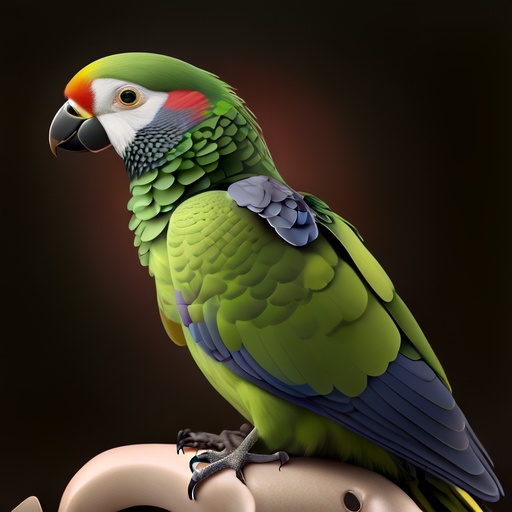} &     \includegraphics[width=0.1\textwidth]{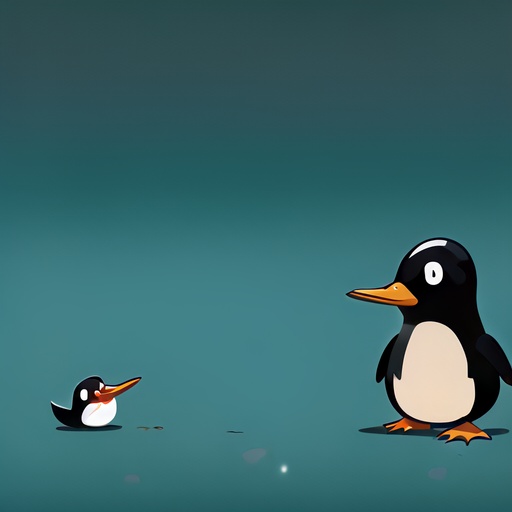} &     \includegraphics[width=0.1\textwidth]{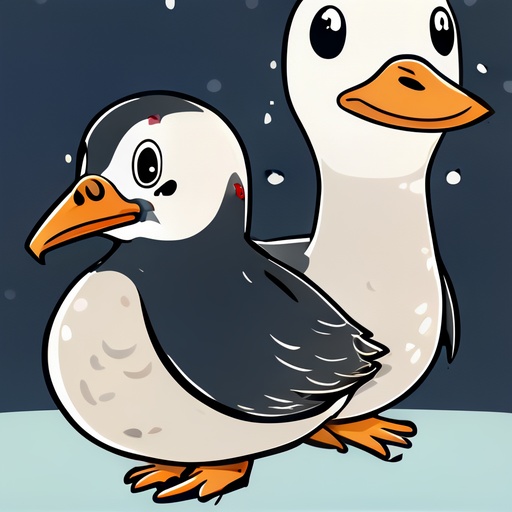} &     \includegraphics[width=0.1\textwidth]{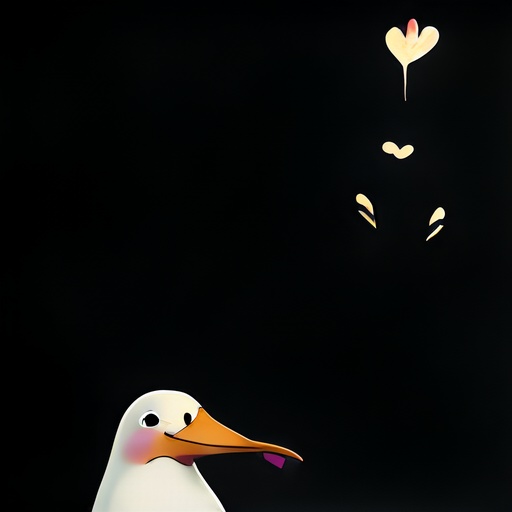} &     \includegraphics[width=0.1\textwidth]{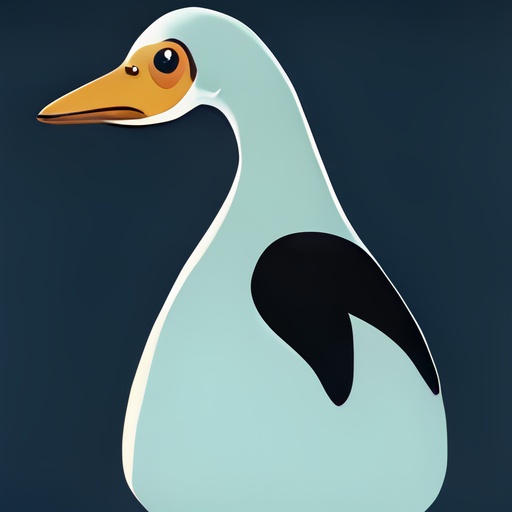} \\
{\raisebox{0.45in}{\multirow{2}{*}{\rotatebox{90}{\normalsize Meissonic (baseline)}}}}& \hspace{0.1cm}
    \includegraphics[width=0.1\textwidth]{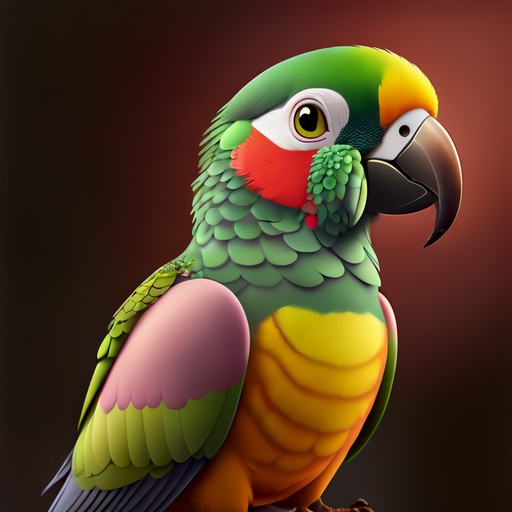} &     \includegraphics[width=0.1\textwidth]{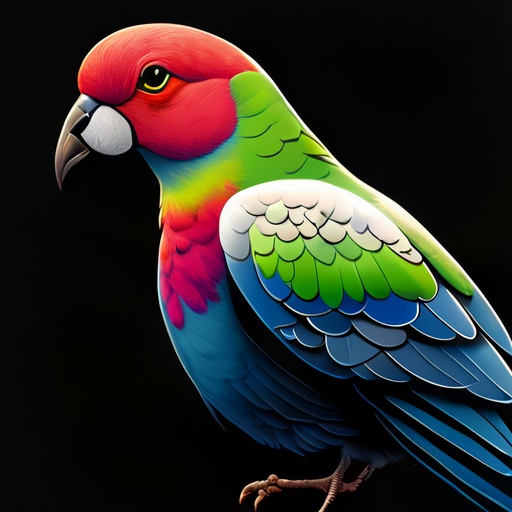} &     \includegraphics[width=0.1\textwidth]{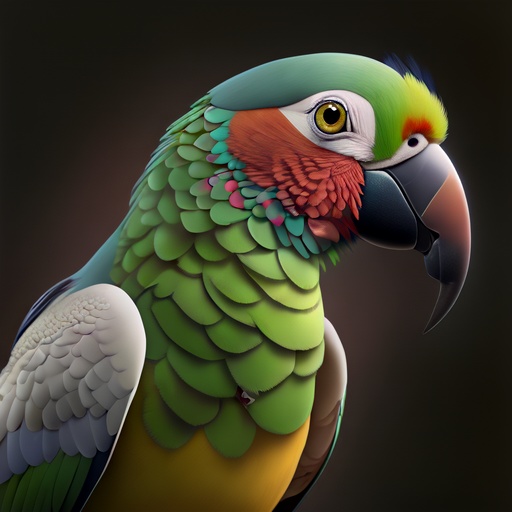} &     \includegraphics[width=0.1\textwidth]{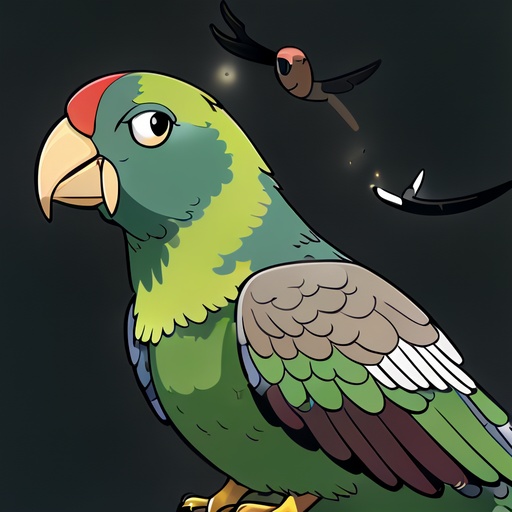} &     \includegraphics[width=0.1\textwidth]{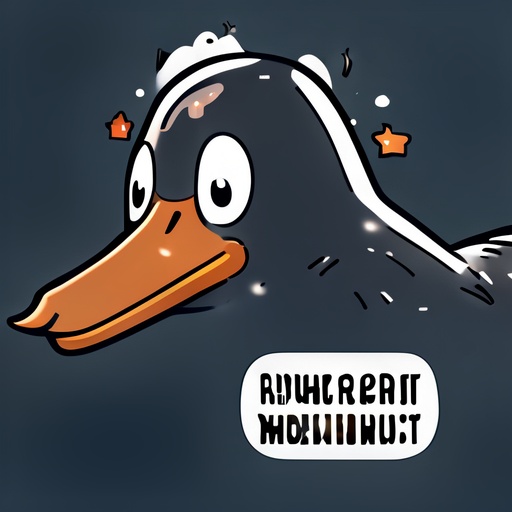} &     \includegraphics[width=0.1\textwidth]{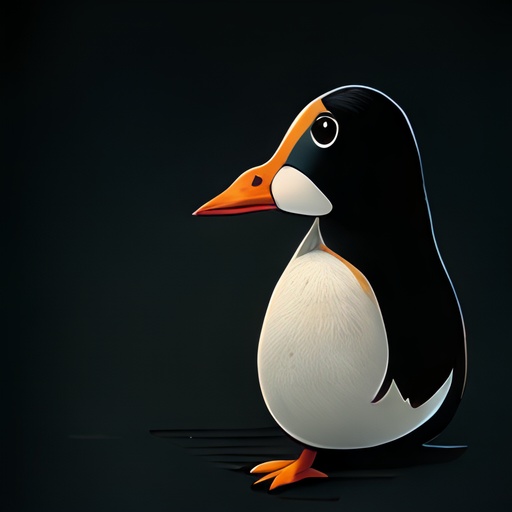} &     \includegraphics[width=0.1\textwidth]{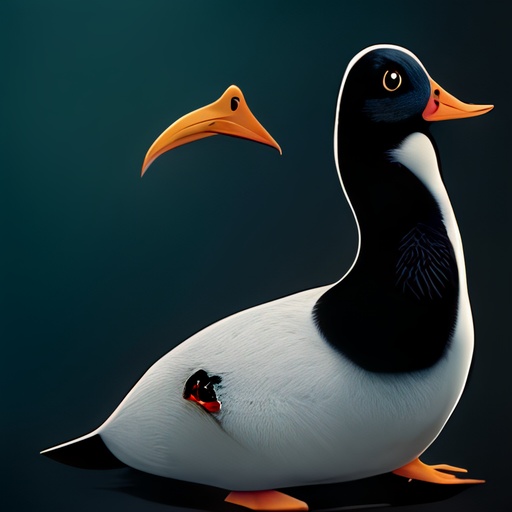} &     \includegraphics[width=0.1\textwidth]{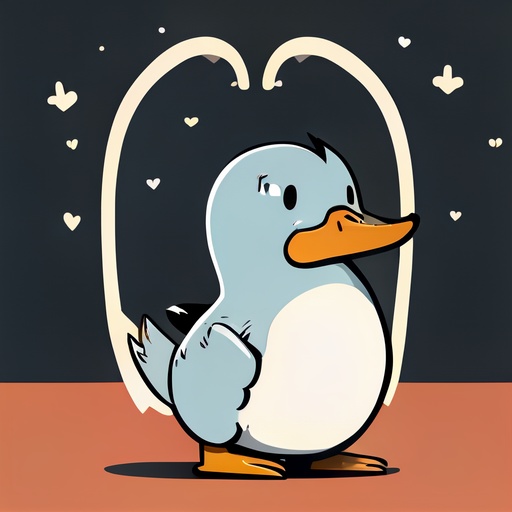} \\
 & \hspace{0.1cm}
    \includegraphics[width=0.1\textwidth]{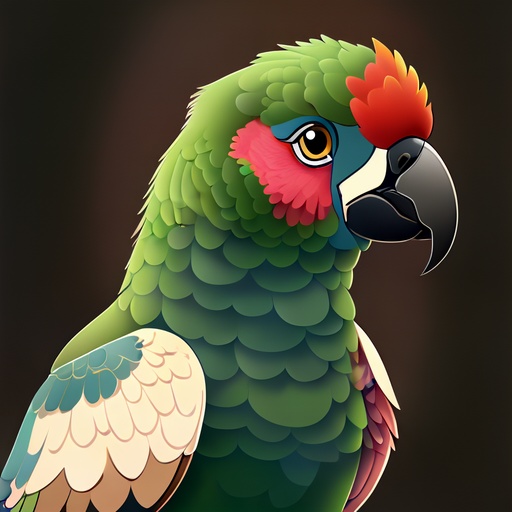} &     \includegraphics[width=0.1\textwidth]{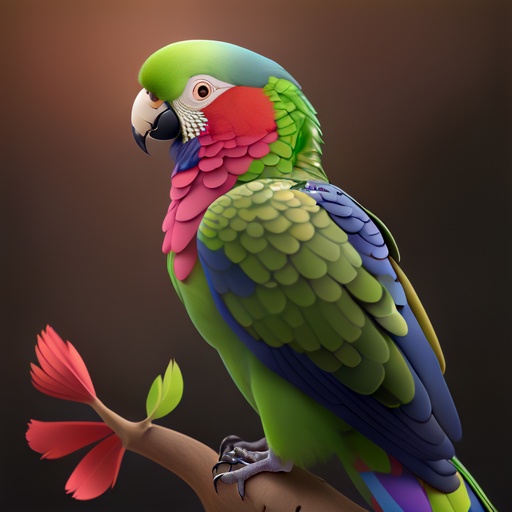} &     \includegraphics[width=0.1\textwidth]{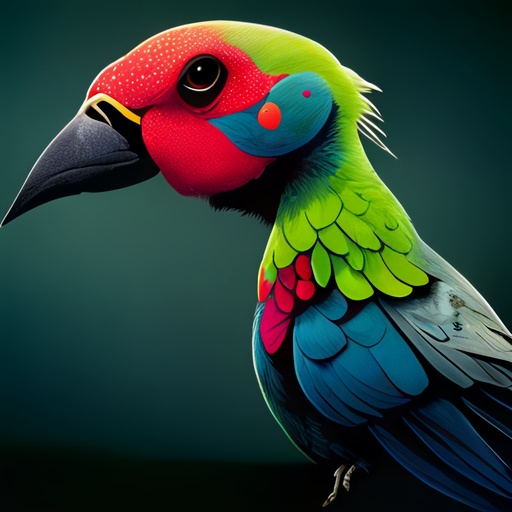} &     \includegraphics[width=0.1\textwidth]{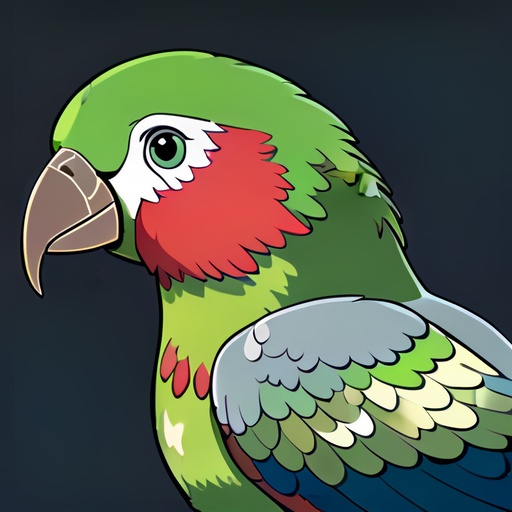} &     \includegraphics[width=0.1\textwidth]{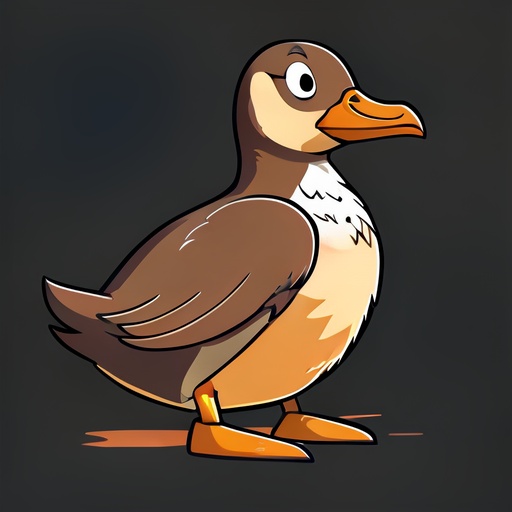} &     \includegraphics[width=0.1\textwidth]{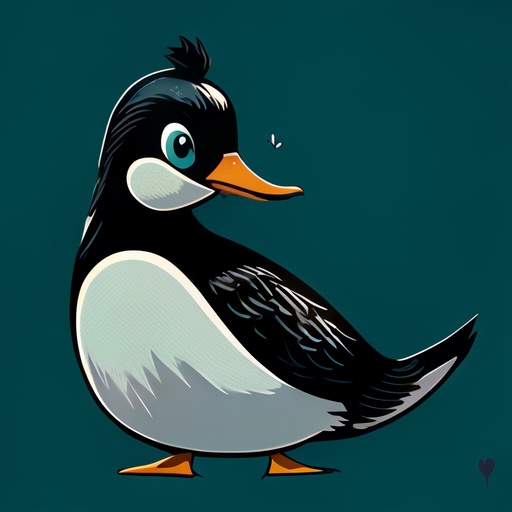} &     \includegraphics[width=0.1\textwidth]{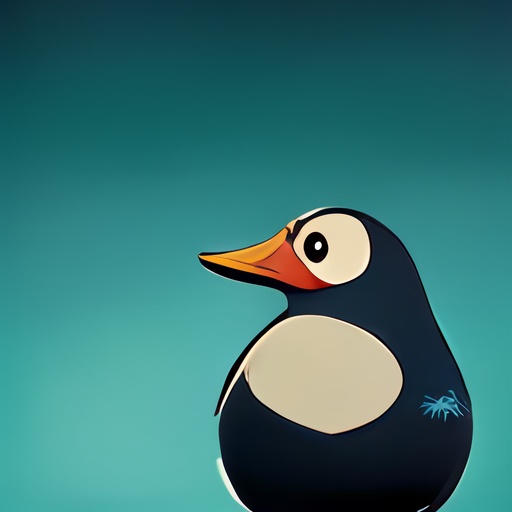} &     \includegraphics[width=0.1\textwidth]{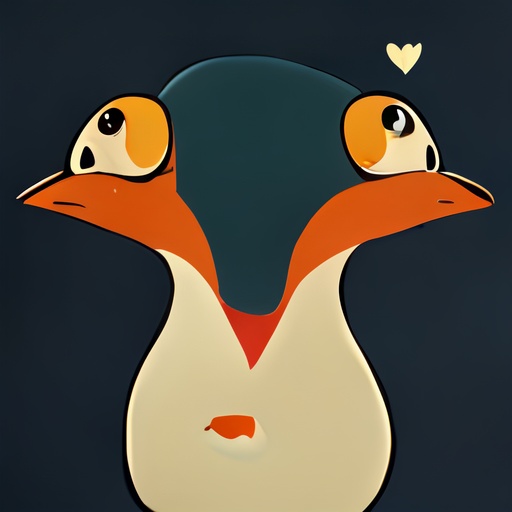} \\
 & \hspace{0.1cm}
    \includegraphics[width=0.1\textwidth]{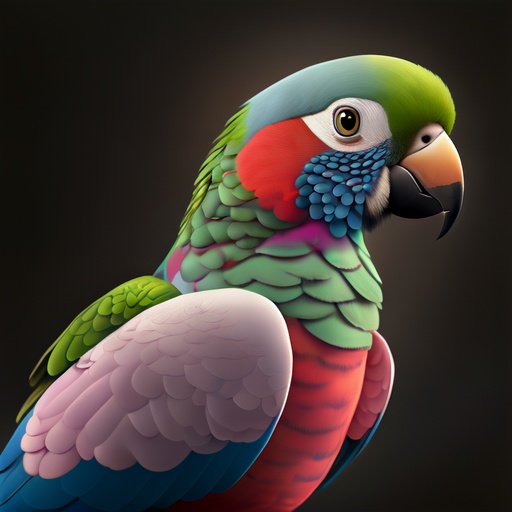} &     \includegraphics[width=0.1\textwidth]{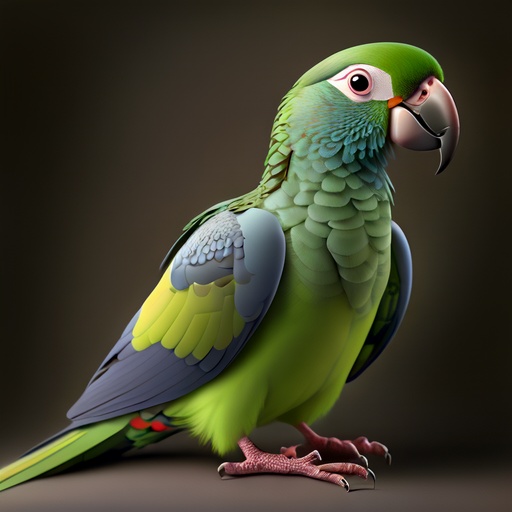} &     \includegraphics[width=0.1\textwidth]{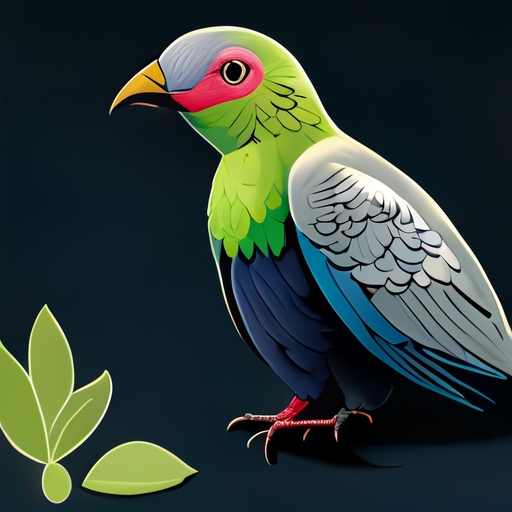} &     \includegraphics[width=0.1\textwidth]{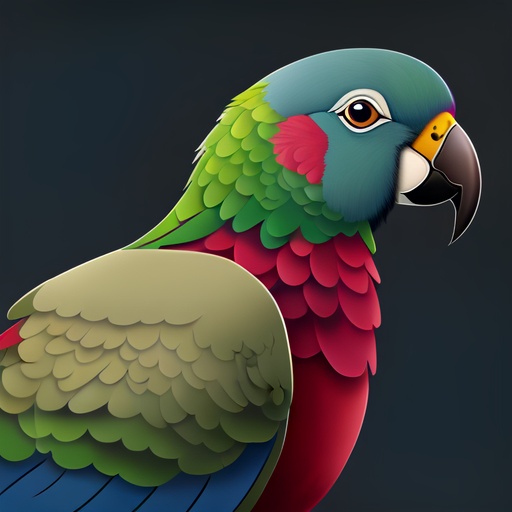} &     \includegraphics[width=0.1\textwidth]{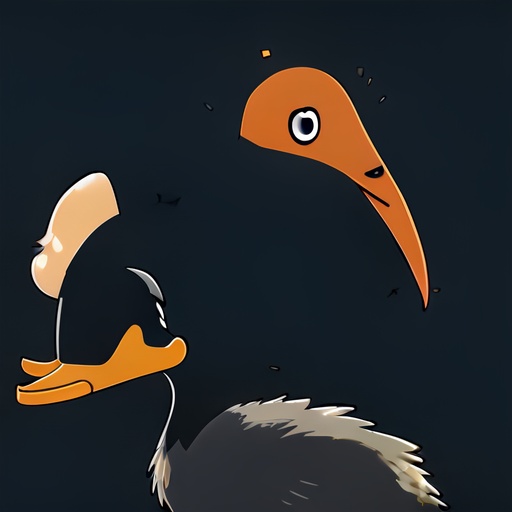} &     \includegraphics[width=0.1\textwidth]{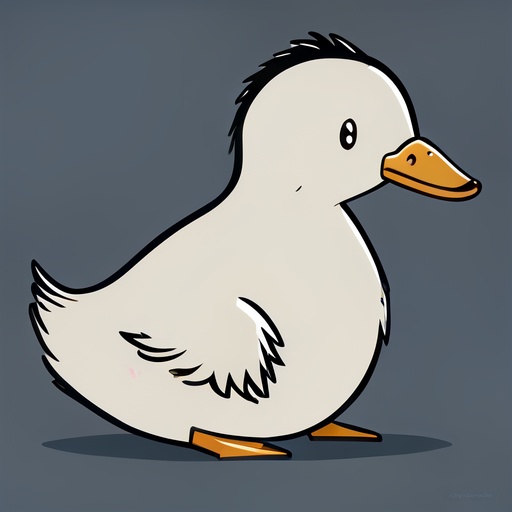} &     \includegraphics[width=0.1\textwidth]{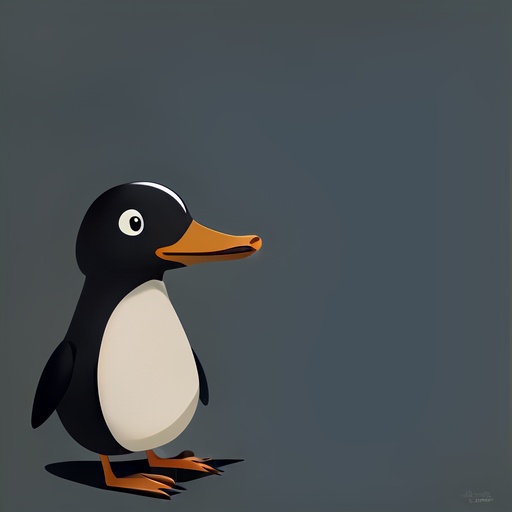} &     \includegraphics[width=0.1\textwidth]{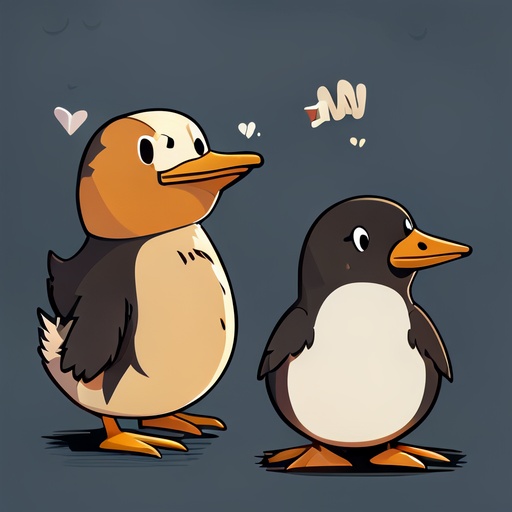} \\
    \\ \\ \\

 & \hspace{0.1cm}
    \includegraphics[width=0.1\textwidth]{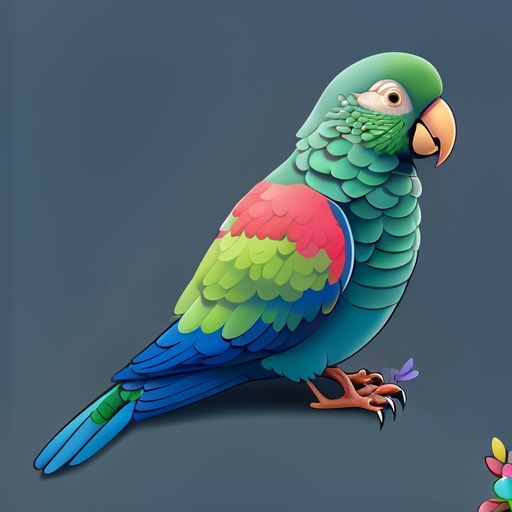} &     \includegraphics[width=0.1\textwidth]{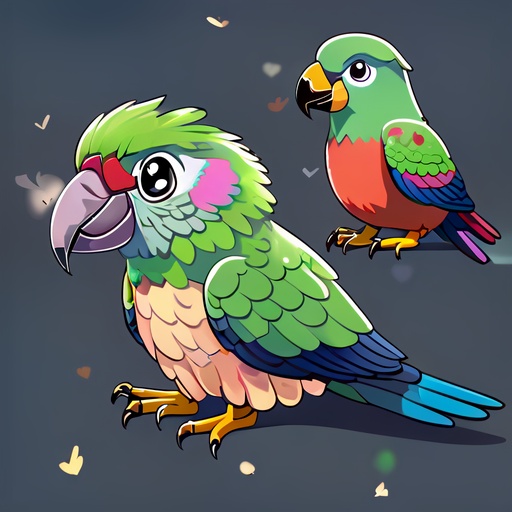} &     \includegraphics[width=0.1\textwidth]{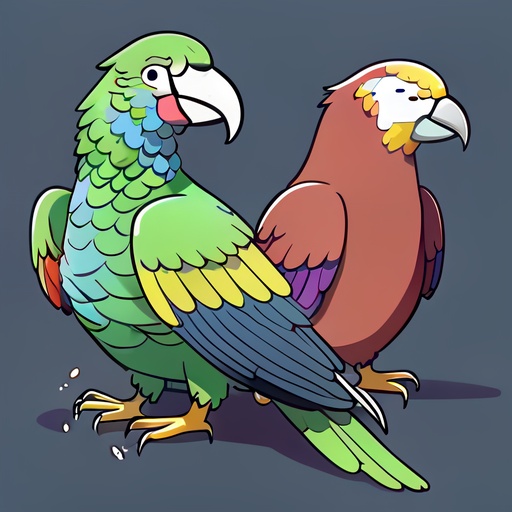} &     \includegraphics[width=0.1\textwidth]{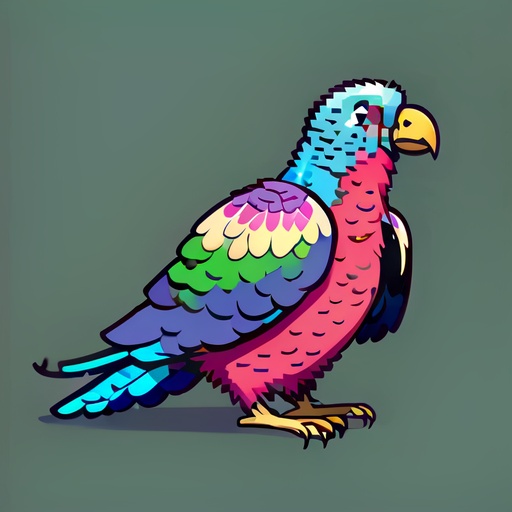} &     \includegraphics[width=0.1\textwidth]{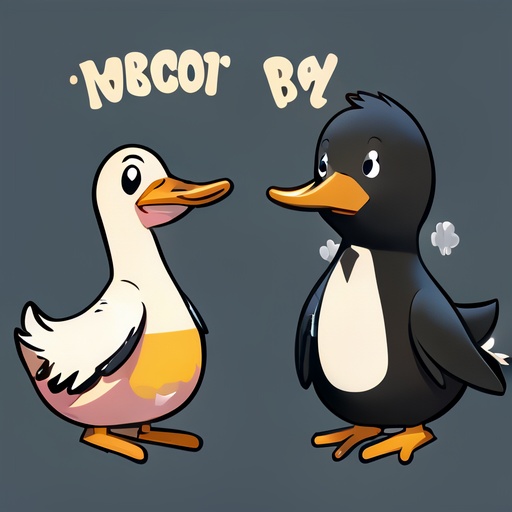} &     \includegraphics[width=0.1\textwidth]{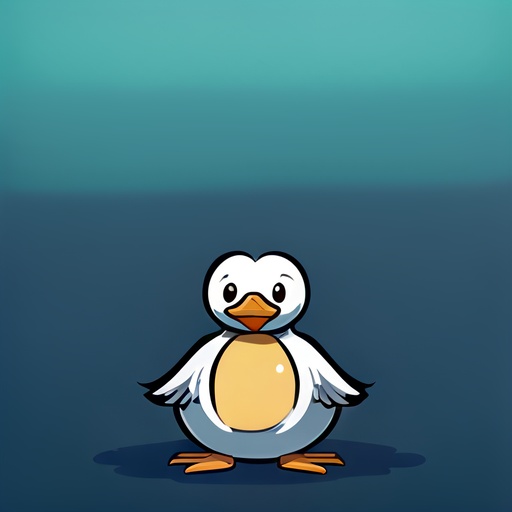} &     \includegraphics[width=0.1\textwidth]{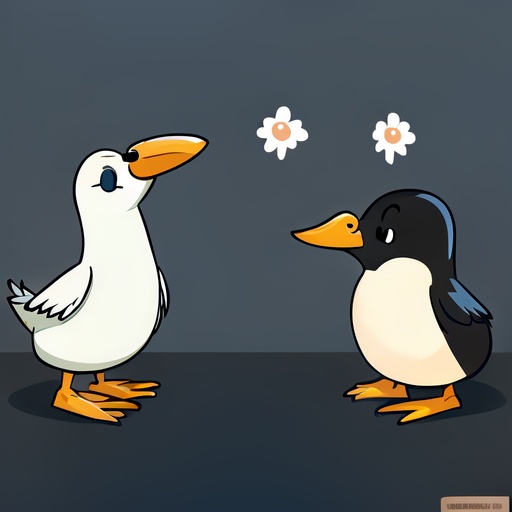} &     \includegraphics[width=0.1\textwidth]{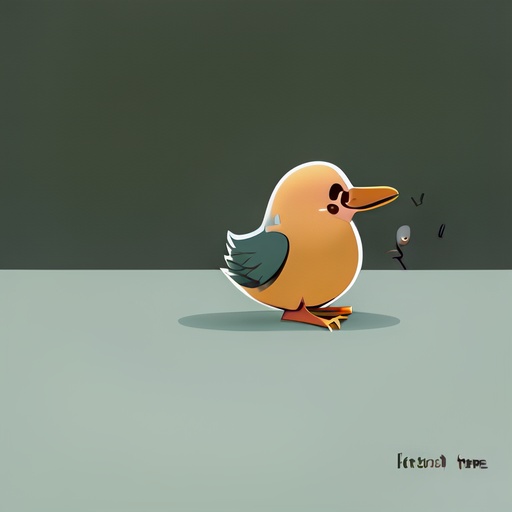} \\
{\raisebox{0.4in}{\multirow{4}{*}{\rotatebox{90}{\normalsize \textbf{\oursabbr{} (ours)}}}}} & \hspace{0.1cm}
    \includegraphics[width=0.1\textwidth]{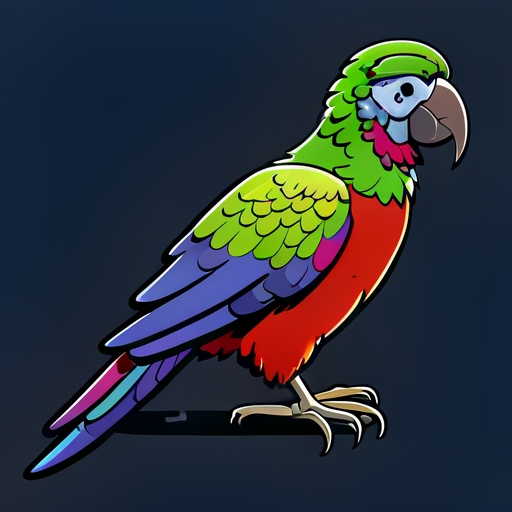} &     \includegraphics[width=0.1\textwidth]{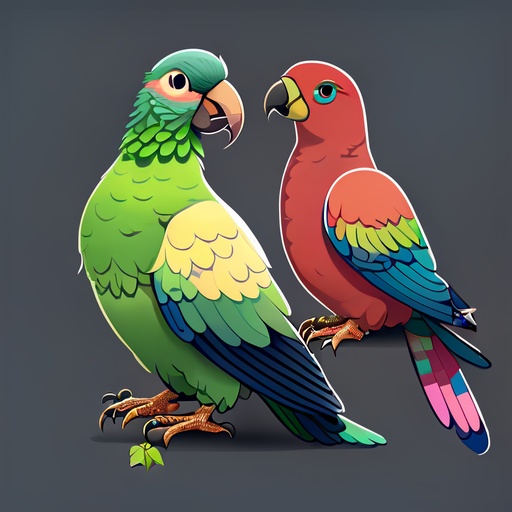} &     \includegraphics[width=0.1\textwidth]{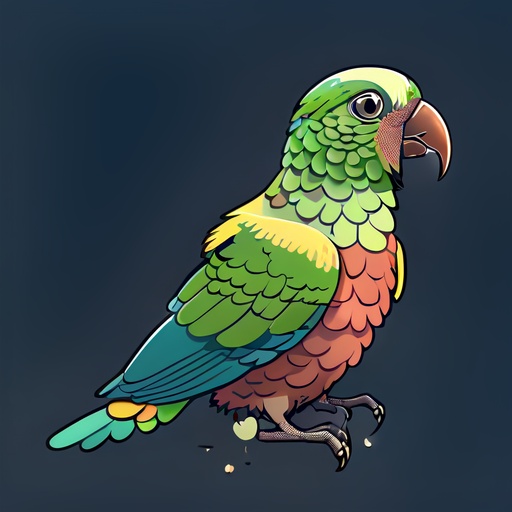} &     \includegraphics[width=0.1\textwidth]{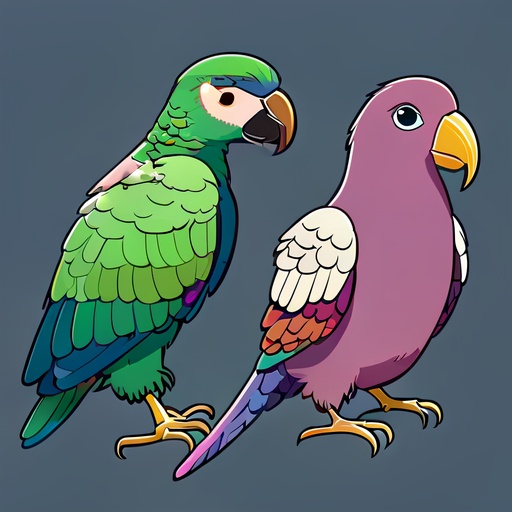} &     \includegraphics[width=0.1\textwidth]{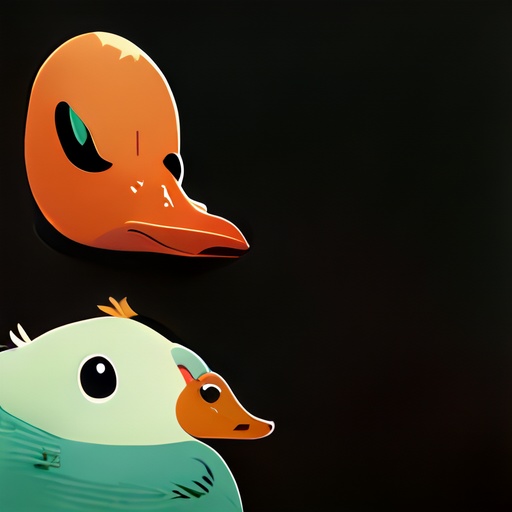} &     \includegraphics[width=0.1\textwidth]{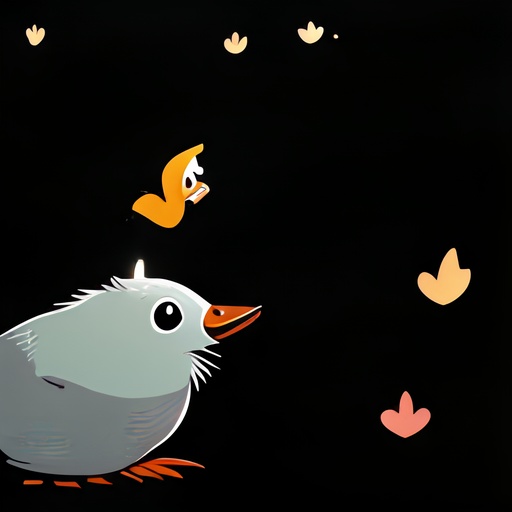} &     \includegraphics[width=0.1\textwidth]{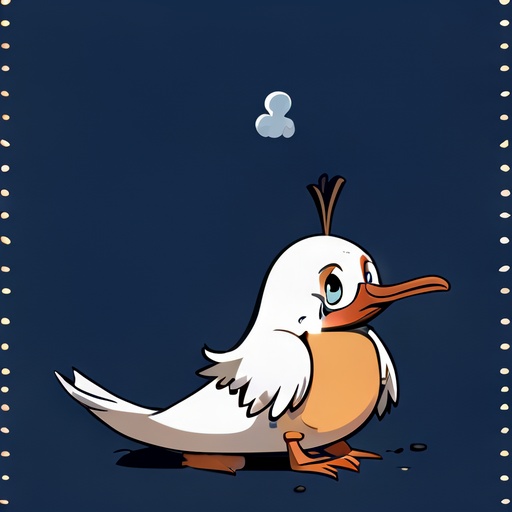} &     \includegraphics[width=0.1\textwidth]{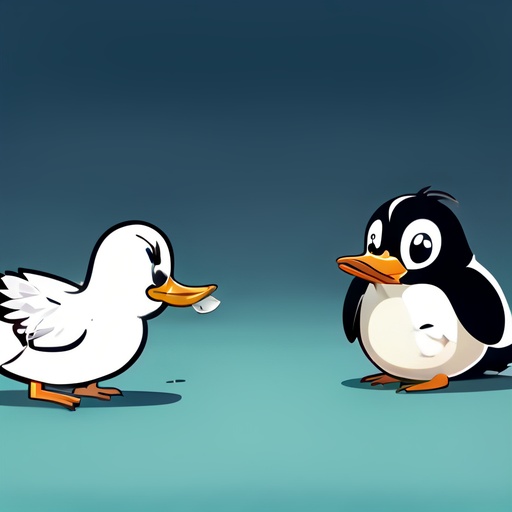} \\
 & \hspace{0.1cm}
    \includegraphics[width=0.1\textwidth]{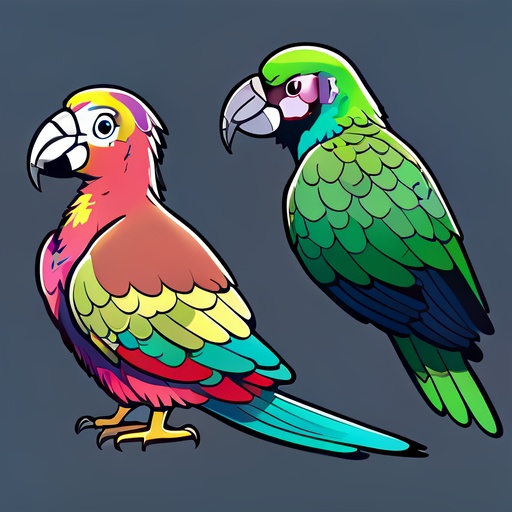} &     \includegraphics[width=0.1\textwidth]{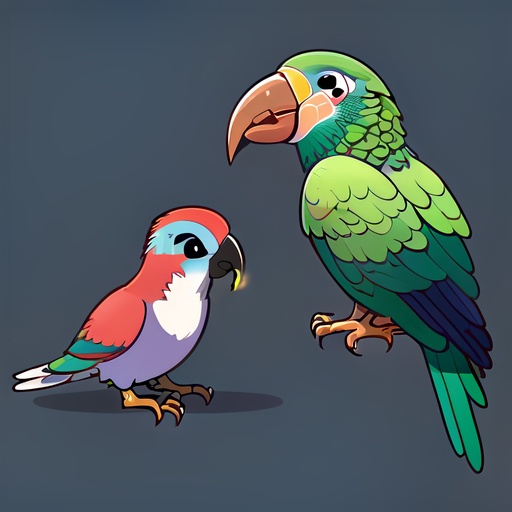} &     \includegraphics[width=0.1\textwidth]{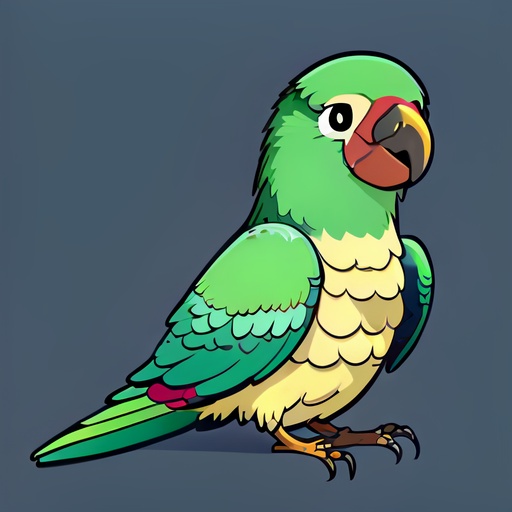} &     \includegraphics[width=0.1\textwidth]{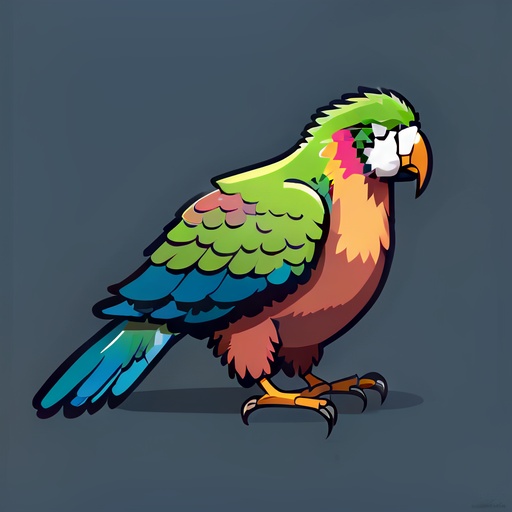} &     \includegraphics[width=0.1\textwidth]{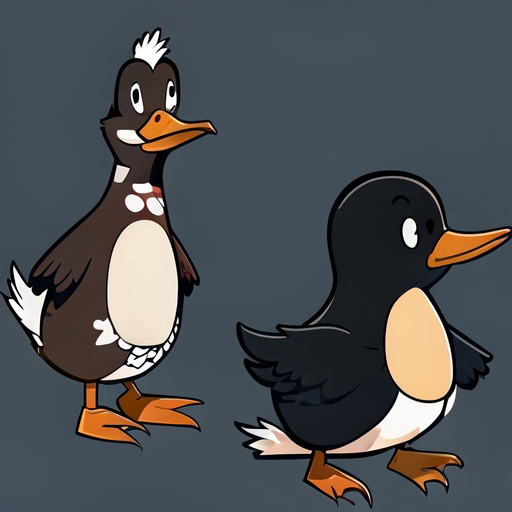} &     \includegraphics[width=0.1\textwidth]{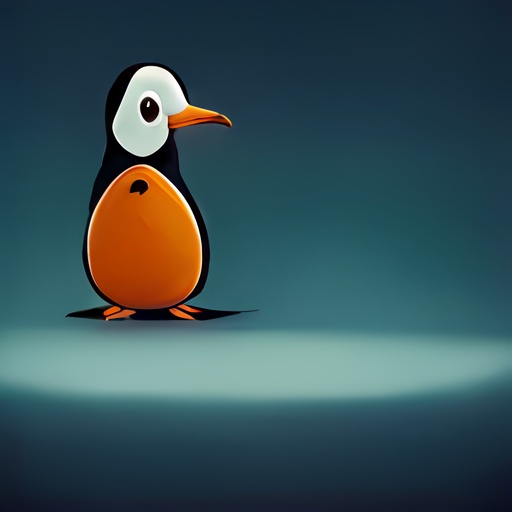} &     \includegraphics[width=0.1\textwidth]{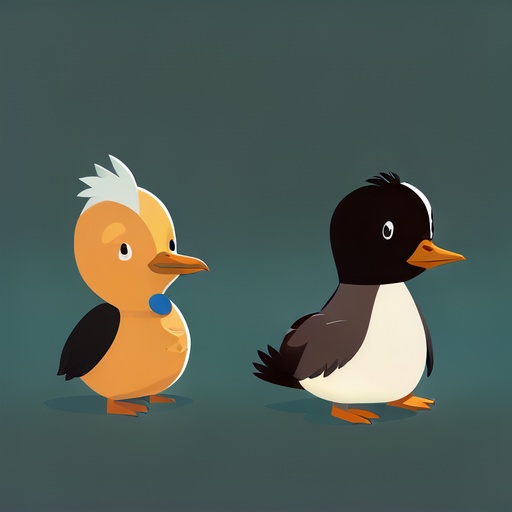} &     \includegraphics[width=0.1\textwidth]{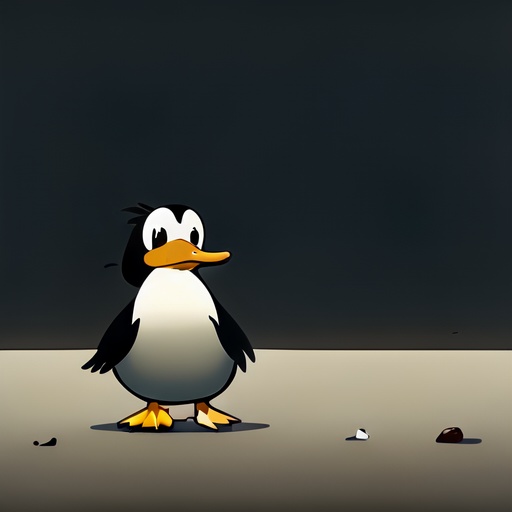} \\
 & \hspace{0.1cm}
    \includegraphics[width=0.1\textwidth]{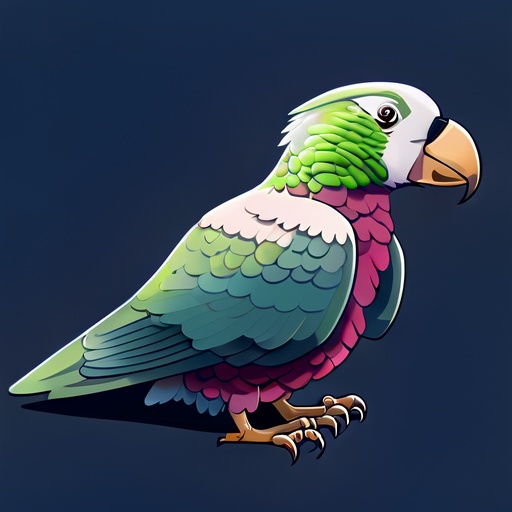} &     \includegraphics[width=0.1\textwidth]{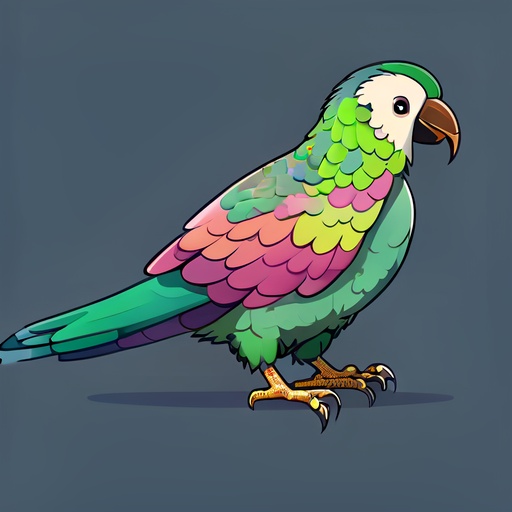} &     \includegraphics[width=0.1\textwidth]{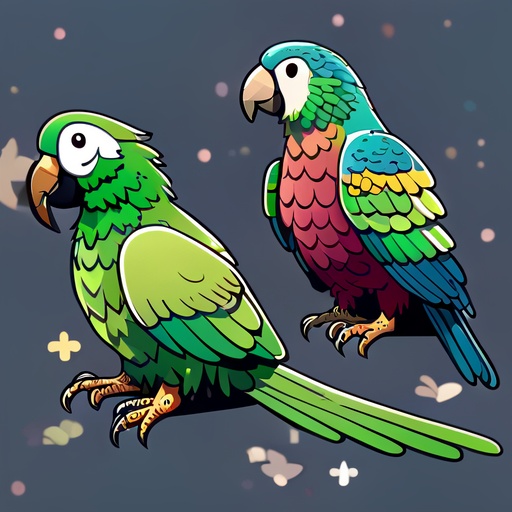} &     \includegraphics[width=0.1\textwidth]{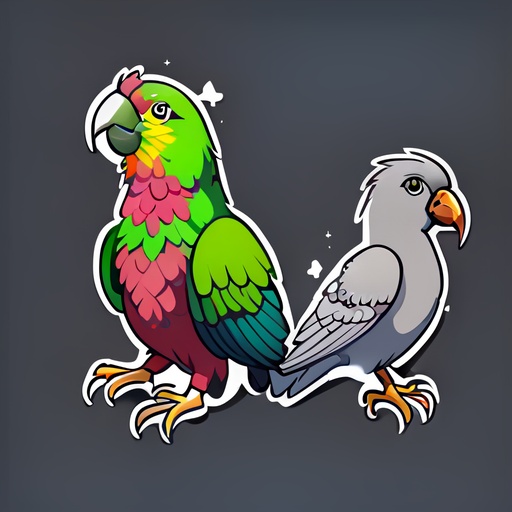} &     \includegraphics[width=0.1\textwidth]{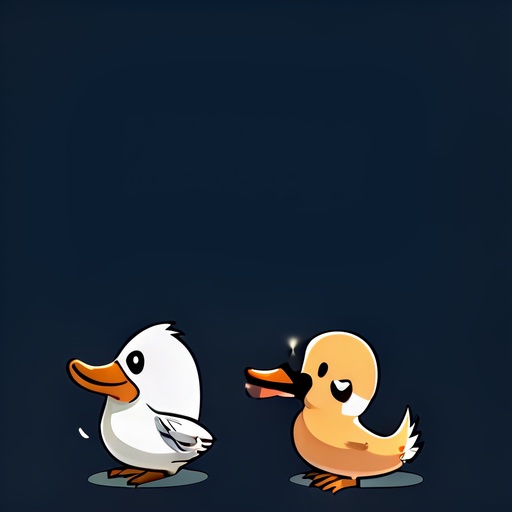} &     \includegraphics[width=0.1\textwidth]{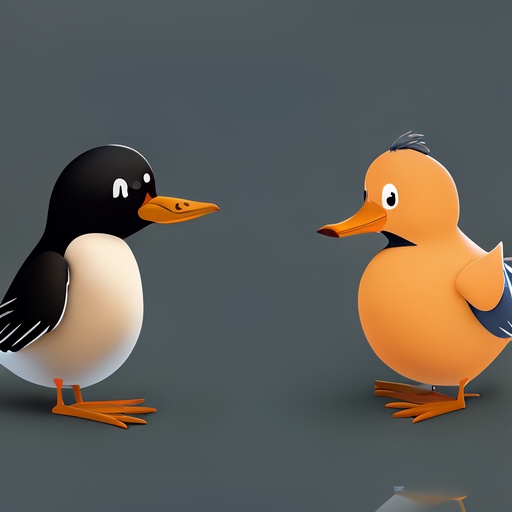} &     \includegraphics[width=0.1\textwidth]{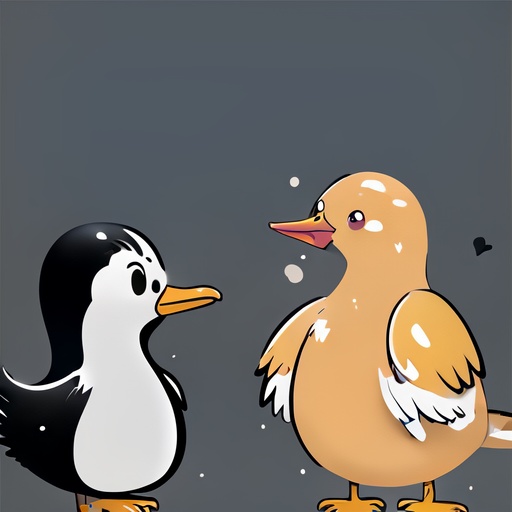} &     \includegraphics[width=0.1\textwidth]{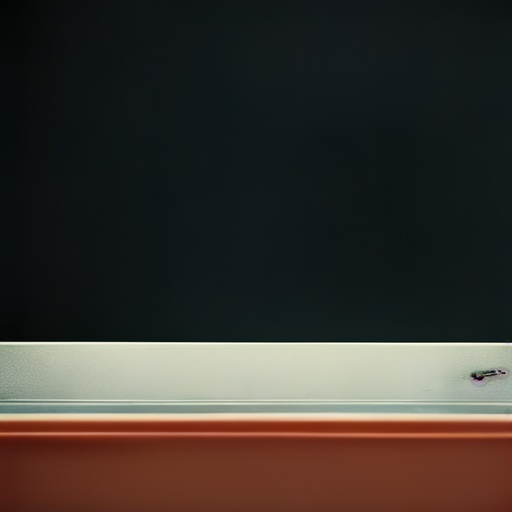} \\

\end{tabular}
    }
    }
\caption{Additional qualitative results on SSD~\citep{ssd} dataset (3). The same seeds are applied to each prompt across all methods.}
\label{fig:app_qual_ssd_3}
\end{figure*}

Additional qualitative results on the SSD dataset are provided in \cref{fig:app_qual_ssd_1,fig:app_qual_ssd_2,fig:app_qual_ssd_3}. The same seeds are applied to each prompt across all methods.

\FloatBarrier

\end{document}